\documentclass{article} 
\usepackage{iclr2027_conference,times}

\usepackage{amsmath,amsfonts,bm}

\def\eqref#1{equation~\ref{#1}}

\def\1{\bm{1}}

\DeclareMathAlphabet{\mathsfit}{\encodingdefault}{\sfdefault}{m}{sl}
\SetMathAlphabet{\mathsfit}{bold}{\encodingdefault}{\sfdefault}{bx}{n}

\usepackage{hyperref}
\usepackage{url}
\usepackage{graphicx}
\usepackage{booktabs}
\usepackage{multirow}
\usepackage{makecell}
\usepackage{pifont}
\usepackage{dsfont}    
\usepackage[T1]{fontenc}

\usepackage[table]{xcolor}
\definecolor{heatlow}{HTML}{D6EAF8}    
\definecolor{heatmid}{HTML}{FEF9E7}    
\definecolor{heathigh}{HTML}{F5B7B1}   

\definecolor{confounded}{HTML}{F5CBA7}
\definecolor{fragile}{HTML}{D2B4DE}

\usepackage{placeins}
\usepackage{morefloats}

\title{OLMo-Detect: A Multi-Stage, Confounder-Controlled Benchmark for Membership Inference on Large Language Models}

\author{
Tao Shi$^{1}$, Chaoyi Xiang$^{1}$, Qiongkai Xu$^{1,2}$ \& Jey Han Lau$^{1}$ \\
$^{1}$The University of Melbourne \quad $^{2}$Macquarie University \\
\texttt{\{tao.shi,chaoyix\}@student.unimelb.edu.au} \\
\texttt{qiongkai.xu@mq.edu.au}, \texttt{jeyhan.lau@gmail.com}
}

\iclrpreprintcopy

\begin{document}

\maketitle

\begin{abstract}

Membership inference on large language models (LLMs) aims to determine whether a given text sample was included in an LLM's training data, without access to its training corpus. Despite recent progress, existing benchmarks suffer from three limitations: limited coverage of training stages, insufficient distributional alignment between members and non-members, and lack of rigorous filtering of non-members against the training corpus. To address these limitations, we propose \textsc{OLMo-Detect}, a multi-stage, confounder-controlled benchmark built upon the fully open OLMo~2 pipeline. \textsc{OLMo-Detect} spans pre-training, mid-training, and post-training, explicitly aligns members and non-members on three key axes, and rigorously filters non-members via infini-gram. To assess robustness to distribution shifts, we further introduce \textsc{OLMo-Detect (Shifted)}, a variant where members are misaligned with non-members. We evaluate 15 unsupervised and 3 supervised membership inference attacks (MIAs) across the OLMo~2 family, finding that: (i) overall performance is limited: the best unsupervised and supervised MIAs both reach an AUC of only 0.68, and supervised MIAs degrade under cross-domain evaluation; (ii) MIA performance peaks at mid-training and is lower at pre-training and post-training, a pattern driven by data type rather than a stage effect: curated math data is far more detectable than other types; (iii) overall scores improve from 1B to 13B but plateau at 32B; and (iv) no unsupervised MIA is robust to distribution shifts, with AUCs shifting by up to 0.42. Finally, we find that our findings on OLMo~2 generalize to OLMo~3 and non-OLMo models.

\end{abstract}

\section{Introduction} \label{intro}
Membership inference on large language models (LLMs) aims to determine whether a given text sample was included in an LLM's training data, without access to its training corpus. Despite recent progress~\citep{shi2023detecting, ye2024data, zhang2024min, dong2024generalization, duan2024membership, fu2025mia, tao2025detecting, chang2025context, zhang2025fine, kim2026detecting}, there is still a lack of \textbf{multi-stage, confounder-controlled benchmarks}.

As illustrated in Table~\ref{table:1}, existing benchmarks suffer from three limitations: 
(i) \textbf{Limited coverage of training stages.} Modern LLM development \citep{grattafiori2024llama, olmo20242, yang2025qwen3} has evolved into a three-stage pipeline of pre-training, mid-training, and post-training, where mid-training refines model capabilities through targeted high-quality data and annealing-style optimization \citep{tu2025survey}. However, most existing benchmarks focus on the pre-training stage, with only a few targeting post-training.
(ii) \textbf{Insufficient distributional alignment between members and non-members.}
Most benchmarks either introduce distribution shifts, such as temporal drift~\citep{shi2023detecting, ye2024data, zhang2024pretraining, fu2025mia} and fine-tuning-induced differences~\citep{dong2024generalization}, or achieve only aggregate-level alignment by construction, such as train/held-out split~\citep{duan2024membership, tao2025detecting}, without explicitly controlling alignment on axes exploitable by existing membership inference attacks (MIAs). The only exception is OLMoMIA \citep{kim2026detecting}, which aligns on semantic similarity. However, most MIAs~\citep{carlini2021extracting, shi2023detecting, chang2023speak, zhang2024min, zhang2024pretraining, ye2024data, xie2024recall, chang2025context, golchin2025data} exploit surface-form signals that manifest through exact wording rather than meaning, so semantically aligned data may still exhibit lexical shifts. (iii) \textbf{Lack of rigorous filtering of non-members against the training corpus.} Most benchmarks construct non-members without filtering out those sharing substantial $n$-gram overlap with the model's training data. MIMIR~\citep{duan2024membership} is the sole exception but uses a 13-gram filter with a permissive threshold of 80\%, leaving its non-members with potentially high residual overlap.


To address these limitations, we propose \textsc{OLMo-Detect}, a multi-stage, confounder-controlled benchmark built upon the fully open training pipeline of OLMo~2~\citep{olmo20242}. Specifically: (i) it spans pre-training, mid-training, and post-training, enabling systematic analysis of how detection performance varies across training stages; (ii) it explicitly aligns members and non-members on three key axes to minimize distribution shifts: text quality, temporal range, and lexical similarity; and (iii) it filters non-members against OLMo~2's training data via infini-gram~\citep{liu2024infini}-based 13-gram filtering with a strict threshold of 20\%, leaving non-members with an average residual overlap of only 5.31\%. Moreover, we introduce \textsc{OLMo-Detect (Shifted)}, a benchmark variant where members are misaligned with non-members, to assess robustness to distribution shifts.


\begin{table*}[t]
\centering
\resizebox{1\textwidth}{!}{%
\begin{tabular}{@{}l@{\;\;}l@{\;\;}l@{\;\;}l@{\;\;}lccc@{}}
\toprule
\multirow{2.5}{*}{\textbf{Benchmark}} &
\multirow{2.5}{*}{\textbf{Stage(s)}} &
\multirow{2.5}{*}{\textbf{Domain(s)}} &
\multirow{2.5}{*}{\textbf{\makecell[l]{Member}}} &
\multirow{2.5}{*}{\textbf{\makecell[l]{Non-Member}}} &
\multicolumn{3}{c}{\textbf{Assessments}} \\
\cmidrule(l){6-8}
& & & & & \textbf{C} & \textbf{A} & \textbf{F} \\
\midrule
WikiMIA \citep{shi2023detecting} & Pre-training & Wikipedia & \makecell[l]{Pre-2017 data} & \makecell[l]{Post-2023 data} & {\ding{55}} & {\ding{55}} & {\ding{55}} \\ \midrule
StackMIA \citep{ye2024data} & Pre-training & \makecell[l]{Stack Exchange} & \makecell[l]{Pre-2017 data} & \makecell[l]{Post-2023 data} & {\ding{55}} & {\ding{55}} & {\ding{55}} \\ \midrule
DETCON \citep{dong2024generalization} & \makecell[l]{Post-training} & \makecell[l]{HumanEval, GSM8K} & \makecell[l]{Outputs of post-SFT LLMs} & \makecell[l]{Outputs of base LLMs} & {\ding{55}} & {\ding{55}} & {\ding{55}} \\ \midrule
MIMIR \citep{duan2024membership} & Pre-training & \makecell[l]{The Pile: 7 domains} & Train splits & \makecell[l]{Held-out data} & {\ding{55}} & {\ding{55}} & {\ding{51}}\textsuperscript{\dag} \\ \midrule
PatentMIA \citep{zhang2024pretraining} & Pre-training & \makecell[l]{Google Patents} & \makecell[l]{Pre-2023 data} & \makecell[l]{Post-2024 data} & {\ding{55}} & {\ding{55}} & {\ding{55}}  \\ \midrule
WikiMIA-24 \citep{fu2025mia} & Pre-training & Wikipedia & \makecell[l]{Pre-2017 data} & \makecell[l]{Post-2024 data} & {\ding{55}} & {\ding{55}} & {\ding{55}} \\ \midrule
RL-MIA \citep{tao2025detecting}
  & \makecell[l]{Post-training}
  & \makecell[l]{AIME24/25, K\&K, SAT}
  & \makecell[l]{Training data}
  & \makecell[l]{Held-out data}
  & {\ding{55}}
  & {\ding{55}}
  & {\ding{55}} \\ \midrule
OLMoMIA \citep{kim2026detecting} & Pre-training & Dolma & \makecell[l]{Pre-checkpoint data} & \makecell[l]{Post-checkpoint data} & {\ding{55}} & {\ding{51}}\textsuperscript{\ddag} & {\ding{55}} \\ \midrule
\textbf{\textsc{OLMo-Detect}} (ours) & \makecell[l]{Pre-training,\\Mid-training,\\Post-training} & \makecell[l]{9 domains of\\OLMo~2} & \makecell[l]{Distribution-matched\\training data} & \makecell[l]{Distribution-matched, \\rigorously filtered\\non-members} & \ding{51} & \ding{51} & \ding{51} \\
\bottomrule
\end{tabular}%
}
\caption{
Comparison of existing benchmarks for membership inference on LLMs. We assess on three criteria: (i) \textbf{Coverage (C)} of training stages (pre-training, mid-training, and post-training); (ii) \textbf{Explicit Alignment (A)} of distributions between members and non-members; (iii) \textbf{Rigorous Filtering (F)} of non-members against the training corpus. \ding{51}\textsuperscript{\dag} indicates that MIMIR uses a filter with a permissive threshold of 80\%, leaving potentially high residual overlap. \ding{51}\textsuperscript{\ddag} means that OLMoMIA aligns on semantic similarity, but most MIAs target exact wording rather than meaning.
}
\label{table:1}
\end{table*}

We evaluate 15 unsupervised and 3 supervised MIAs across the OLMo~2 family (1B/7B/13B/32B) on \textsc{OLMo-Detect}, finding that:
(i) \textbf{Overall performance is limited.} The best unsupervised method is Zlib~\citep{carlini2021extracting}, one of the simplest MIAs, while the best supervised method is Sup-CAMIA~\citep{chang2025context}; both achieve an average AUC of 0.68. Moreover, supervised MIAs degrade under cross-domain evaluation, indicating that learned signals do not transfer well.
(ii) \textbf{Detectability is driven by data type rather than training stage.} Overall performance peaks at mid-training and is lower at pre-training and post-training, but this ordering reflects data type rather than a stage effect: curated math data (e.g., GSM8K~\citep{cobbe2021trainingverifierssolvemath}) is considerably more detectable than other types.
(iii) Overall scores improve from 1B to 13B but plateau at 32B, suggesting that \textbf{scale alone may not continue to make detection easier.}
Additionally, we evaluate unsupervised MIAs on \textsc{OLMo-Detect (Shifted)} and find that \textbf{no unsupervised MIA is robust to distribution shifts}, with AUCs shifting by up to 0.42. Finally, we evaluate the top five unsupervised MIAs on OLMo~3 (7B/32B)~\citep{olmo2025olmo} and two non-OLMo families, DCLM-Baseline (1B/7B)~\citep{li2024datacomp} and SmolLM2-1.7B~\citep{allal2025smollm2smolgoesbig}, on the \textsc{OLMo-Detect} domains they were trained on, and find that \textbf{our main findings generalize beyond OLMo~2}.



\section{Related Work} \label{related_work}

As summarized in Table~\ref{table:1}, existing benchmarks for membership inference on LLMs can be grouped into four categories based on how members and non-members are constructed: (i) temporal-cutoff~\citep{shi2023detecting, ye2024data, zhang2024pretraining, fu2025mia}; (ii) partition-based~\citep{duan2024membership, tao2025detecting}; (iii) checkpoint-based~\citep{kim2026detecting}; and (iv) model-output-based~\citep{dong2024generalization}. Detailed descriptions are provided in Appendix~\ref{appendix:related_work_benchmarks}. 

Existing MIAs can be grouped into unsupervised and supervised methods based on whether they use labeled members or non-members. Unsupervised MIAs can be classified into four categories: (i) likelihood-based \citep{yeom2018privacy, carlini2021extracting, shi2023detecting, zhang2024min, zhang2024pretraining, ye2024data, xie2024recall, fu2024membership, chang2025context, kim2026detecting}; (ii) perturbation-based \citep{mattern2023membership}; (iii) prompt-based \citep{golchin2023time, chang2023speak, deng2024investigating, golchin2025data}; and (iv) output-distribution-based \citep{dong2024generalization, tao2025detecting}. Supervised MIAs include a supervised variant of CAMIA \citep{chang2025context} and tuning-based methods \citep{fu2025mia, zhang2025fine}. Details are provided in Appendix~\ref{appendix:related_work_methods}.

\section{Task Definition} \label{task_def}
Let $\mathcal{M}$ denote an LLM trained on an unknown corpus $\mathcal{D}$. Given a candidate text $x$, the task of membership inference on LLMs is to infer whether $x \in \mathcal{D}$ via a scoring function $s(x; \mathcal{M})$, thresholded at $\lambda$: $\text{prediction}(x; \mathcal{M}) = \mathds{1}[s(x; \mathcal{M}) \geq \lambda]$, where $1$ and $0$ denote member and non-member, respectively. Following prior work \citep{zhang2024min}, we assume grey-box access to $\mathcal{M}$: output statistics (loss, logits, token probabilities) are available, internal parameters (weights, gradients) are not. Some methods further assume the ability to fine-tune $\mathcal{M}$, described in Appendix~\ref{appendix:related_work_methods_fine_tuning}.

\section{Benchmark Construction}

\subsection{Motivation for Using OLMo~2}
Constructing benchmarks requires LLMs with open training data to verify members and filter non-members against the corpus. Among existing open-source LLMs, GPT-NeoX \citep{black2022gpt} and Pythia \citep{biderman2023pythia} are limited to pre-training; OLMo~1 \citep{groeneveld-etal-2024-olmo} additionally includes post-training, but only for its 7B variant, and lacks mid-training; only OLMo~2 \citep{olmo20242} and OLMo~3 \citep{olmo2025olmo} provide complete access across all three stages. We construct \textsc{OLMo-Detect} based on OLMo~2, as it was the available model when we started this work. Nonetheless, we show that patterns found on OLMo~2 generalize to OLMo~3 (Section~\ref{generalization}).

\subsection{Training Data of OLMo~2} \label{sec:olmo2_data}

We summarize the training data of OLMo~2 as follows: (i) \textbf{pre-training}: DCLM-Baseline (web), StarCoder (code), peS2o and arXiv (academic papers), OpenWebMath (web math), Algebraic Stack (math proofs and code), and Wikipedia/Wikibooks (encyclopedic); (ii) \textbf{mid-training}: three reused from pre-training (filtered DCLM-Baseline, peS2o, and Wikipedia/Wikibooks) and three new domains: FLAN (instruction data), Stack Exchange (Q\&A), and Math Mix (curated math); and (iii) \textbf{post-training}: curated data for supervised fine-tuning (SFT), direct preference optimization (DPO), and reinforcement learning with verifiable rewards (RLVR). Details are provided in Appendix~\ref{appendix:olmo2_training_data}.


\begin{table*}[h]
\centering
\resizebox{1\textwidth}{!}{%
\begin{tabular}{@{}c@{\;\;}l@{\;\;}l@{\;\;}l@{\;\;}l@{\;\;}r@{\;\;}r@{\;\;}l@{\;\;}c@{\;\;}l@{}}
\toprule
\textbf{Stage} & \textbf{Domain} & \textbf{Subsets} & \textbf{Split} & \textbf{Samples} & \textbf{Tokens} & \textbf{Quality} & \textbf{Temporal} & \textbf{Lexical} \\
\midrule
\multirow{13}{*}{Pre-training}
& \multirow{2}{*}{DCLM-Baseline} & 
  & NM & 150 & 217{,}117 & FastText: 0.01809 $\pm$ 0.00007 & 2018-01-19 $\pm$ 930d & \multirow{2}{*}{0.07} \\
& & & M & 150 & 217{,}016 & FastText: 0.01811 $\pm$ 0.00000 & 2018-01-18 $\pm$ 940d & \\
\cmidrule(l){2-9}
& \multirow{2}{*}{peS2o} & 
  & NM & 268 & 216{,}981 & --- & 2022-12-14 $\pm$ 11d & \multirow{2}{*}{0.10} \\
& & & M & 268 & 217{,}230 & --- & 2022-12-15 $\pm$ 13d & \\
\cmidrule(l){2-9}
& \multirow{2}{*}{OpenWebMath} & 
  & NM & 151 & 217{,}183 & --- & 2019-10-13 $\pm$ 853d & \multirow{2}{*}{0.11} \\
& & & M & 151 & 217{,}252 & --- & 2019-10-17 $\pm$ 864d & \\
\cmidrule(l){2-9}
& \multirow{4}{*}{StarCoder} & \multirow{4}{*}{\makecell[l]{Python,\\Java}}
  & NM & 156 & 217{,}156
  & \begin{tabular}[c]{@{}l@{\hspace{4pt}}r@{\hspace{2pt}}l@{}}
      GitHub Stars: & 35.65 & $\pm$ 66.66 \\
      Top-1 Freq:   & 0.30 & $\pm$ 0.02 \\
      Top-2 Freq:   & 0.38 & $\pm$ 0.05
    \end{tabular}
  & ---
  & \multirow{4}{*}{0.15} \\
& & & M & 156 & 217{,}222
  & \begin{tabular}[c]{@{}l@{\hspace{4pt}}r@{\hspace{2pt}}l@{}}
      GitHub Stars: & 36.15 & $\pm$ 64.67 \\
      Top-1 Freq:   & 0.29 & $\pm$ 0.01 \\
      Top-2 Freq:   & 0.36 & $\pm$ 0.05
    \end{tabular}
  & ---
  & \\
\midrule
\multirow{6}{*}{Mid-training}
& \multirow{2}{*}{Math Mix} & \multirow{2}{*}{GSM8K}
  & NM & 1{,}040 & 217{,}163 & --- & --- & \multirow{2}{*}{0.04} \\
& & & M & 1{,}040 & 217{,}132 & --- & --- & \\
\cmidrule(l){2-9}
& \multirow{3}{*}{Stack Exchange} & \multirow{3}{*}{\makecell[l]{Stack Overflow,\\Math Stack Exchange}}
  & NM & 430 & 217{,}208
  & \begin{tabular}[c]{@{}l@{\hspace{4pt}}l@{}}
      Question Vote: & 3.45 $\pm$ 2.19 \\
      Answer Vote:   & 5.12 $\pm$ 1.99
    \end{tabular}
  & 2015-03-31 $\pm$ 1378d
  & \multirow{3}{*}{0.08} \\
& & & M & 430 & 217{,}202
  & \begin{tabular}[c]{@{}l@{\hspace{4pt}}l@{}}
      Question Vote: & 3.52 $\pm$ 1.74 \\
      Answer Vote:   & 5.54 $\pm$ 1.94
    \end{tabular}
  & 2015-03-30 $\pm$ 1367d
  & \\
\midrule
\multirow{7}{*}{Post-training}
& \multirow{2}{*}{SFT$^{\dagger}$} & \multirow{2}{*}{\makecell[l]{WildChat,\\Aya}}
  & NM & 225 & 217{,}057 & --- & --- & \multirow{2}{*}{0.11} \\
& & & M & 225 & 217{,}159 & --- & --- & \\
\cmidrule(l){2-9}
& \multirow{2}{*}{DPO$^{\dagger}$} & \multirow{2}{*}{UF-WildChat}
  & NM & 113 & 217{,}045 & --- & --- & \multirow{2}{*}{0.09} \\
& & & M & 113 & 217{,}287 & --- & --- & \\
\cmidrule(l){2-9}
& \multirow{2}{*}{RLVR} & \multirow{2}{*}{\makecell[l]{CoT-GSM8K,\\CoT-MATH}}
  & NM & 290 & 217{,}110 & --- & --- & \multirow{2}{*}{0.01} \\
& & & M & 290 & 217{,}025 & --- & --- & \\
\bottomrule
\end{tabular}%
}
\caption{Overall per-domain statistics of \textsc{OLMo-Detect}. Within each domain, members and non-members are aligned on up to three axes: Text \textbf{Quality}, \textbf{Temporal} Range, and \textbf{Lexical} Similarity. Text Quality uses OLMo~2's metric and is enforced for quality-thresholded domains; Temporal Range is enforced when timestamps are available; and Lexical Similarity, measured by Jensen--Shannon divergence (lower is better), is enforced for all domains. ``---'' denotes uncontrolled axes. \textbf{NM}: non-member; \textbf{M}: member. \textbf{Subsets} lists constituent subsets; blanks indicate single-subset domains whose subset shares the domain's name. $^{\dagger}$SFT and DPO are constructed per model size due to differing training data; values shown are for 1B. Each domain is split into dev/test sets at approximately 1:9 sample and token ratios; only totals are shown. Full statistics are in Appendix~\ref{appendix:full_benchmark_statistics}.}
\label{table:2}
\end{table*}

\subsection{Benchmark Overview}
\label{sec4.3}

As summarized in Table~\ref{table:2}, \textsc{OLMo-Detect} comprises nine domains spanning three training stages of OLMo~2's training data. Mid-training domains reused from pre-training are excluded; four domains are additionally excluded: pre-training's Wikipedia/Wikibooks (no source-matched non-members available), pre-training's arXiv and Algebraic Stack (no 13-gram-filtered non-members survived), and mid-training's FLAN (no instance-length-matched members available).

\subsection{Construction of Non-Members} \label{sec4.4}
For each domain, we construct non-members by (i) identifying source-matched candidate pools, (ii) sampling candidates with strategies differing by domain type, (iii) retaining those that pass the infini-gram-based 13-gram filter, and (iv) allocating token budgets. Details are provided in Appendix~\ref{appendix:non_member_construction}.

\textbf{Source-Matched Candidate Pools.}  
We first identify domains for which non-members can be drawn from the same underlying source as OLMo~2's training data, from either held-out data or the remainder of the original source corpus not used to train OLMo~2. (i) \textbf{Domains with Held-Out Data.} Pre-training: peS2o and OpenWebMath; mid-training: Math Mix's GSM8K; post-training: RLVR's CoT-GSM8K and CoT-MATH. (ii) \textbf{Domains with the Remainder of Original Source Corpora.} Pre-training: DCLM-Baseline and StarCoder; mid-training: Stack Exchange; post-training: SFT, DPO, and RLVR's IFEval. Among them, three are obtained by quality-thresholding the original corpora: DCLM-Baseline (fastText score $\geq$ 0.01811), StarCoder (GitHub stars $\geq$ 2; top-1 word frequency $\leq$ 30\% and top-2 word frequencies $\leq$ 50\%), and Stack Exchange (question votes $\geq$ 3 and answer votes $\geq$ 5); the rest are obtained via random sampling from the original source corpora. For domains with multiple sampleable subsets, we select two from the top-5 subsets by training provenance to capture within-domain diversity.

\textbf{Domain-Dependent Sampling Strategies.} \label{sec:sampling_strategies}
We then sample candidates based on whether a domain is quality-thresholded. (i) \textbf{Random Sampling for Domains without Quality Thresholding.}
For domains whose candidate pools come from either held-out data or the remainder of original source corpora without quality thresholding, we randomly sample candidate non-members. (ii) \textbf{Boundary Sampling for Quality-Thresholded Domains.}
For quality-thresholded domains, random sampling would yield data of systematically lower quality than that used to train OLMo~2. We thus apply \textbf{boundary sampling}, selecting candidates closest to OLMo~2's quality threshold. Since these candidates are drawn from the rejected remainder, they fall immediately below the threshold.

\textbf{Infini-Gram-based 13-Gram Filtering.} \label{13_gram_filtering}
Finally, we filter out candidates with substantial training-data overlap via infini-gram. Infini-gram \citep{liu2024infini} is a suffix-array-based indexing system that supports efficient $n$-gram search over massive text corpora for arbitrarily large $n$. For each model size, we index its full training corpus, and apply 13-gram filtering with a strict threshold of $20\%$, a strategy established by \citet{xu2025infini}, rather than the permissive 80\% used by MIMIR. Specifically, for each candidate, we extract all 13-grams $G$ and compute the fraction that appear in the index, $\eta = \frac{|\{g \in G : \mathrm{count}(g) > 0\}|}{|G|}$. A candidate is retained as a non-member if $\eta < 20\%$.

\textbf{Token Budget Allocation.} \label{token_budget}
From the resulting non-members, we sample toward a uniform budget of approximately 217K tokens per domain, determined by Math Mix's non-member token count, and allocate this budget proportionally to training provenance for multi-subset domains.

\subsection{Construction of Members} \label{sec4.5}
Given the constructed non-members per domain, we construct members that are distributionally aligned with them, via (i) identifying candidate pools, (ii) determining the alignment axes, and (iii) constrained sampling using simulated annealing. We provide detailed descriptions in Appendix~\ref{appendix:member_construction}.

\textbf{Candidate Pools.} 
For domains without quality thresholding, candidate pools are the corresponding OLMo~2 training data. For quality-thresholded domains, to obtain candidates matched in quality to their non-member counterparts, we apply boundary sampling: since members are drawn from data that passed OLMo~2's quality threshold, this selects candidates immediately above the threshold.

\textbf{Alignment Axes.}
We identify three key axes for aligning members with non-members: (i) \textbf{text quality}, already enforced via boundary sampling for quality-thresholded domains; (ii) \textbf{temporal range}, enforced when timestamp data are available, since temporal drift can inflate detection scores~\citep{duan2024membership}; and (iii) \textbf{lexical similarity}, measured by Jensen--Shannon divergence (JSD) between their token-frequency distributions and enforced across all domains, as most MIAs target exact wording rather than meaning (Section~\ref{intro}), making lexical mismatch a direct confounder.

\begin{table*}[t]
\centering
\resizebox{1\textwidth}{!}{%
\begin{tabular}{@{}l@{\;\;}c@{\;\;}c@{\;\;}c@{\;\;}c@{\;\;}c@{\;\;}c@{\;\;}c@{\;\;}c@{\;\;}c@{\;\;}c@{}}
\toprule
& \multicolumn{4}{c}{\textbf{Pre-training}} & \multicolumn{2}{c}{\textbf{Mid-training}} & \multicolumn{3}{c}{\textbf{Post-training}} & \\
\cmidrule(lr){2-5} \cmidrule(lr){6-7} \cmidrule(lr){8-10}
& DCLM-Baseline & OpenWebMath & peS2o & StarCoder & Math Mix & Stack Exchange & SFT & DPO & RLVR & \textbf{Avg} \\
\midrule
\textbf{Overlap} & 4.07\% & 8.33\% & 3.11\% & 9.31\% & 6.65\% & 5.71\% & 6.32\% & 2.20\% & 2.06\% & \textbf{5.31\%} \\
\textbf{BoW AUC} & 0.49 & 0.52 & 0.48 & 0.50 & 0.52 & 0.48 & 0.51 & 0.49 & 0.51 & \textbf{0.50} \\
\bottomrule
\end{tabular}%
}
\caption{Per-domain verification results, averaged across model sizes. \textbf{Overlap}: average 13-gram overlap of each domain's non-members against OLMo~2's training corpus. \textbf{BoW AUC}: blind detectability, measured by the AUC of a BoW classifier. Per-model-size results are in Tables~\ref{table:13_gram_overlap_model_size} and~\ref{table:blind_auc_model_size}.}
\label{table:benchmark_verification}
\end{table*}


\textbf{Constrained Sampling via Simulated Annealing.} 
Given the candidates, we select members that match their non-member counterparts subject to two constraints: (i) alignment on axes beyond text quality; (ii) matched instance and token counts, eliminating average instance length as a confounder. Let $B$ and $A$ denote the sets of non-members and candidate members for a domain. We search for a member subset $A^* \subseteq A$ that minimizes a distributional loss $\mathcal{L}$ via simulated annealing \citep{lecchini2007simulated}, which probabilistically accepts suboptimal moves to escape local minima:
\begin{equation}
\begin{aligned}
A^* = \arg\min_{A' \subseteq A} \quad & \mathrm{JSD}(p_{A'}, p_B) + \mathds{1}[\text{temporal metadata available}] \cdot \mathcal{L}_{\text{temp}} \\
\text{s.t.} \quad & |A'| = |B|, \quad \left| \textstyle\sum_{x \in A'} \mathrm{tok}(x) - \sum_{x \in B} \mathrm{tok}(x) \right| \leq \delta,
\end{aligned}
\label{eq:2}
\end{equation}
where $p_{A'}$ and $p_B$ denote token distributions of the candidate member subset and non-members, respectively; $\mathrm{tok}(x)$ denotes the token count of instance $x$; $\delta$ is the token margin, which is set to 250; $\mathrm{JSD}$ enforces lexical alignment and is bounded in $[0,1]$; and $\mathcal{L}_{\text{temp}}$, which is normalized to the same scale as $\mathrm{JSD}$, aligns members and non-members' mean and standard deviation of timestamps.

\subsection{Benchmark Verification}

\textbf{Residual Overlap of Non-Members.}
Table~\ref{table:benchmark_verification} shows that non-members of \textsc{OLMo-Detect} exhibit an average 13-gram overlap of 5.31\% against OLMo~2's full training corpus, with a maximum of just 9.31\%, confirming the effectiveness of our infini-gram-based filtering scheme. 

\textbf{Verification of Lexical Alignment.}
Since lexical similarity has no ground truth and is measured using a summary statistic (JSD), residual lexical shifts may still remain. To rule this out, we follow \citet{das2025blind} and probe each domain with a \emph{blind} attack: a bag-of-words (BoW) classifier that predicts membership from text alone, without access to the LLM. If residual lexical confounders exist, the AUC of BoW will exceed chance. As shown in Table~\ref{table:benchmark_verification}, every domain attains a blind AUC within $0.02$ of $0.50$, confirming that members and non-members are lexically well aligned.

\subsection{OLMo-Detect (Shifted)} \label{sec4.7}
To examine the effect of distribution shifts on detection robustness, we introduce \textsc{OLMo-Detect (Shifted)}. For each domain, non-members remain unchanged, but members are randomly sampled to satisfy only the constraint of matched instance and token counts, without enforcing alignment on text quality, temporal range, or lexical similarity. Mid-training's Math Mix and post-training's RLVR are excluded from \textsc{OLMo-Detect (Shifted)}, as randomly sampled members do not yield substantially different $\mathrm{JSD}$ from their corresponding non-members. Overall statistics of \textsc{OLMo-Detect (Shifted)} are reported in Table~\ref{appendix:table:olmo_detect_shifted:overall_statistics}, with detailed descriptions in Appendix~\ref{appendix:olmo_detect_shifted}.

\begin{table*}[t]
\centering
\resizebox{1\textwidth}{!}{%
\begin{tabular}{@{}lc@{\;\;}c@{\;\;}c@{\;\;}cc@{\;\;}c@{\;\;}c@{\;\;}cc@{\;\;}c@{\;\;}c@{\;\;}cc@{\;\;}c@{\;\;}c@{\;\;}cc@{\;\;}c@{\;\;}c@{\;\;}c}
\toprule
\multirow{2}{*}{\textbf{Method}} & \multicolumn{4}{c}{\textbf{1B}} & \multicolumn{4}{c}{\textbf{7B}} & \multicolumn{4}{c}{\textbf{13B}} & \multicolumn{4}{c}{\textbf{32B}} & \multicolumn{4}{c}{\textbf{Model-Avg}} \\
\cmidrule(lr){2-5} \cmidrule(lr){6-9} \cmidrule(lr){10-13} \cmidrule(lr){14-17} \cmidrule(lr){18-21}
 & \textbf{Pre} & \textbf{Mid} & \textbf{Post} & \textbf{All} & \textbf{Pre} & \textbf{Mid} & \textbf{Post} & \textbf{All} & \textbf{Pre} & \textbf{Mid} & \textbf{Post} & \textbf{All} & \textbf{Pre} & \textbf{Mid} & \textbf{Post} & \textbf{All} & \textbf{Pre} & \textbf{Mid} & \textbf{Post} & \textbf{All} \\
\midrule
PPL$^{\mathrm{L}}$ & 0.54 & 0.64 & 0.61 & 0.60 & 0.55 & 0.72 & 0.68 & 0.65 & 0.56 & 0.79 & 0.66 & 0.67 & 0.58 & 0.80 & 0.68 & 0.69 & 0.56 & 0.74 & 0.66 & 0.65 \\
Zlib$^{\mathrm{L}}$ & 0.57 & \textbf{0.67} & \textbf{0.66} & \textbf{0.63} & 0.59 & \textbf{0.76} & 0.66 & 0.67 & 0.59 & \textbf{0.83} & \textbf{0.67} & \textbf{0.70} & 0.61 & \textbf{0.85} & \textbf{0.69} & \textbf{0.72} & 0.59 & \textbf{0.78} & \textbf{0.67} & \textbf{0.68} \\
Lower$^{\mathrm{L}}$ & 0.57 & 0.61 & 0.59 & 0.59 & 0.57 & 0.69 & 0.62 & 0.63 & 0.58 & 0.75 & 0.65 & 0.66 & 0.58 & 0.73 & 0.67 & 0.66 & 0.58 & 0.69 & 0.63 & 0.63 \\
MinK$^{\mathrm{L}}$ & 0.55 & 0.65 & 0.62 & 0.61 & 0.56 & 0.73 & 0.67 & 0.65 & 0.57 & 0.80 & 0.66 & 0.68 & 0.59 & 0.81 & 0.69 & 0.70 & 0.57 & 0.75 & 0.66 & 0.66 \\
MinK++$^{\mathrm{L}}$ & 0.48 & 0.53 & 0.59 & 0.54 & 0.50 & 0.54 & 0.62 & 0.56 & 0.51 & 0.59 & 0.59 & 0.57 & 0.54 & 0.56 & 0.63 & 0.57 & 0.51 & 0.56 & 0.61 & 0.56 \\
DCPDD$^{\mathrm{L}}$ & \textbf{0.59} & 0.61 & 0.59 & 0.60 & \textbf{0.60} & 0.76 & 0.66 & \textbf{0.67} & 0.61 & 0.83 & 0.64 & 0.69 & \textbf{0.63} & 0.83 & 0.67 & 0.71 & \textbf{0.61} & 0.76 & 0.64 & 0.67 \\
PAC$^{\mathrm{L}}$ & 0.58 & 0.65 & 0.61 & 0.61 & 0.59 & 0.73 & 0.60 & 0.64 & \textbf{0.62} & 0.81 & 0.61 & 0.68 & 0.63 & 0.80 & 0.63 & 0.69 & 0.60 & 0.75 & 0.61 & 0.66 \\
ReCaLL$^{\mathrm{L}}$ & 0.50 & 0.43 & 0.50 & 0.48 & 0.54 & 0.29 & 0.49 & 0.44 & 0.55 & 0.34 & 0.48 & 0.45 & 0.52 & 0.37 & 0.54 & 0.48 & 0.53 & 0.36 & 0.50 & 0.46 \\
EMMIA$^{\mathrm{L}}$ & 0.56 & 0.53 & 0.61 & 0.57 & 0.58 & 0.54 & 0.61 & 0.57 & 0.57 & 0.56 & 0.60 & 0.58 & 0.61 & 0.59 & 0.47 & 0.56 & 0.58 & 0.56 & 0.57 & 0.57 \\
CAMIA$^{\mathrm{L}}$ & 0.55 & 0.67 & 0.59 & 0.60 & 0.53 & 0.73 & \textbf{0.72} & 0.66 & 0.54 & 0.80 & 0.66 & 0.67 & 0.54 & 0.82 & 0.66 & 0.67 & 0.54 & 0.76 & 0.66 & 0.65 \\
\cmidrule(lr){1-21}
Neigh.$^{\mathrm{P}}$ & 0.55 & 0.59 & 0.53 & 0.56 & 0.54 & 0.64 & 0.55 & 0.58 & 0.55 & 0.73 & 0.53 & 0.60 & 0.55 & 0.76 & 0.57 & 0.63 & 0.55 & 0.68 & 0.55 & 0.59 \\
\cmidrule(lr){1-21}
Guided$^{\mathrm{Pr}}$ & 0.47 & 0.52 & 0.50 & 0.50 & 0.50 & 0.47 & 0.50 & 0.49 & 0.47 & 0.48 & 0.49 & 0.48 & 0.48 & 0.55 & 0.46 & 0.50 & 0.48 & 0.51 & 0.49 & 0.49 \\
DCQ$^{\mathrm{Pr}}$ & 0.52 & 0.50 & 0.46 & 0.50 & 0.41 & 0.45 & 0.40 & 0.42 & 0.49 & 0.59 & 0.48 & 0.52 & 0.56 & 0.58 & 0.47 & 0.54 & 0.50 & 0.53 & 0.45 & 0.49 \\
\cmidrule(lr){1-21}
CDD$^{\mathrm{O}}$ & 0.51 & 0.51 & 0.51 & 0.51 & 0.50 & 0.54 & 0.50 & 0.52 & 0.51 & 0.60 & 0.50 & 0.54 & 0.50 & 0.62 & 0.49 & 0.54 & 0.51 & 0.57 & 0.50 & 0.52 \\
SelfCrit$^{\mathrm{O}\dagger}$ & --- & --- & 0.46 & 0.46 & --- & --- & 0.50 & 0.50 & --- & --- & 0.44 & 0.44 & --- & --- & 0.48 & 0.48 & --- & --- & 0.47 & 0.47 \\
\midrule
\textbf{Method-Avg} & 0.54 & 0.58 & 0.56 & 0.56 & 0.54 & 0.61 & 0.59 & 0.58 & 0.55 & 0.68 & 0.58 & 0.60 & 0.57 & 0.69 & 0.59 & 0.61 & 0.55 & 0.64 & 0.58 & 0.58 \\
\bottomrule
\end{tabular}%
}
\caption{AUCs of unsupervised MIAs on \textsc{OLMo-Detect}, grouped by family: $^{\mathrm{L}}$ likelihood-based, $^{\mathrm{P}}$ perturbation-based, $^{\mathrm{Pr}}$ prompt-based, and $^{\mathrm{O}}$ output-distribution-based. \textbf{Pre}/\textbf{Mid}/\textbf{Post} macro-average AUC over each stage's domains, and \textbf{All} macro-averages all stages. \textbf{Model-Avg} averages across model sizes, and \textbf{Method-Avg} averages across methods. The best method per column is in bold. $^\dagger$SelfCrit is evaluated on post-training's DPO and RLVR. Per-domain results are in Appendix~\ref{appendix:auc_results_unsup}.}
\label{table:olmo_detect_overall_unsupervised}
\end{table*}

\section{Experiments} \label{exp:setting}

\subsection{Experimental Settings}
We describe our experimental settings below, with details in Appendix~\ref{appendix:experiment_setting}. (i) \textbf{Models.} We evaluate four instruction-tuned OLMo~2 models: 1B, 7B, 13B, and 32B. (ii) \textbf{Compared Methods.} We evaluate 15 unsupervised MIAs across four families: (a) likelihood-based: Perplexity (PPL)~\citep{yeom2018privacy}, Zlib and Lowercase (Lower)~\citep{carlini2021extracting}, Min-K\% (MinK)~\citep{shi2023detecting}, Min-K\%++ (MinK++)~\citep{zhang2024min}, DCPDD~\citep{zhang2024pretraining}, PAC~\citep{ye2024data}, ReCaLL~\citep{xie2024recall}, EMMIA~\citep{kim2026detecting}, and CAMIA~\citep{chang2025context}; (b) perturbation-based: Neighborhood Attack (Neigh.)~\citep{mattern2023membership}; (c) prompt-based: Guided Instruction (Guided)~\citep{golchin2023time} and DCQ~\citep{golchin2025data}; and (d) output-distribution-based: CDD~\citep{dong2024generalization} and Self-Critique (SelfCrit)~\citep{tao2025detecting} (evaluated only on DPO and RLVR, following its original setup). We also evaluate three supervised MIAs: the supervised variant of CAMIA (Sup-CAMIA) \citep{chang2025context}, MIA-Tuner \citep{fu2025mia}, and FSD \citep{zhang2025fine}. (iii) \textbf{Evaluation Metrics.} Following prior work~\citep{shi2023detecting}, we report AUC and TPR@5\%FPR. (iv) \textbf{Score Aggregation.} To aggregate subset-level scores into domain-level scores for multi-subset domains, domain-level scores into stage-level scores, and stage-level scores into final scores, we use macro-averaging, following~\citet{duan2024membership}.

\subsection{Overall Performance}

\textbf{Unsupervised Methods.} AUCs of unsupervised MIAs on \textsc{OLMo-Detect} are reported in Table~\ref{table:olmo_detect_overall_unsupervised}, with TPR@5\%FPR results in Appendix~\ref{appendix:tpr_results_unsup}. Our main findings are: (i) \textbf{Overall performance is limited.} The best method, Zlib, reaches an overall AUC of only 0.68, while the average across methods is 0.58. This suggests that, under distribution-matched settings with minimal confounders, membership signals are substantially harder to detect than existing benchmarks suggest. Notably, Zlib, one of the earliest and simplest MIAs, outperforms every more sophisticated method. (ii) \textbf{Likelihood-based methods outperform other families.} Across families, likelihood-based MIAs generally perform best, with one exception: ReCaLL. ReCaLL assumes that a non-member prefix disrupts predictions on members more than on non-members, but this effect is inverted on  GSM8K, CoT-GSM8K, and CoT-MATH (AUC 0.10, 0.09, and 0.45; Tables~\ref{table:auc_gsm8k} and~\ref{table:auc_rlvr_subsets}): their rigid format turns the prefix into an in-context demonstration, improving member predictions instead of disrupting them. (iii) \textbf{Token-level normalization hurts performance.} MinK++, which applies token-level vocabulary normalization to MinK, performs markedly worse (0.56 vs.\ 0.66), suggesting that fine-grained normalization removes part of the membership signal under distribution-matched settings.


\begin{table*}[t]
\centering
\resizebox{1\textwidth}{!}{%
\begin{tabular}{@{}llc@{\;\;}c@{\;\;}c@{\;\;}cc@{\;\;}c@{\;\;}c@{\;\;}cc@{\;\;}c@{\;\;}c@{\;\;}cc@{\;\;}c@{\;\;}c@{\;\;}cc@{\;\;}c@{\;\;}c@{\;\;}c@{}}
\toprule
\multirow{2.5}{*}{\textbf{Method}} & \multirow{2.5}{*}{\textbf{Setting}} & \multicolumn{4}{c}{\textbf{1B}} & \multicolumn{4}{c}{\textbf{7B}} & \multicolumn{4}{c}{\textbf{13B}} & \multicolumn{4}{c}{\textbf{32B}} & \multicolumn{4}{c}{\textbf{Model-Avg}} \\
\cmidrule(lr){3-6} \cmidrule(lr){7-10} \cmidrule(lr){11-14} \cmidrule(lr){15-18} \cmidrule(lr){19-22}
 & & Pre & Mid & Post & All & Pre & Mid & Post & All & Pre & Mid & Post & All & Pre & Mid & Post & All & Pre & Mid & Post & All \\
\midrule
\multirow{2}{*}{\textbf{Sup-CAMIA}} & In-domain & \textbf{0.56} & \textbf{0.72} & \textbf{0.61} & \textbf{0.63} & 0.54 & \textbf{0.78} & \textbf{0.72} & \textbf{0.68} & 0.55 & \textbf{0.86} & \textbf{0.72} & \textbf{0.71} & 0.55 & \textbf{0.84} & \textbf{0.67} & \textbf{0.69} & 0.55 & \textbf{0.80} & \textbf{0.68} & \textbf{0.68} \\
 & LODO & 0.56 & 0.63 & 0.52 & 0.57 & \textbf{0.60} & 0.73 & 0.62 & 0.65 & \textbf{0.57} & 0.66 & 0.55 & 0.59 & \textbf{0.59} & 0.70 & 0.62 & 0.64 & \textbf{0.58} & 0.68 & 0.58 & 0.61 \\
\midrule
\multirow{2}{*}{\textbf{MIA-Tuner}} & In-domain & 0.54 & 0.60 & 0.53 & 0.56 & 0.53 & 0.61 & 0.55 & 0.56 & 0.54 & 0.60 & 0.53 & 0.56 & 0.54 & 0.53 & 0.51 & 0.52 & 0.54 & 0.59 & 0.53 & 0.55 \\
 & LODO & 0.52 & 0.50 & 0.47 & 0.50 & 0.53 & 0.48 & 0.46 & 0.49 & 0.47 & 0.48 & 0.46 & 0.47 & 0.51 & 0.53 & 0.50 & 0.52 & 0.51 & 0.50 & 0.47 & 0.49 \\
\bottomrule
\end{tabular}%
}

\vspace{4pt}

\resizebox{1\textwidth}{!}{
\begin{tabular}{@{}lcc@{\;\;}c@{\;\;}c@{\;\;}c@{\;\;}cc@{\;\;}c@{\;\;}c@{\;\;}c@{\;\;}c@{}}
\toprule
\multirow{2.5}{*}{\textbf{Method}} & \multirow{2.5}{*}{\textbf{Base AUC}} & \multicolumn{5}{c}{\textbf{In-domain $\Delta$AUC}} & \multicolumn{5}{c}{\textbf{LODO $\Delta$AUC}} \\
\cmidrule(lr){3-7} \cmidrule(lr){8-12}
 & & 1B & 7B & 13B & 32B & \textbf{Model-Avg} & 1B & 7B & 13B & 32B & \textbf{Model-Avg} \\
\midrule
\textbf{FSD (PPL)} & 0.65 & $+$0.00 & $-$0.03 & \textbf{$+$0.01} & $-$0.03 & $-$0.01 & $-$0.04 & $-$0.02 & $-$0.02 & $-$0.05 & $-$0.03 \\
\textbf{FSD (Zlib)} & 0.68 & $-$0.04 & $-$0.04 & $-$0.02 & $-$0.07 & $-$0.04 & $-$0.07 & $-$0.03 & $-$0.04 & $-$0.07 & $-$0.06 \\
\textbf{FSD (Lower)} & 0.63 & $-$0.03 & $-$0.02 & $-$0.03 & $-$0.05 & $-$0.03 & $-$0.04 & $-$0.02 & $-$0.02 & $-$0.05 & $-$0.03 \\
\textbf{FSD (MinK)} & 0.66 & $-$0.05 & $-$0.12 & $-$0.07 & $-$0.09 & $-$0.08 & $-$0.06 & $-$0.03 & $-$0.03 & $-$0.06 & $-$0.05 \\
\bottomrule
\end{tabular}%
}
\caption{Supervised MIAs on \textsc{OLMo-Detect} under \textbf{In-domain} and \textbf{LODO} settings. \textbf{Top:} AUCs of Sup-CAMIA and MIA-Tuner; the best method per column is in bold. \textbf{Bottom:} FSD measures the change in a base MIA's score after fine-tuning the target model on non-members; we therefore report its AUC gain ($\Delta$AUC) over the base MIA (in parentheses), averaged across stages. \textbf{Base AUC}: base MIA's AUC averaged across model sizes; positive gains are bolded. Details are in Appendix~\ref{appendix:auc_results_sup}.}
\label{table:olmo_detect_overall_supervised}
\end{table*}



\textbf{Supervised Methods.} \label{sec:sup_results}
We evaluate supervised MIAs in two settings: (i) \textbf{In-domain}, trained on the evaluated domain's dev set, which measures the gain from same-distribution supervision over unsupervised MIAs; and (ii) \textbf{LODO} (leave-one-domain-out), trained on the other eight domains' dev sets and evaluated on the held-out domain, which measures cross-domain transfer. AUCs of supervised MIAs on \textsc{OLMo-Detect} are reported in Table~\ref{table:olmo_detect_overall_supervised}, with TPR@5\%FPR results in Appendix~\ref{appendix:tpr_results_sup}. Our main findings are: (i) \textbf{In-domain supervision adds little over unsupervised methods.} Under the \textbf{In-domain} setting, the best supervised method, Sup-CAMIA, reaches an overall AUC of 0.68, matching the best unsupervised MIA, Zlib (0.68), and exceeding its unsupervised counterpart CAMIA (0.65) by only 0.03; MIA-Tuner remains near chance (0.55), and FSD lowers the AUC of every base MIA on average. This indicates that matched-distribution labels add little beyond unsupervised membership signals.  To rule out the small dev set as a confounder, we hold out 50\% of each domain for evaluation and vary the supervision budget from 10\% to 50\% (Appendix~\ref{appendix:learning_curve}, Table~\ref{table:lc_supcamia}). We find that increasing the budget improves Sup-CAMIA by only 0.04--0.06 AUC, confirming that the limited performance is not an artifact of the labeled-data budget. 
(ii) \textbf{Learned signals do not transfer well across domains.} Compared with \textbf{In-domain}, all supervised MIAs perform worse or no better under \textbf{LODO}: Sup-CAMIA drops to 0.61 ($-0.07$), MIA-Tuner to 0.49 ($-0.06$), and FSD's average gain over its base MIAs remains $-0.04$. This suggests that supervised MIAs largely learn domain-specific statistics rather than transferrable membership signals.

\textbf{Domain-Unknown Setting.} To simulate a real-world setting in which an auditor is unaware of a text's domain, we compute a single AUC over members and non-members pooled with equal instance counts across all domains, retraining supervised MIAs on the pooled dev sets. A full discussion is provided in Appendix~\ref{appendix:three_settings}. In summary, unsupervised and supervised MIAs degrade slightly, by 0.02 and 0.03 AUC respectively, with the supervised drop concentrated in Sup-CAMIA ($-0.06$). The relative ranking is largely preserved (Spearman $\rho = 0.95$ over all 17 MIAs in Table~\ref{table:three_settings}).


\subsection{Performance Across Stages and Model Sizes}
\textbf{Cross-Stage Performance.} As shown in the \textbf{Model-Avg} columns of Tables~\ref{table:olmo_detect_overall_unsupervised} and~\ref{table:olmo_detect_overall_supervised}, overall MIA performance peaks at mid-training and is considerably lower at post-training and pre-training. However, this pattern reflects data type rather than a stage effect: curated math data, i.e., instances drawn from purpose-built math datasets rather than harvested from the web, is far more detectable than any other data type. As the top half of Table~\ref{table:cross_stage} shows, the three curated math subsets, GSM8K, CoT-GSM8K, and CoT-MATH, achieve the highest AUCs across all MIAs, and their advantage is even more pronounced among effective MIAs (overall AUC $\geq 0.60$), where all three exceed 0.70 and CoT-GSM8K reaches 0.90. Notably, the effect is specific to curated math data rather than to mathematical content in general: Math Stack Exchange (free-form forum math Q\&A) and OpenWebMath (web math text) score only 0.61 and 0.57 among effective MIAs, in line with non-math subsets. To test this difference statistically, we conduct a one-sided Wilcoxon signed-rank test with $\alpha = 0.01$ on the paired per-MIA AUC differences between the curated math subsets and the others. As the bottom half of Table~\ref{table:cross_stage} shows, curated math subsets are significantly more detectable under both the all-MIA ($p=0.008$) and the effective-MIA ($p=0.004$) setting, and once they are excluded, the spread across stages shrinks substantially ($0.10\to0.05$ in the all-MIA and $0.17\to0.06$ in the effective-MIA setting), indicating that stage ordering is largely a composition effect. Finally, we examine duplication count, i.e., how many times each member is seen during training, as a potential confounder, and find that it does not explain the observed pattern: the most duplicated domain, DCLM, is less detectable than CoT-MATH, which has a duplication count of only 1.5.

\textbf{Cross-Model-Size Performance.} The \textbf{Method-Avg} row of Table~\ref{table:olmo_detect_overall_unsupervised} shows that the average AUC of unsupervised MIAs increases with model size, from 0.56 at 1B to 0.58 at 7B and 0.60 at 13B, but reaches only 0.61 at 32B. This trend is more pronounced for effective unsupervised MIAs (overall AUC $\geq 0.60$): their average AUC rises from 0.61 at 1B to 0.68 at 13B, but gains only 0.01 more at 32B. Supervised methods (Table~\ref{table:olmo_detect_overall_supervised}) even decline at 32B under the In-domain setting: Sup-CAMIA rises from 0.63 at 1B to 0.71 at 13B before falling to 0.69 at 32B, while MIA-Tuner remains at 0.56 through 13B and drops to 0.52 at 32B. The increase from 1B to 13B is consistent with MIMIR~\citep{duan2024membership}, which attributed a similar trend in Pythia models to greater overfitting. Beyond 13B, gains diminish and In-domain supervised performance declines, echoing \citet{hayes2026exploringlimitsstrongmembership}, who report that MIA AUC does not increase monotonically with model size.

\begin{table*}[t]
\centering
\resizebox{1\textwidth}{!}{%
\begin{tabular}{@{}lc@{\;\;}c@{\;\;}c@{\;\;}c@{\;\;}cc@{\;\;}c@{\;\;}cc@{\;\;}c@{\;\;}c@{\;\;}c@{\;\;}c@{}}
\toprule
 & \multicolumn{5}{c}{\textbf{Pre-training}} & \multicolumn{3}{c}{\textbf{Mid-training}} & \multicolumn{5}{c}{\textbf{Post-training}} \\
\cmidrule(lr){2-6} \cmidrule(lr){7-9} \cmidrule(lr){10-14}
 & \multirow{2.5}{*}{DCLM} & \multirow{2.5}{*}{peS2o} & \multirow{2.5}{*}{OWM} & \multicolumn{2}{c}{StarCoder} & Math Mix & \multicolumn{2}{c}{SE} & \multicolumn{2}{c}{SFT} & DPO & \multicolumn{2}{c}{RLVR} \\
\cmidrule(lr){5-6} \cmidrule(lr){7-7} \cmidrule(lr){8-9} \cmidrule(lr){10-11} \cmidrule(lr){12-12} \cmidrule(lr){13-14}
 & & & & Python & Java & GSM8K & Math SE & SO & Aya & WildChat & UF-WildChat & CoT-GSM8K & CoT-MATH \\
\midrule
\textbf{AUC} (all MIAs) & 0.56 & 0.52 & 0.54 & 0.56 & 0.59 & \textbf{0.71} & 0.59 & 0.60 & 0.57 & 0.60 & 0.50 & \textbf{0.70} & 0.63 \\
\textbf{AUC} (eff.\ MIAs) & 0.62 & 0.50 & 0.57 & 0.59 & 0.64 & \textbf{0.88} & 0.61 & 0.61 & 0.59 & 0.62 & 0.50 & \textbf{0.90} & \textbf{0.74} \\
\textit{Dup.\ count} & 9.00 & 1.12 & 1.00 & 1.00 & 1.00 & 6.00 & 6.00 & 6.00 & 2.09 & 2.00 & 1.13 & 1.25 & 1.50 \\
\bottomrule
\end{tabular}%
}
\vspace{4pt}

\resizebox{1\textwidth}{!}{%
\begin{tabular}{@{}lc@{\;\;}c@{\;\;}cc@{\;\;}c@{\;\;}cc@{\;\;}c@{\;\;}c@{}}
\toprule
 & \multicolumn{3}{c}{\textbf{Stage (all subsets)}} & \multicolumn{3}{c}{\textbf{Stage (w/o curated math)}} & \multicolumn{3}{c}{\textbf{Curated math vs.\ others}} \\
\cmidrule(lr){2-4} \cmidrule(lr){5-7} \cmidrule(lr){8-10}
 & Pre & Mid & Post & Pre & Mid & Post & Curated math & Others & $\Delta$ (\#MIAs, $p$) \\
\midrule
\textbf{AUC} (all MIAs) & 0.55 & 0.65 & 0.59 & 0.55 & 0.59 & 0.54 & 0.69 & 0.56 & $+0.12$ (13/16, 0.008) \\
\textbf{AUC} (eff.\ MIAs) & 0.58 & 0.75 & 0.65 & 0.57 & 0.61 & 0.55 & 0.86 & 0.59 & $+0.27$ (8/8, 0.004) \\
\bottomrule
\end{tabular}%
}
\caption{Cross-stage analysis on \textsc{OLMo-Detect}. \textbf{Top:} Per-subset AUCs averaged across model sizes and over all MIAs (excluding FSD, which reports gain, and SelfCrit, which is evaluated on DPO/RLVR) or \textit{effective (eff.)} MIAs (overall AUC $\geq 0.60$); bold means AUC $>0.70$. \textit{Dup.\ count}: training-corpus occurrences per member in each subset. DCLM: DCLM-Baseline; OWM: OpenWebMath; SE: Stack Exchange; SO: Stack Overflow. \textbf{Bottom:} Stage-level AUCs with and without the three curated math subsets (GSM8K, CoT-GSM8K, CoT-MATH), and AUCs on the curated math subsets versus the others. $\Delta$: average per-MIA AUC gain on curated math subsets over the others; parentheses report the number of MIAs with positive gains and the one-sided Wilcoxon $p$-value.}
\label{table:cross_stage}
\end{table*}

\subsection{Robustness to Distribution Shifts} \label{sec:5.4}
To test the robustness of unsupervised MIAs to distribution shifts, we compare their per-domain AUCs on \textsc{OLMo-Detect (Shifted)} and \textsc{OLMo-Detect}, testing each difference using a paired instance-level bootstrap (Table~\ref{table:distribution_shifts}). Our main findings are: (i) \textbf{No unsupervised MIA is robust.} Every MIA's AUC changes significantly on three to six of the seven domains, with each MIA's largest change ranging from 0.09 to 0.42 (mean 0.22); the largest changes concentrate in pre-training domains. (ii) \textbf{Whether distribution shifts inflate or degrade AUC follows no consistent pattern.} For all methods except ReCaLL and CDD, whose significant AUC changes are all inflations, there is no consistent shifting pattern across method--domain pairs: OpenWebMath inflates PPL's AUC by 0.29 but degrades Zlib's by 0.31; peS2o inflates CAMIA's AUC by 0.28 yet degrades EMMIA's by 0.17 (Table~\ref{table:distribution_shifts_per_domain}). Thus, shifted benchmarks distort AUC unpredictably. (iii) \textbf{Likelihood-based MIAs are most affected.} Eight of the ten likelihood-based MIAs change by 0.20--0.42 AUC, while the methods from other families (Neigh., Guided, DCQ, and CDD) change by at most 0.12.

\begin{table*}[t]
\centering
\resizebox{1\textwidth}{!}{%
\begin{tabular}{@{}lc@{\;\;}c@{\;\;}c@{\;\;}c@{\;\;}c@{\;\;}c@{\;\;}c@{\;\;}c@{\;\;}c@{\;\;}c@{\;\;}c@{\;\;}c@{\;\;}c@{\;\;}c@{}}
\toprule
 & \textbf{PPL} & \textbf{Zlib} & \textbf{Lower} & \textbf{MinK} & \textbf{MinK++} & \textbf{DCPDD} & \textbf{ReCaLL} & \textbf{CAMIA} & \textbf{PAC} & \textbf{Neigh.} & \textbf{EMMIA} & \textbf{DCQ} & \textbf{CDD} & \textbf{Guided} \\
\midrule
\textbf{I / D} & 4 / 2 & 2 / 2 & 5 / 1 & 2 / 1 & 1 / 3 & 2 / 3 & 4 / 0 & 4 / 1 & 5 / 1 & 5 / 1 & 3 / 2 & 2 / 1 & 3 / 0 & 1 / 2 \\
\textbf{Largest $\Delta$} & $+$0.29 & $-$0.31 & $-$0.30 & $+$0.23 & $-$0.16 & $+$0.16 & $+$0.24 & $+$0.28 & $-$0.20 & $+$0.12 & $+$0.42 & $+$0.09 & $+$0.12 & $+$0.10 \\
 & (OWM) & (OWM) & (OWM) & (OWM) & (SC) & (SC) & (peS2o) & (peS2o) & (OWM) & (SC) & (OWM) & (peS2o) & (OWM) & (SC) \\
\bottomrule
\end{tabular}%
}
\caption{Effect of distribution shift on unsupervised MIAs (excluding SelfCrit), over the seven domains with shifted variants. \textbf{I / D}: number of domains on which the AUC increases / decreases significantly under the shift ($\Delta$ = \textsc{OLMo-Detect (Shifted)} $-$ \textsc{OLMo-Detect}, averaged over model sizes; 95\% paired instance bootstrap). \textbf{Largest $\Delta$}: the largest change and the domain where it occurs. OWM: OpenWebMath; SC: StarCoder. Per-domain values are in Appendix~\ref{appendix:distribution_shifts}.}
\label{table:distribution_shifts}
\end{table*}

\begin{table}[t]
\centering
\resizebox{1\columnwidth}{!}{%
\begin{tabular}{@{}llc@{\;\;}c@{\;\;}c@{\;\;}c@{\;\;}cc@{}}
\toprule
\textbf{Domain} & \textbf{Comparison} & \textbf{PPL} & \textbf{Zlib} & \textbf{MinK} & \textbf{DCPDD} & \textbf{PAC} & \textbf{Avg} \\
\midrule
\multirow{2}{*}{GSM8K} & OLMo~2-7B / OLMo~3-7B & 0.85 / 0.97 & 0.89 / 0.98 & 0.86 / 0.97 & 0.91 / 0.98 & 0.83 / 0.95 & 0.87 / 0.97 \\
 & OLMo~2-32B / OLMo~3-32B & 0.98 / 0.99 & 0.99 / 1.00 & 0.97 / 0.99 & 0.98 / 0.98 & 0.92 / 0.98 & 0.97 / 0.99 \\
\midrule
\multirow{2}{*}{DCLM-Baseline} & OLMo~2-1B / DCB-1B & 0.60 / 0.63 & 0.59 / 0.61 & 0.57 / 0.60 & 0.57 / 0.60 & 0.61 / 0.63 & 0.59 / 0.61 \\
 & OLMo~2-1B / SmolLM2-1.7B & 0.60 / 0.62 & 0.59 / 0.62 & 0.57 / 0.59 & 0.57 / 0.61 & 0.61 / 0.64 & 0.59 / 0.62 \\
\midrule
\multirow{2}{*}{OpenWebMath} & OLMo~2-1B / DCB-1B & 0.51 / 0.51 & 0.56 / 0.54 & 0.55 / 0.54 & 0.61 / 0.58 & 0.54 / 0.56 & 0.55 / 0.54 \\
 & OLMo~2-7B / DCB-7B & 0.51 / 0.53 & 0.57 / 0.56 & 0.55 / 0.55 & 0.57 / 0.59 & 0.56 / 0.57 & 0.55 / 0.56 \\
\bottomrule
\end{tabular}%
}
\caption{AUCs of the five best-performing unsupervised MIAs on OLMo~2 and other model families.}
\label{table:generalization}
\end{table}

\subsection{Generalization Beyond OLMo 2}
\label{generalization}

To show that our findings are not artifacts of OLMo~2, we evaluate the five best-performing unsupervised MIAs (Zlib, DCPDD, MinK, PAC, and PPL) on OLMo~3 \citep{olmo2025olmo} and two non-OLMo model families, comparing them with size-matched OLMo~2 models. (i) \textbf{OLMo~3.} The 7B and 32B instruction-following OLMo~3 models share the GSM8K subset of \textsc{OLMo-Detect}'s Math Mix domain with OLMo~2, which occurs at mid-training in both families. We further verify that \textsc{OLMo-Detect}'s GSM8K members are included in OLMo~3's training data while its non-members are excluded. As shown in Table~\ref{table:generalization}, GSM8K remains highly detectable on OLMo~3 (AUC 0.95--1.00), with a method ranking similar to that on OLMo~2, PAC being the weakest in both families. While both families approach an AUC of 1.0 at 32B, OLMo~3 is more detectable at 7B, likely due to OLMo~3's stronger memorization capability. (ii) \textbf{Non-OLMo Models.} Eligible non-OLMo models must share training data with an \textsc{OLMo-Detect} domain, whose non-members must either be excluded from the model's training data by construction based on clear documentation or be verifiable as absent, which requires open access to its training corpus. To the best of our knowledge, only two model families satisfy these strict conditions: DCLM-Baseline-1B/7B (DCB-1B/7B)~\citep{li2024datacomp} and SmolLM2-1.7B~\citep{allal2025smollm2smolgoesbig}. DCB-1B is trained on the full DCLM-Baseline corpus and OpenWebMath, covering two \textsc{OLMo-Detect} domains, whereas DCB-7B is trained on an undisclosed subset of DCLM-Baseline, leaving only OpenWebMath valid for evaluation; SmolLM2-1.7B is trained on the full DCLM-Baseline corpus, covering the DCLM domain. Conveniently, the entire training data of DCB-1B/7B is a subset of OLMo~2's pre-training data, so the non-members of \textsc{OLMo-Detect}'s DCLM and OpenWebMath domains are clean for these models by construction; for SmolLM2, we apply the same 13-gram filter used in benchmark construction to each of its remaining training corpora, all of which are publicly released, and score both models on the non-members that are verified to be clean (Appendix~\ref{appendix:smollm2_data}). As shown in Table~\ref{table:generalization}, all three models closely track their size-matched OLMo~2 counterparts: every method is within 0.04 AUC, and both domains remain weakly detectable (AUC 0.51--0.64). Method rankings also transfer for all three models, with PAC performing best on DCLM and DCPDD best on OpenWebMath.



\section{Conclusion}

We introduced \textsc{OLMo-Detect}, the first benchmark for membership inference on LLMs that spans pre-training, mid-training, and post-training, explicitly aligns members and non-members on three key axes, and rigorously filters non-members against the training corpus. After extensive evaluation, we find that existing MIAs remain limited, with the best MIA reaching an AUC of only 0.68. Moreover, curated math data is substantially more detectable than any other data type. Importantly, the main patterns generalize to OLMo~3 and non-OLMo models, showing that our findings are not artifacts of OLMo~2. We hope \textsc{OLMo-Detect} serves as a rigorous testbed for future MIAs.

\subsection*{AI use statement}

In this work, we used generative AI tools for the following purposes:

\begin{enumerate}
    \item \textbf{To aid or polish writing.} We used generative AI tools to refine wording, grammar, and clarity and to check the consistency of terminology and citation formatting. All AI-suggested edits were reviewed by the authors, who take full responsibility for the final text.

    \item \textbf{For research ideation or execution.} We used generative AI tools to refine and double-check our implementations of the 18 compared MIA methods. All AI-refined components were verified and tested for correctness by the authors before use in our experiments. Additionally, we used generative AI tools to provide feedback on the benchmark construction methodology and experiments. All feedback was critically evaluated by the authors, who made all final methodological and experimental decisions.

    \item \textbf{To draft sections of the paper.} We used generative AI tools to draft Appendix~\ref{appendix:olmo2_training_data} (Detailed Descriptions of OLMo~2 Training Data), which catalogs extensive factual details about OLMo~2's publicly documented training pipeline. The authors checked all factual claims in the draft against OLMo~2's released documentation and wrote the final version themselves based on the draft.
\end{enumerate}

We have not used generative AI tools to generate synthetic datasets, help develop theoretical models or conceptual frameworks, propose or refine hypotheses, clean and reformat datasets, support qualitative and thematic data analysis, or interpret results. Formulating mathematical claims, providing critical ingredients for proving mathematical claims, assisting in the writing of proofs, and assisting with translation are not applicable to this work.

We take responsibility for the final content of this work, including text, claims, and artifacts produced with the aid of generative AI.

\subsection*{Ethics Statement}

\textsc{OLMo-Detect} is constructed entirely from data that OLMo~2 was trained on (members) or data drawn from the same source distributions (non-members). We do not scrape or introduce data from sources beyond those in OLMo~2's publicly documented training pipeline. We do not perform additional PII screening ourselves, but two domains inherit filtering from their sources~\citep{olmo20242}: peS2o applies regex-based PII detection and removes documents containing substantial PII~\citep{soldaini2024dolma}; SFT includes WildChat, which anonymizes names, contact information, and other identifiers through a dedicated pipeline before release~\citep{zhao2024wildchat}.

\subsection*{Reproducibility statement}

\textbf{Code and Data Release.}
We release our benchmark and full evaluation code at a GitHub repository: \url{https://github.com/LuckyDaydreamer/OLMo-Detect}.

\textbf{Benchmark Construction.} 
Benchmark construction is described in Sections~\ref{sec4.3}--\ref{sec4.7}, with detailed constructions of non-members and members provided in Appendices~\ref{appendix:non_member_construction} and~\ref{appendix:member_construction}, respectively. Overall statistics for \textsc{OLMo-Detect} are reported in Table~\ref{table:2}, with detailed per-domain statistics provided in Appendix~\ref{appendix:full_benchmark_statistics}. Statistics for \textsc{OLMo-Detect (Shifted)} are provided in Appendix~\ref{appendix:olmo_detect_shifted}.

\textbf{Method Implementation and Evaluation.}  
The overall experimental settings, including evaluated models, compared methods, and evaluation metrics, are described in Section~\ref{exp:setting}. Detailed implementation information is provided in Appendix~\ref{appendix:experiment_setting}, including (i) each method's original public repository, tuned hyperparameters, and search spaces; (ii) determinism and statistical reporting; (iii) continuous scoring for categorical methods; and (iv) reasons for excluding certain existing MIAs.

\bibliography{iclr2027_conference}
\bibliographystyle{iclr2027_conference}

\appendix

\section{Detailed Descriptions of Existing Benchmarks for Membership Inference on LLMs}
\label{appendix:related_work_benchmarks}

This section provides detailed descriptions of each benchmark surveyed in Section~\ref{related_work}.

\paragraph{Temporal-Cutoff Benchmarks.} The central difficulty in benchmark construction is obtaining reliable membership labels without access to the model's training corpus. Temporal-cutoff benchmarks solve this by exploiting the model's training cutoff date: texts published before the cutoff are treated as members, while texts published after it are treated as non-members. WikiMIA \citep{shi2023detecting} first instantiated this strategy on Wikipedia event pages, treating pre-2017 articles as members and post-2023 articles as non-members. Subsequent benchmarks extend it along two axes: (i) \textbf{broadening the coverage}, where StackMIA \citep{ye2024data} applies the same January 2017 cutoff to Stack Exchange posts, treating those after May 2023 as non-members, and PatentMIA \citep{zhang2024pretraining} targets Chinese-language patents, treating those published before January 2023 as members and those with priority dates after March 2024 as non-members, to evaluate Chinese LLMs released in that window; and (ii) \textbf{refreshing the cutoff}, where WikiMIA-24 \citep{fu2025mia} retains WikiMIA's original January 2017 members but updates its non-members to articles published after March 2024, extending applicability to more recent models. However, text naturally drifts over time in vocabulary, topic, and style, so pre- and post-cutoff splits differ in more than membership alone, entangling genuine membership signals with distribution shifts. As a result, high performance on them likely reflects the shift rather than genuine membership detection \citep{duan2024membership}.

\paragraph{Partition-Based Benchmarks.} While temporal-cutoff benchmarks infer membership from publication dates, partition-based benchmarks obtain exact labels from a known training corpus, treating instances included in training as members and held-out or later-introduced instances as non-members. MIMIR~\citep{duan2024membership} partitions the documented corpora of open models, namely The Pile~\citep{gao2020pile} and Dolma~\citep{soldaini2024dolma}, spanning a broad range of domains from general web text to academic and code corpora. To decontaminate non-members, it applies a Bloom-filter–based 13-gram deduplication filter against the training corpus, using a permissive overlap threshold of $\leq 80\%$: a candidate non-member is discarded only if \emph{more than} $80\%$ of its 13-grams already appear among training data, so a document sharing up to $80\%$ of its 13-grams with the training data is still retained as a ``non-member.'' Because the threshold is permissive, MIMIR reports substantial residual overlap between members and non-members across all measured domains, consistently exceeding 30\%. In contrast, \textsc{OLMo-Detect} applies 13-gram filtering with a far stricter $\leq 20\%$ threshold, reducing residual overlap to $5.31\%$ on average (Table~\ref{table:benchmark_verification}). Rather than partitioning an existing corpus, RL-MIA \citep{tao2025detecting} manufactures its partition to simulate contamination during RL post-training, drawing from AIME24, AIME25, K\&K \citep{xie2025memorization}, and SAT \citep{liu2025saturn}. Specifically, for AIME24 and AIME25, half of each benchmark's items are injected into the OpenR1-Math-46K \citep{guha2025openthoughts} RL training corpus as members, with the remainder held out as non-members, while for K\&K and SAT, original training instances serve as members and additional items synthesized via the same generation procedure serve as non-members. However, similar to MIMIR, RL-MIA provides no explicit mechanism to verify or control distributional alignment between its members and non-members.

\paragraph{Checkpoint-Based Benchmarks.} OLMoMIA \citep{kim2026detecting} constructs members and non-members from the training checkpoints of OLMo-7B \citep{groeneveld-etal-2024-olmo}: members are examples seen before step 100k, and non-members are examples introduced between steps 400k and 500k, leaving the intervening steps unused. OLMoMIA is the only existing benchmark to explicitly control distributional alignment between members and non-members, via semantic similarity. Specifically, it embeds members and non-members using a text encoder, NV-Embed-v2 \citep{lee2025nv}, clusters each split separately with $k$-means, and selects cluster pairs based on their inter-cluster distance, producing settings that range from maximally separated to near-indistinguishable. However, most MIA methods exploit surface-form signals that manifest through exact wording rather than meaning, while semantic similarity does not entail lexical similarity. For example, two texts describing the same event using synonyms can lie close in NV-Embed-v2's embedding space while sharing few overlapping $n$-grams. Consequently, two instances considered well-aligned under OLMoMIA's construction may still differ substantially in surface-form statistics that MIAs actually exploit, leading to a lexical shortcut that can inflate performance independent of membership signals. 


\paragraph{Model-Output-Based Benchmarks.} Both preceding strategies label texts, model-output-based benchmarks instead label model outputs, constructing members and non-members from the generations of models with and without exposure to the evaluation data. DETCON \citep{dong2024generalization} simulates membership during the supervised fine-tuning (SFT) stage by fine-tuning a base LLM via low-rank adaptation (LoRA) \citep{hu2021lora} on HumanEval \citep{chen2021evaluating} (code generation) and GSM8K \citep{cobbe2021trainingverifierssolvemath} (mathematical reasoning), varying the mixing ratio, learning rate, and occurrence count of these benchmarks in the fine-tuning data. Outputs from the fine-tuned model are treated as members, while outputs from the corresponding base model are treated as non-members. 
However, since members and non-members come from different model states rather than a single model, differences arising from the fine-tuning procedure, such as LoRA's own effect on output style independent of the injected content, may provide a shortcut to separate the data. Moreover, LoRA fine-tuning exhibits dynamics that diverge from full-parameter training \citep{biderman2024loralearnsforgets}, so detection signals calibrated on DETCON's LoRA-simulated membership may not transfer to membership arising from real-world LLM training.

\section{Detailed Descriptions of Existing MIAs}
\label{appendix:related_work_methods}

This section provides a taxonomy of existing MIAs. Notably, we restrict our scope to those applicable in the gray-box setting defined in Section~\ref{task_def}. These methods can be grouped into unsupervised and supervised methods based on whether they require labeled members and non-members.

\paragraph{Unsupervised Methods.} We classify unsupervised methods into four categories based on the signal they exploit: likelihood-based, prompt-based, perturbation-based, and output-distribution-based.

\subparagraph{Likelihood-based.} These methods infer membership from the model's token-level likelihoods, exploiting the tendency of LLMs to assign higher probability to previously seen text. The earliest instantiation, Perplexity \citep{yeom2018privacy}, directly thresholds the model's loss on the candidate text, based on the observation that the gap between member and non-member loss reflects overfitting. However, raw likelihood conflates two causes: a text may receive high probability because it was memorized, or merely because it is inherently predictable. Subsequent methods differ primarily in what they calibrate the likelihood against to remove this confound. The first family calibrates against properties of the candidate text itself: Zlib and Lowercase \citep{carlini2021extracting} normalize the model's perplexity by the text's zlib-compressed entropy and the perplexity of its lowercased variant, respectively; Min-K\% \citep{shi2023detecting} restricts scoring to the $k\%$ least-probable tokens, hypothesizing that unseen text contains outlier tokens with abnormally low probability, and its extension Min-K\%++ \citep{zhang2024min} further normalizes each token's log-probability by the mean and standard deviation over the vocabulary at that position, exploiting the insight that maximum-likelihood training pushes member tokens towards local maxima of the output distribution; DCPDD \citep{zhang2024pretraining} calibrates token probabilities against a token-frequency prior estimated from a reference corpus, computing the cross-entropy between the two distributions to avoid misclassifying unseen text composed of frequent tokens. The second family calibrates against augmented variants of the candidate: PAC \citep{ye2024data} extends Min-K\% by constructing adjacent samples via word-level augmentation and computing a polarized distance that contrasts near and far probability regions to expose overfitted training samples; although it shares the augmentation idea with the perturbation-based Neighborhood Attack below, its final score remains a Min-K\%-style likelihood statistic, so we group it here. The third family calibrates against other models: Smaller Ref \citep{carlini2021extracting} scores candidates by the ratio of the target model's perplexity to that of a smaller model from the same family, exploiting the fact that smaller models memorize less of the shared training distribution, and SPV-MIA \citep{fu2024membership} removes the need for an externally trained reference by fine-tuning one on text elicited from the target model itself through self-prompting, scoring membership via a probabilistic-variation signal computed over perturbed variants of the candidate. Finally, the fourth family calibrates against the conditioning context rather than the text or model: ReCaLL \citep{xie2024recall} measures the relative drop in conditional log-likelihood when the candidate is prefixed with known non-member text, based on the finding that such conditioning reduces the likelihood of members substantially more than that of non-members; EMMIA \citep{kim2026detecting} addresses ReCaLL's dependence on prefix selection by iteratively refining prefix scores through expectation-maximization; CAMIA \citep{chang2025context} extracts richer signals from the per-token loss sequence, conditioning on prefix properties such as length, token diversity, and prefix perplexity, and computing loss-dynamics statistics including slope, fluctuation, and outlier counts, motivated by the observation that member losses decrease more smoothly as context accumulates.

\subparagraph{Perturbation-based.} Rather than calibrating the likelihood of the candidate alone, perturbation-based methods treat the candidate's synthetic neighbors as the reference: the neighbors match the candidate in content and fluency, so any gap between the candidate and its neighbors isolates memorization of surface form from inherent predictability. Neighborhood Attack \citep{mattern2023membership} generates neighbors via masked-word substitution with a mask-filling model and scores membership by the difference between the candidate's log-likelihood and the mean log-likelihood of its neighbors, flagging texts whose likelihood stands out from their own neighborhood.

\subparagraph{Prompt-based.} Rather than reading likelihoods, prompt-based methods probe memorization through the model's responses to prompts, differing primarily in how much they withhold and what they ask the model to do with the rest. Guided Instruction \citep{golchin2023time} withholds the most: by prompting the model with the dataset name, partition, and the initial segment of an instance, it measures whether the completion reproduces the entire withheld remainder, scoring the completion against the reference continuation with ROUGE-L \citep{lin-2004-rouge} and BLEURT \citep{sellam-etal-2020-bleurt} and additionally employing an LLM judge to identify exact or near-exact matches. Comparing this guided completion with one generated from a general instruction that omits dataset identifiers isolates the contribution of dataset-specific memorization. Narrowing the target from a free-form continuation to a single slot, \citet{chang2023speak} employ a name-cloze formulation in which a proper name is masked in a book passage and the model is asked to recover it without additional context, making successful recovery indicative of memorization of the source text, while TS-Guessing \citep{deng2024investigating} adapts the same cloze principle, masking an incorrect answer option in a multiple-choice question and flagging membership when the exact missing option is reproduced. This narrowing reflects a trade-off: free-form completion probes more of the memorized content but requires soft matching, whereas slot filling admits exact-match verification at the cost of probing only a fragment. At the far end of this spectrum, DCQ \citep{golchin2025data} withholds nothing and shifts from production to recognition: it casts membership inference as a five-way multiple-choice quiz in which the original candidate appears alongside four word-level paraphrases that preserve its semantics. Since all options are semantically equivalent, a consistent preference for the original is attributable to memorization of surface form rather than content understanding.

\subparagraph{Output-distribution-based.} These methods share a common premise: membership leaves a fingerprint of abnormal concentration in the model's output distribution. CDD \citep{dong2024generalization} reads this fingerprint at the string level, sampling multiple outputs for the same input and measuring the peakedness of their distribution around the greedy output via edit distance, flagging membership when the outputs are abnormally concentrated. Self-Critique \citep{tao2025detecting} reads the same fingerprint in a setting where it is sharpest: membership introduced during RL post-training. Motivated by the observation that RL collapses the model's output entropy onto narrow reasoning paths, it prompts the model to produce two distinct responses to the same problem and flags membership when the two exhibit high similarity in the entropy space.

\paragraph{Supervised Methods.} We classify supervised methods into two categories based on how they make use of labeled members and non-members: likelihood-based and tuning-based.

\subparagraph{Likelihood-based.} A supervised variant of CAMIA \citep{chang2025context} uses labeled members and non-members to train a downstream classifier, leaving the target model untouched. It represents each candidate as a feature vector of its per-token loss-dynamics signals and trains a logistic regression classifier to combine them, applying PCA within each signal group to control dimensionality. The underlying likelihood-derived signals are unchanged, only their combination is learned.

\subparagraph{Tuning-based.} \label{appendix:related_work_methods_fine_tuning} These methods use labeled data to tune the target model, differing in what is tuned and what the tuning is used to achieve. MIA-Tuner \citep{fu2025mia} keeps the model weights frozen and tunes a soft prompt injected into the target model, using both members and non-members, so that the tuned model becomes the detector directly: an instruction-tuning objective trains aligned models to answer \textit{Yes} or \textit{No} to a membership query, while a contrastive objective trains unaligned models, which cannot follow such instructions, to amplify the loss gap between members and non-members. FSD \citep{zhang2025fine} instead fine-tunes the model weights and is motivated by an empirical finding: briefly fine-tuning the target model on a small set of non-members disproportionately lowers non-members' scores under an existing method (e.g.\ Perplexity, Min-K\%), while leaving members' scores largely unchanged. Rather than tuning the model to become the detector itself, as MIA-Tuner does, FSD uses this asymmetric shift as its detection signal: membership is inferred from a candidate's score deviation before versus after fine-tuning, and it requires only non-members for tuning.

\section{Detailed Descriptions of OLMo~2 Training Data}
\label{appendix:olmo2_training_data}

This appendix provides a detailed breakdown of OLMo~2's training data across the three stages described in Section~\ref{sec:olmo2_data}. Token counts, document counts, and filtering details are reproduced from the original OLMo~2 paper \citep{olmo20242}.

\subsection{Pre-training: OLMo~2 Mix~1124}
\label{sec:app_olmo2_pretrain}
OLMo~2 Mix~1124 comprises seven domains drawn from three source corpora. From DCLM \citep{li2024datacomp}, OLMo~2 uses DCLM-Baseline (web pages), the top 10\% quality subset of deduplicated DCLM-RefinedWeb (a RefinedWeb \citep{penedo2023refinedwebdatasetfalconllm}-style heuristically filtered version of the raw DCLM-Pool corpus), selected using a fastText classifier \citep{joulin2017bag}. From Dolma~1.7 \citep{soldaini2024dolma}, it incorporates five subsets: peS2o \citep{peS2o} (academic papers), arXiv (STEM papers; originally sourced from RedPajama \citep{weber2024redpajama}), OpenWebMath \citep{paster2024openwebmath} (math web pages), Algebraic Stack \citep{azerbayev2023llemma} (math proofs and code), and Wikipedia/Wikibooks (encyclopedic content). The seventh domain, StarCoder \citep{li2023starcoder} (code), is derived from permissively licensed GitHub repositories. StarCoder is filtered to retain only repositories with at least 2 GitHub stars; within the retained set, documents whose most-frequent word constitutes over 30\% of tokens or whose top-2 most-frequent words constitute over 50\% are discarded.

The full per-domain composition of OLMo~2 Mix~1124 is summarized in Table~\ref{table:app:olmo2_pretrain}. The 1B and 7B models are pretrained on 4 trillion tokens of this mix (1 epoch), the 13B on 5 trillion (1.2 epochs), and the 32B on 6 trillion (1.5 epochs).

\begin{table*}[t]
\centering
\resizebox{\textwidth}{!}{%
\begin{tabular}{@{}l l r r r r@{}}
\toprule
\textbf{Domain} & \textbf{Type} & \textbf{Tokens} & \textbf{Words} & \textbf{Bytes} & \textbf{Docs} \\ \midrule
DCLM-Baseline           & Web pages         & 3.71T  & 3.32T  & 21.32T & 2.95B \\
StarCoder (filtered)    & Code              & 83.0B  & 70.0B  & 459B   & 78.7M \\
peS2o                   & Academic papers   & 58.6B  & 51.1B  & 413B   & 38.8M \\
arXiv                   & STEM papers       & 20.8B  & 19.3B  & 77.2B  & 3.95M \\
OpenWebMath             & Math web pages    & 12.2B  & 11.1B  & 47.2B  & 2.89M \\
Algebraic Stack         & Math proofs/code  & 11.8B  & 10.8B  & 44.0B  & 2.83M \\
Wikipedia/Wikibooks  & Encyclopedic      & 3.7B   & 3.16B  & 16.2B  & 6.17M \\ \midrule
\textbf{Total}          &                   & \textbf{3.90T} & \textbf{3.48T} & \textbf{22.38T} & \textbf{3.08B} \\
\bottomrule
\end{tabular}%
}
\caption{Composition of OLMo~2 Mix~1124, the pre-training corpus of the OLMo~2 suite, reproduced from Table~4 of \citet{olmo20242}.}
\label{table:app:olmo2_pretrain}
\end{table*}

\begin{table*}[t]
\centering
\resizebox{\textwidth}{!}{%
\begin{tabular}{@{}l l r r r r@{}}
\toprule
\textbf{Domain} & \textbf{Type} & \textbf{Tokens} & \textbf{Words} & \textbf{Bytes} & \textbf{Docs} \\ \midrule
\multicolumn{6}{@{}l}{\textit{High-Quality Subset (five domains)}} \\
Filtered DCLM-Baseline                             & High-quality web & 752B   & 670B  & 4.56T  & 606M  \\
FLAN                                               & Instruction data & 17.0B  & 14.4B & 98.2B  & 57.3M \\
peS2o                                              & Academic papers  & 58.6B  & 51.1B & 413B   & 38.8M \\
Wikipedia/Wikibooks                                & Encyclopedic     & 3.7B   & 3.16B & 16.2B  & 6.17M \\
Stack Exchange                                     & Q\&A             & 1.26B  & 1.14B & 7.72B  & 2.48M \\
\textbf{High-quality total}                        &                  & \textbf{832.6B} & \textbf{739.8B} & \textbf{5.09T} & \textbf{710.8M} \\ \midrule
\multicolumn{6}{@{}l}{\textit{Dolmino Math Mix (one domain, seven subsets)}} \\
TuluMath                                       & Synthetic math   & 230M   & 222M  & 1.03B  & 220K  \\
DolminoSynthMath                               & Synthetic math   & 28.7M  & 35.1M & 163M   & 725K  \\
TinyGSM-MIND                                   & Synthetic math   & 6.48B  & 5.68B & 25.52B & 17M   \\
MathCoder2 Synth Books                         & Synthetic math   & 3.87B  & 3.71B & 18.4B  & 2.83M \\
Metamath (OWM-filtered)                        & Math             & 84.2M  & 76.6M & 741M   & 383K  \\
CodeSearchNet (OWM-filtered)$^{\dagger}$       & Code             & 1.78M  & 1.41M & 29.8M  & 7.27K \\
GSM8K (training split)                         & Math             & 2.74M  & 3.00M & 25.3M  & 17.6K \\
\textbf{Dolmino Math Mix total}                &                  & \textbf{10.7B} & \textbf{9.73B} & \textbf{45.9B} & \textbf{21.37M} \\
\bottomrule
\end{tabular}%
}
\caption{
Composition of Dolmino Mix~1124, the mid-training corpus of the OLMo~2 suite, reproduced from Table~5 of \citet{olmo20242}. The High-Quality Subset contributes five domains directly; the Dolmino Math Mix is itself a single domain whose composition is given by seven subsets. ``OWM-filtered'' indicates that the subset is further filtered using the OpenWebMath classifier designed by \citet{olmo20242}. $^{\dagger}$Although CodeSearchNet (OWM-filtered) is listed as part of the Dolmino Math Mix in \citet{olmo20242}, it was not used in the final training.
}
\label{table:app:olmo2_midtrain}
\end{table*}

\begin{table*}[t]
\centering
\resizebox{\textwidth}{!}{%
\begin{tabular}{@{}lrcccccc@{}}
\toprule
\multirow{2}{*}{\textbf{Domain}} & \multirow{2}{*}{\textbf{Tokens}} & \multicolumn{2}{c}{\textbf{50B sample}} & \multicolumn{2}{c}{\textbf{100B sample}} & \multicolumn{2}{c}{\textbf{300B sample}} \\
\cmidrule(lr){3-4} \cmidrule(lr){5-6} \cmidrule(lr){7-8}
& & \textbf{Domain \%} & \textbf{Mix \%} & \textbf{Domain \%} & \textbf{Mix \%} & \textbf{Domain \%} & \textbf{Mix \%} \\
\midrule
Filtered DCLM-Baseline & 752B  & 3.23  & 47.2 & 6.85  & 50.2 & 20.78 & 51.9 \\
FLAN    & 17.0B & 50.0  & 16.6 & 100   & 16.7 & 200   & 11.3 \\
Stack Exchange         & 1.26B & 100   & 2.45 & 200   & 2.47 & 400   & 1.68 \\
peS2o                  & 58.6B & 5.15  & 5.85 & 16.7  & 9.52 & 100   & 19.4 \\
Wikipedia/Wikibooks    & 3.7B  & 100   & 7.11 & 100   & 3.57 & 400   & 4.86 \\
Dolmino Math Mix       & 10.7B & 100   & 20.8 & 200   & 17.5 & 400   & 10.8 \\
\bottomrule
\end{tabular}%
}
\caption{
Per-domain composition of the three token-budget samples (50B, 100B, 300B) drawn from Dolmino Mix~1124 for mid-training, reproduced from Table~13 of \citet{olmo20242}. For each domain: ``Domain \%'' is the fraction of that domain's full token count drawn into the sample, with values above 100 indicating repetition (e.g., 400 = 4$\times$ repeat); ``Mix \%'' is the domain's share of the sample's total tokens (each Mix \% column sums to 100\%).
}
\label{table:app:olmo2_midtrain_samples}
\end{table*}

\subsection{Mid-training: Dolmino Mix~1124}
\label{sec:app_olmo2_midtrain}
Mid-training is an intermediate annealing stage between pre-training and post-training, in which the pretrained model is further trained on a smaller, higher-quality data mixture under a decaying learning rate to improve performance on targeted skills such as math and reasoning. OLMo~2's mid-training mixture, Dolmino Mix~1124, comprises six domains organized into a High-Quality Subset and a Dolmino Math Mix. The High-Quality Subset contains five domains: filtered DCLM-Baseline, peS2o, Wikipedia/Wikibooks, FLAN \citep{wei2021finetuned}, and Stack Exchange. Dolmino Math Mix (abbreviated as Math Mix in the main paper) is a heterogeneous mixture aggregating synthetic math data (TuluMath, generated with GPT-4o \citep{hurst2024gpt}; DolminoSynthMath and TinyGSM-MIND, generated with Qwen2.5-7B-Instruct \citep{yang2024qwen2}; and MathCoder2-Synthetic, collected from external sources and filtered using a FastText classifier trained on GPT-4o annotations), the GSM8K training split, OpenWebMath-filtered Metamath, and OpenWebMath-filtered CodeSearchNet. The full per-domain composition of Dolmino Mix~1124 is given in Table~\ref{table:app:olmo2_midtrain}.

Within the High-Quality Subset, peS2o and Wikipedia/Wikibooks are reused from pre-training without modification. Filtered DCLM-Baseline applies additional filtering on top of DCLM-Baseline: documents are first retained by the DCLM FastText classifier with threshold 0.0331 (retaining $\sim \text{65.6}\%$), then restricted to FineWeb-Edu \citep{penedo2024fineweb} scores $\geq \text{2}$ ($\sim \text{20.3}\%$ retained overall). FLAN is decontaminated by removing training, validation, and test instances of evaluation benchmarks. Stack Exchange is taken from the 2024-09-30 dump, restructured into Q\&A format, and filtered to retain only pairs with an accepted answer, at least 3 question votes, and at least 5 answer votes.

Three sample sizes (50B, 100B, 300B tokens) are drawn from the full mix, their per-domain compositions are summarized in Table~\ref{table:app:olmo2_midtrain_samples}. The 1B model anneals on a single 50B sample without averaging. The 7B, 13B, and 32B models apply \textit{model soup} \citep{wortsman2022model}: multiple checkpoints are annealed on the same data mix using different randomized data orders, then averaged into the final midtrained model. The 7B averages three checkpoints trained on 50B samples, while the 13B and 32B each average four checkpoints (three trained on 100B samples and one on a 300B sample).

\subsection{Post-training: Tülu~3 Recipe}
\label{sec:app_olmo2_posttrain}

\begin{table*}[t]
\centering
\resizebox{0.85\textwidth}{!}{%
\begin{tabular}{@{}l r r c@{}}
\toprule
\textbf{Subset} & \textbf{Count} & \textbf{\# in SFT} & \textbf{Sampleable} \\
\midrule
\multicolumn{4}{@{}l}{\textit{General}} \\
Tülu~3 Hardcoded          &            240 &           240 & \\
OpenAssistant             &         88{,}838 &         7{,}132 & \ding{51} \\
No Robots                 &          9{,}500 &         9{,}500 & \\
WildChat (GPT-4 subset)   &        241{,}307 &       100{,}000 & \ding{51} \\
\midrule
\multicolumn{4}{@{}l}{\textit{Knowledge Recall}} \\
FLAN v2                   &         89{,}982 &        89{,}982 & \\
SciRIFF                   &         35{,}357 &        10{,}000 & \ding{51} \\
TableGPT                  &         13{,}222 &         5{,}000 & \ding{51} \\
\midrule
\multicolumn{4}{@{}l}{\textit{Math and Reasoning}} \\
Tülu~3 Persona MATH$^{\dagger}$    &        149{,}960 &       149{,}960 & \\
Tülu~3 Persona GSM$^{\dagger}$     &         49{,}980 &        49{,}980 & \\
Tülu~3 Persona Algebra    &         20{,}000 &        20{,}000 & \\
OpenMathInstruct~2        & 21{,}972{,}791 &        50{,}000 & \ding{51} \\
NuminaMath-TIR            &         64{,}312 &        64{,}312 & \\
\midrule
\multicolumn{4}{@{}l}{\textit{Coding}} \\
Tülu~3 Persona Python     &         34{,}999 &        34{,}999 & \\
Evol CodeAlpaca           &        107{,}276 &       107{,}276 & \\
\midrule
\multicolumn{4}{@{}l}{\textit{Safety and Non-Compliance}} \\
Tülu~3 CoCoNot            &         10{,}983 &        10{,}983 & \\
Tülu~3 WildJailbreak      &         50{,}000 &        50{,}000 & \\
Tülu~3 WildGuardMix       &         50{,}000 &        50{,}000 & \\
\midrule
\multicolumn{4}{@{}l}{\textit{Multilingual}} \\
Aya                       &        202{,}285 &       100{,}000 & \ding{51} \\
\midrule
\multicolumn{4}{@{}l}{\textit{Precise Instruction Following}} \\
Tülu~3 Persona IF         &         29{,}980 &        29{,}980 & \\
\midrule
\textbf{Total}            & \textbf{23{,}221{,}012} & \textbf{939{,}344} & \textbf{6} \\
\bottomrule
\end{tabular}%
}
\caption{Subsets of OLMo~2's SFT mix, reproduced from Table~7 of \citet{lambert2025tulu3pushingfrontiers}. ``Count'' is the total number of prompts available in the upstream source corpus; ``\# in SFT'' corresponds to \texttt{tulu-3-sft-olmo-2-mixture} (7B and 13B; 939{,}344 prompts). The 1B and 32B models use \texttt{tulu-3-sft-olmo-2-mixture-0225} (866{,}138 prompts), which draws from the same 19 subsets but further removes date-cutoff instructions and applies majority voting over 5 completions to the $^{\dagger}$-marked subsets. The \textbf{Sampleable} column marks subsets with non-member candidates available for \textsc{OLMo-Detect}: \ding{51}~=~sampleable remainder available; blank~=~source corpus fully consumed (\# in SFT~=~Count).}
\label{table:app:olmo2_sft}
\end{table*}

\begin{table*}[t]
\centering
\resizebox{\textwidth}{!}{%
\begin{tabular}{@{}lllrcccc c@{}}
\toprule
\textbf{Stage} & \textbf{Domain} & \textbf{Subset} & \textbf{Prompts}$^{*}$ & \textbf{1B} & \textbf{7B} & \textbf{13B} & \textbf{32B} & \textbf{Sampleable} \\ 
\midrule
\multirow{8}{*}{Post-training} & \multirow{8}{*}{DPO} 
    & SFT Reused                   & 117{,}025 & \ding{51} & \ding{51} & \ding{51} & \ding{51} & $\sim$ \\
  & & SFT IF                       & 65{,}792  & \ding{51} & \ding{51} & \ding{51} & \ding{51} & \\
\cmidrule(lr){3-9}
  & & WildChat Unused$^{\ddagger}$ & 84{,}105  & \ding{51} & \ding{51} & \ding{51} & \ding{51} & \ding{51} \\
  & & WildChat Reused$^{\ddagger}$ & 17{,}703  & \ding{51} & \ding{51} & \ding{51} & \ding{51} & \ding{51} \\
  & & WildChat IF$^{\ddagger}$     & 10{,}794  & \ding{51} &           & \ding{51} & \ding{51} & \ding{51} \\
\cmidrule(lr){3-9}
  & & UltraFeedback (cleaned)      & 60{,}816  & \ding{51} & \ding{51} & \ding{51} & \ding{51} & \\
  & & DaringAnteater IF            &  1{,}618  & \ding{51} & \ding{51} & \ding{51} & \ding{51} & \ding{51} \\
  & & Tülu~3 Personas IF           & 19{,}890  & \ding{51} & \ding{51} & \ding{51} & \ding{51} & \ding{51} \\
\midrule
  & & \textbf{Total prompts}       &           & 378.3K    & 366.7K    & 377.7K    & 377.9K    & \textbf{3} \\
\bottomrule
\end{tabular}%
}
\caption{
Prompt subsets for OLMo~2's DPO stage across all four model sizes. $^{*}$Per-subset prompt counts are reproduced from Table~27 of \citet{olmo20242} and apply to the 7B and 13B mixes; the 1B and 32B mixes draw from the same eight subset families with comparable but not identical per-subset counts. The 7B mix omits WildChat IF; chosen responses in the 1B and 32B mixes are additionally filtered to remove date-cutoff mentions. Subsets labeled ``IF'' contain prompts augmented with verifiable instruction-following constraints (IFEval-style). The \textbf{Sampleable} column marks subsets with uncontaminated candidates available for \textsc{OLMo-Detect}: \ding{51}~=~sampleable remainder available; blank~=~source corpus fully consumed; $\sim$~=~unknown (SFT Reused aggregates prompts from many SFT source datasets with varying sampleability, making it impossible to determine remainder without per-source attribution). $^{\ddagger}$All three WildChat subsets draw from the same upstream source corpus and \textbf{together count as one of the three sampleable sources}; the remaining two are DaringAnteater IF and Tülu~3 Personas IF.
}
\label{table:app:olmo2_dpo}
\end{table*}

The post-training stage follows the Tülu~3 recipe \citep{lambert2025tulu3pushingfrontiers} adapted to permissively-licensed sources, and proceeds in three sequential phases: SFT, DPO, and RLVR.

\paragraph{Supervised Fine-Tuning (SFT).}
Supervised fine-tuning (SFT) trains a pretrained language model on a collection of (prompt, response) pairs using next-token-prediction loss, teaching the model to follow instructions and generate well-formatted responses. OLMo~2 draws its SFT prompts from the Tülu~3 SFT data sources \citep{lambert2025tulu3pushingfrontiers}, organized into two mixes: \texttt{tulu-3-sft-olmo-2-mixture} (applied to 7B and 13B models), contains 939{,}344 prompts\footnote{The OLMo~2 paper \citep{olmo20242} reports 939{,}104 prompts for this mix, whereas the released HuggingFace dataset (\url{https://huggingface.co/datasets/allenai/tulu-3-sft-olmo-2-mixture}) contains 939{,}344 prompts. In this paper, we adopt the released dataset as the ground-truth count.}; and \texttt{tulu-3-sft-olmo-2-mixture-0225} (applied to 1B and 32B models), contains 866{,}138 prompts and applies two additional prompt-level filters relative to the first mix: (i) prompts whose instruction text mentions an explicit date cutoff are removed, as training on such prompts caused the resulting model to hallucinate its own date cutoff and emit ``As an AI language model...'' prefaces at inference time; (ii) for the synthetic Persona MATH and Grade School Math subsets, five completions are sampled per prompt, and only prompts whose completions reach a majority answer are retained, reducing the rate of incorrect math answers absorbed during SFT. Both mixes inherit the Tülu~3 subset structure, as detailed in Table~\ref{table:app:olmo2_sft}; the six subsets marked \ding{51}: OpenAssistant, WildChat, SciRIFF, TableGPT, OpenMathInstruct~2, and Aya, are downsampled from larger upstream corpora.

\paragraph{Direct Preference Optimization (DPO).}
Direct preference optimization (DPO) \citep{rafailov2023direct} is a preference-alignment method that fine-tunes a language model directly on pairwise (chosen, rejected) data, bypassing the reward-model-based RL training loop used in conventional reinforcement learning from human feedback (RLHF) \citep{ouyang2022training}. DPO preference data is constructed for each model size via a common synthetic-data pipeline adapted from Tülu~3: (i) candidate responses are sampled from a pool of 20 external models together with on-policy data generated from intermediate OLMo~2 SFT checkpoints; (ii) GPT-4o-2024-08-06 serves as the judge, scoring responses along four dimensions (helpfulness, truthfulness, honesty, and instruction following); and (iii) scores are binarized using the Argilla procedure \citep{cui2023ultrafeedback}, which averages ratings across dimensions, selects the highest-scoring completion as the chosen response, and samples one of the remaining completions as the rejected response. All four mixes draw prompts from the eight Tülu~3 subset families listed in Table~\ref{table:app:olmo2_dpo}, yielding 378.3K prompts for the 1B, 366.7K for the 7B (excluding WildChat IF), 377.7K for the 13B, and 377.9K for the 32B. Chosen responses in the 1B and 32B mixes are additionally filtered to remove instances mentioning a date cutoff.

\paragraph{Reinforcement Learning with Verifiable Rewards (RLVR).}
RLVR \citep{lambert2025tulu3pushingfrontiers} is a reinforcement learning fine-tuning method that replaces the learned reward model used in RLHF with a deterministic rule-based verifier. Completions receive a binary reward based on whether they can be programmatically verified as correct, e.g., through exact-match comparison against a ground-truth mathematical solution or satisfaction of explicit instruction-following constraints. The training data combines three subsets: the GSM8K training set, the MATH \citep{hendrycks2021measuring} training set, and prompts from \citet{lambert2025tulu3pushingfrontiers} containing verifiable instruction-following constraints (IFEval-style). GSM8K and MATH prompts are prepended with 8-shot and 3-shot chain-of-thought (CoT) in-context exemplars, respectively. The 7B and 13B models are trained with proximal policy optimization (PPO) \citep{schulman2017proximal}, with the value function initialized from a reward model trained on the same preference data used for DPO. The 1B and 32B models are trained with group relative policy optimization (GRPO) \citep{shao2024deepseekmath}, which forgoes a reward model. All four models receive a single RLVR pass on the combined mix; the 13B model additionally undergoes two further passes, one on GSM8K alone and one on MATH alone, with the value function re-initialized from the reward model at each subsequent pass. The 1B model additionally undergoes one MATH-only pass (see Table~\ref{table:app:infini_gram_corpora} footnote).

\begin{table*}[t]
\centering
\resizebox{\textwidth}{!}{%
\begin{tabular}{@{}llcccc@{}}
\toprule
\textbf{Stage} & \textbf{Component} & \textbf{1B} & \textbf{7B} & \textbf{13B} & \textbf{32B} \\
\midrule
Pre-training & OLMo~2 Mix~1124 & \ding{51} & \ding{51} & \ding{51} & \ding{51} \\
\midrule
\multirow{3}{*}{Mid-training} 
  & Dolmino Mix~1124, 50B sample  & $\times 1$ & $\times 3$ &            &            \\
  & Dolmino Mix~1124, 100B sample &            &            & $\times 3$ & $\times 3$ \\
  & Dolmino Mix~1124, 300B sample &            &            & $\times 1$ & $\times 1$ \\
\midrule
\multirow{2}{*}{SFT}
  & \texttt{tulu-3-sft-olmo-2-mixture-0225} & $\times 2$ &            &            & $\times 2$ \\
  & \texttt{tulu-3-sft-olmo-2-mixture}      &            & $\times 2$ & $\times 2$ &            \\
\midrule
DPO & Model-specific preference mix$^{*}$ & \ding{51} & \ding{51} & \ding{51} & \ding{51} \\
\midrule
\multirow{3}{*}{RLVR}
  & Combined pass (GSM8K + MATH + IFEval) & \ding{51}           & \ding{51} & \ding{51} & \ding{51} \\
  & Additional GSM8K-only pass            &                     &           & \ding{51} &           \\
  & Additional MATH-only pass$^{\dagger}$ & \ding{51} &           & \ding{51} &           \\
\bottomrule
\end{tabular}%
}
\caption{Per-model training corpora for infini-gram indexing. Entries denote inclusion in the model's indexed corpus, with multiplicities reflecting epoch and model-soup repetition (\ding{51}~=~included once; $\times N$~=~included $N$ times). $^{*}$Model-specific preference mixes: \texttt{olmo-2-0425-1b-preference-mix} (1B), \texttt{olmo-2-1124-7b-preference-mix} (7B), \texttt{olmo-2-1124-13b-preference-mix} (13B), \texttt{olmo-2-0325-32b-preference-mix} (32B). $^{\dagger}$The 1B model's additional MATH-only RLVR pass is inferred from the OLMo~2 training documentation in the \texttt{open-instruct} repository (\url{https://github.com/allenai/open-instruct/blob/main/docs/olmo2.md}).}
\label{table:app:infini_gram_corpora}
\end{table*}

\section{Detailed Per-Domain Statistics of \textsc{OLMo-Detect}}
\label{appendix:full_benchmark_statistics}

We provide detailed per-domain statistics for \textsc{OLMo-Detect}, as introduced in Section~\ref{sec4.3}. Per-domain statistics for pre-training, mid-training, and post-training stages are reported in Tables~\ref{table:appendix_olmo_detect_statistics_1}--\ref{table:appendix_olmo_detect_statistics_3}, while per-subset statistics for domains comprising multiple subsets (StarCoder, Stack Exchange, SFT, and RLVR) are reported in Tables~\ref{table:appendix_olmo_detect_statistics_4}--\ref{table:appendix_olmo_detect_statistics_7}. For each domain, we partition the data into a development set (\textbf{dev}; for hyperparameter tuning) and a test set (\textbf{test}; for method evaluation), with both sample and token ratios set to approximately 1:9.

\paragraph{Interpreting the Lexical Column.} JSD does not reach zero even between two samples drawn from the same source, because each sample misses some of the other's rare tokens. This floor rises as the token budget falls, and StarCoder Java has the smallest budget in the benchmark, 22.1K tokens against 184K--217K elsewhere. Splitting Java's non-members into two halves and comparing them against each other already yields 0.40, above the 0.30 reported for its members against its non-members. Restricting StarCoder Python to Java's token budget reproduces the effect, raising its value from 0.16 to 0.36. Java's 0.30 therefore reflects its size rather than residual misalignment, which the blind bag-of-words classifier confirms independently of sample size: it reaches an AUC of 0.51 on Java (Table~\ref{table:blind_auc_subsets}), indistinguishable from chance.

\begin{table*}[t]
\centering
\resizebox{\textwidth}{!}{%
\begin{tabular}{@{}c l l l l r r l l l@{}}
\toprule
\textbf{Stage} & \textbf{Domain} & \textbf{Subset} & \textbf{Split} & \textbf{Partition} & \textbf{Samples} & \textbf{Tokens} & \textbf{Quality} & \textbf{Temporal} & \textbf{Lexical} \\ \midrule

\multirow{40}{*}{Pre-training} 

& \multirow{7}{*}{DCLM-Baseline}
&
& \multirow{3}{*}{NM} 
& Dev & 16 & 21,745 & FastText: 0.01808 $\pm$ 0.00004 & 2018-01-18 $\pm$ 930d & \multirow{7}{*}{0.07} \\
& & & & Test & 134 & 195,372 & FastText: 0.01809 $\pm$ 0.00008 & 2018-01-19 $\pm$ 930d &  \\ \cmidrule(l){5-9}
& & & & \textbf{Total} & \textbf{150} & \textbf{217,117} & FastText: 0.01809 $\pm$ 0.00007 & 2018-01-19 $\pm$ 930d &  \\ \cmidrule(l){4-9}

& & &\multirow{3}{*}{M} 
& Dev & 16 & 21,721 & FastText: 0.01811 $\pm$ 0.00000 & 2018-01-17 $\pm$ 933d &  \\
& & & & Test & 134 & 195,295 & FastText: 0.01811 $\pm$ 0.00000 & 2018-01-18 $\pm$ 941d &  \\ \cmidrule(l){5-9}
& & & & \textbf{Total} & \textbf{150} & \textbf{217,016} & FastText: 0.01811 $\pm$ 0.00000 & 2018-01-18 $\pm$ 940d &  \\ \cmidrule(l){2-10}

& \multirow{7}{*}{peS2o}
&
& \multirow{3}{*}{NM} 
& Dev & 27 & 21,793 & --- & 2022-12-12 $\pm$ 10d & \multirow{7}{*}{0.10} \\
& & & & Test & 241 & 195,188 & --- & 2022-12-14 $\pm$ 11d &  \\ \cmidrule(l){5-9}
& & & & \textbf{Total} & \textbf{268} & \textbf{216,981} & --- & 2022-12-14 $\pm$ 11d &  \\ \cmidrule(l){4-9}

& & &\multirow{3}{*}{M} 
& Dev & 27 & 21,804 & --- & 2022-12-15 $\pm$ 14d &  \\
& & & & Test & 241 & 195,426 & --- & 2022-12-15 $\pm$ 13d &  \\ \cmidrule(l){5-9}
& & & & \textbf{Total} & \textbf{268} & \textbf{217,230} & --- & 2022-12-15 $\pm$ 13d &  \\ \cmidrule(l){2-10}

& \multirow{7}{*}{OpenWebMath}
&
& \multirow{3}{*}{NM} 
& Dev & 15 & 21,761 & --- & 2019-10-13 $\pm$ 851d & \multirow{7}{*}{0.11} \\
& & & & Test & 136 & 195,422 & --- & 2019-10-13 $\pm$ 853d &  \\ \cmidrule(l){5-9}
& & & & \textbf{Total} & \textbf{151} & \textbf{217,183} & --- & 2019-10-13 $\pm$ 853d &  \\ \cmidrule(l){4-9}

& & &\multirow{3}{*}{M} 
& Dev & 15 & 21,793 & --- & 2019-10-15 $\pm$ 851d &  \\
& & & & Test & 136 & 195,459 & --- & 2019-10-17 $\pm$ 865d &  \\ \cmidrule(l){5-9}
& & & & \textbf{Total} & \textbf{151} & \textbf{217,252} & --- & 2019-10-17 $\pm$ 864d &  \\ \cmidrule(l){2-10}

& \multirow{17}{*}{StarCoder}
& \multirow{17}{*}{\makecell[l]{Python,\\Java}}
& \multirow{5}{*}{NM} 
& Dev & 16 & 21,536 
  & \begin{tabular}[c]{@{}l@{\hspace{6pt}}r@{\hspace{3pt}}l@{}}
      GitHub Stars: & 36.19 & $\pm$ 67.09 \\
      Top-1 Freq:   & 0.30 & $\pm$ 0.01 \\
      Top-2 Freq:   & 0.37 & $\pm$ 0.04
    \end{tabular}
  & ---
  & \multirow{17}{*}{0.15} \\
& & & & Test & 140 & 195,620 
  & \begin{tabular}[c]{@{}l@{\hspace{6pt}}r@{\hspace{3pt}}l@{}}
      GitHub Stars: & 35.59 & $\pm$ 66.61 \\
      Top-1 Freq:   & 0.30 & $\pm$ 0.02 \\
      Top-2 Freq:   & 0.38 & $\pm$ 0.05
    \end{tabular}
  & ---
  &  \\ \cmidrule(l){5-9}
& & & & \textbf{Total} & \textbf{156} & \textbf{217,156} 
  & \begin{tabular}[c]{@{}l@{\hspace{6pt}}r@{\hspace{3pt}}l@{}}
      GitHub Stars: & 35.65 & $\pm$ 66.66 \\
      Top-1 Freq:   & 0.30 & $\pm$ 0.02 \\
      Top-2 Freq:   & 0.38 & $\pm$ 0.05
    \end{tabular}
  & ---
  &  \\ \cmidrule(l){4-9}

& & &\multirow{5}{*}{M} 
& Dev & 16 & 21,587 
  & \begin{tabular}[c]{@{}l@{\hspace{6pt}}r@{\hspace{3pt}}l@{}}
      GitHub Stars: & 36.25 & $\pm$ 65.18 \\
      Top-1 Freq:   & 0.28 & $\pm$ 0.01 \\
      Top-2 Freq:   & 0.36 & $\pm$ 0.04
    \end{tabular}
  & ---
  &  \\
& & & & Test & 140 & 195,635 
  & \begin{tabular}[c]{@{}l@{\hspace{6pt}}r@{\hspace{3pt}}l@{}}
      GitHub Stars: & 36.14 & $\pm$ 64.61 \\
      Top-1 Freq:   & 0.29 & $\pm$ 0.01 \\
      Top-2 Freq:   & 0.36 & $\pm$ 0.05
    \end{tabular}
  & ---
  &  \\ \cmidrule(l){5-9}
& & & & \textbf{Total} & \textbf{156} & \textbf{217,222} 
  & \begin{tabular}[c]{@{}l@{\hspace{6pt}}r@{\hspace{3pt}}l@{}}
      GitHub Stars: & 36.15 & $\pm$ 64.67 \\
      Top-1 Freq:   & 0.29 & $\pm$ 0.01 \\
      Top-2 Freq:   & 0.36 & $\pm$ 0.05
    \end{tabular}
  & ---
  &  \\

\bottomrule
\end{tabular}%
}
\caption{Detailed per-domain statistics for the pre-training stage of \textsc{OLMo-Detect}. \textbf{NM}: non-member; \textbf{M}: member. \textbf{Dev}: development set; \textbf{Test}: test set. The \textbf{Subset} column lists constituent subsets; blank entries indicate single-subset domains whose subset shares the domain's name.}
\label{table:appendix_olmo_detect_statistics_1}
\end{table*}

\begin{table*}[t]
\centering
\resizebox{\textwidth}{!}{%
\begin{tabular}{@{}c l l l l r r l l l@{}}
\toprule
\textbf{Stage} & \textbf{Domain} & \textbf{Subset} & \textbf{Split} & \textbf{Partition} & \textbf{Samples} & \textbf{Tokens} & \textbf{Quality} & \textbf{Temporal} & \textbf{Lexical} \\ \midrule
\multirow{20}{*}{Mid-training} 
& \multirow{7}{*}{Math Mix} 
& \multirow{7}{*}{GSM8K}
& \multirow{3}{*}{NM} 
& Dev & 103 & 21,755 & --- & --- & \multirow{7}{*}{0.04} \\
& & & & Test & 937 & 195,408 & --- & --- & \\ \cmidrule(l){5-9}
& & & & \textbf{Total} & \textbf{1,040} & \textbf{217,163} &  &  & \\ \cmidrule(l){4-9}
& & &\multirow{3}{*}{M} 
& Dev & 103 & 21,729 & --- & --- & \\
& & & & Test & 937 & 195,403 & --- & --- & \\ \cmidrule(l){5-9}
& & & & \textbf{Total} & \textbf{1,040} & \textbf{217,132} &  &  & \\ \cmidrule(l){2-10}
& \multirow{12}{*}{Stack Exchange} 
& \multirow{12}{*}{\makecell[l]{Stack Overflow,\\Math Stack Exchange}}
& \multirow{4}{*}{NM} 
& Dev & 43 & 21,760 &
\begin{tabular}[c]{@{}l@{\hspace{6pt}}r@{\hspace{3pt}}l@{}}
Question Vote: & 3.47 & $\pm$ 2.20 \\
Answer Vote:   & 5.12 & $\pm$ 1.94
\end{tabular}
& 2015-03-27 $\pm$ 1378d
& \multirow{12}{*}{0.08} \\
& & & & Test & 387 & 195,448 &
\begin{tabular}[c]{@{}l@{\hspace{6pt}}r@{\hspace{3pt}}l@{}}
Question Vote: & 3.45 & $\pm$ 2.19 \\
Answer Vote:   & 5.12 & $\pm$ 2.00
\end{tabular}
& 2015-03-31 $\pm$ 1378d
& \\ \cmidrule(l){5-9}
& & & & \textbf{Total} & \textbf{430} & \textbf{217,208} &
\begin{tabular}[c]{@{}l@{\hspace{6pt}}r@{\hspace{3pt}}l@{}}
Question Vote: & 3.45 & $\pm$ 2.19 \\
Answer Vote:   & 5.12 & $\pm$ 1.99
\end{tabular}
& 2015-03-31 $\pm$ 1378d
& \\ \cmidrule(l){4-9}
& & &\multirow{4}{*}{M} 
& Dev & 43 & 21,845 &
\begin{tabular}[c]{@{}l@{\hspace{6pt}}r@{\hspace{3pt}}l@{}}
Question Vote: & 3.56 & $\pm$ 2.18 \\
Answer Vote:   & 5.37 & $\pm$ 1.89
\end{tabular}
& 2015-03-29 $\pm$ 1302d
& \\
& & & & Test & 387 & 195,357 &
\begin{tabular}[c]{@{}l@{\hspace{6pt}}r@{\hspace{3pt}}l@{}}
Question Vote: & 3.52 & $\pm$ 1.68 \\
Answer Vote:   & 5.56 & $\pm$ 1.94
\end{tabular}
& 2015-03-29 $\pm$ 1374d
& \\ \cmidrule(l){5-9}
& & & & \textbf{Total} & \textbf{430} & \textbf{217,202} &
\begin{tabular}[c]{@{}l@{\hspace{6pt}}r@{\hspace{3pt}}l@{}}
Question Vote: & 3.52 & $\pm$ 1.74 \\
Answer Vote:   & 5.54 & $\pm$ 1.94
\end{tabular}
& 2015-03-29 $\pm$ 1367d
& \\
\bottomrule
\end{tabular}%
}
\caption{Sampling statistics for the mid-training stage of \textsc{OLMo-Detect}. \textbf{NM}: non-member; \textbf{M}: member. \textbf{Dev}: development set; \textbf{Test}: test set. The \textbf{Subset} column lists constituent subsets.}
\label{table:appendix_olmo_detect_statistics_2}
\end{table*}

\begin{table*}[t]
\centering
\resizebox{0.93\textwidth}{!}{%
\begin{tabular}{@{}c l l l l l r r l l l@{}}
\toprule
\textbf{Stage} & \textbf{Domain} & \textbf{Subset} & \textbf{Model} & \textbf{Split} & \textbf{Partition} & \textbf{Samples} & \textbf{Tokens} & \textbf{Quality} & \textbf{Temporal} & \textbf{Lexical} \\ \midrule
\multirow{50}{*}{Post-training}
& \multirow{14.5}{*}{SFT}
& \multirow{14.5}{*}{\makecell[l]{WildChat,\\Aya}}
& \multirow{7}{*}{1B, 32B}
& \multirow{4}{*}{NM}
& Dev   & 21    & 21,697   & --- & --- & \multirow{7}{*}{0.11} \\
& & & & & Test  & 204   & 195,360  & --- & --- &  \\ \cmidrule(l){6-10}
& & & & & \textbf{Total} & \textbf{225} & \textbf{217,057} & & & \\ \cmidrule(l){5-10}
& & & & \multirow{3}{*}{M}
& Dev   & 21    & 21,696   & --- & --- &  \\
& & & & & Test  & 204   & 195,463  & --- & --- &  \\ \cmidrule(l){6-10}
& & & & & \textbf{Total} & \textbf{225} & \textbf{217,159} & & & \\ \cmidrule(l){4-11}
& & & \multirow{7}{*}{7B, 13B}
& \multirow{4}{*}{NM}
& Dev   & 16    & 21,959   & --- & --- & \multirow{7}{*}{0.14} \\
& & & & & Test  & 148   & 195,165  & --- & --- &  \\ \cmidrule(l){6-10}
& & & & & \textbf{Total} & \textbf{164} & \textbf{217,124} & & & \\ \cmidrule(l){5-10}
& & & & \multirow{3}{*}{M}
& Dev   & 16    & 21,738   & --- & --- &  \\
& & & & & Test  & 148   & 195,180  & --- & --- &  \\ \cmidrule(l){6-10}
& & & & & \textbf{Total} & \textbf{164} & \textbf{216,918} & & & \\ \cmidrule(l){2-11}
& \multirow{30}{*}{DPO}
& \multirow{30}{*}{UF-WildChat}
& \multirow{7}{*}{1B}
& \multirow{3}{*}{NM}
& Dev   & 11    & 21,705   & --- & --- & \multirow{7}{*}{0.09} \\
& & & & & Test  & 102   & 195,340  & --- & --- &  \\ \cmidrule(l){6-10}
& & & & & \textbf{Total} & \textbf{113} & \textbf{217,045} & & & \\ \cmidrule(l){5-10}
& & & & \multirow{3}{*}{M}
& Dev   & 11    & 21,722   & --- & --- &  \\
& & & & & Test  & 102   & 195,565  & --- & --- &  \\ \cmidrule(l){6-10}
& & & & & \textbf{Total} & \textbf{113} & \textbf{217,287} & & & \\ \cmidrule(l){4-11}
& & & \multirow{7}{*}{7B}
& \multirow{3}{*}{NM}
& Dev   & 10    & 21,704   & --- & --- & \multirow{7}{*}{0.09} \\
& & & & & Test  & 100   & 195,163  & --- & --- &  \\ \cmidrule(l){6-10}
& & & & & \textbf{Total} & \textbf{110} & \textbf{216,867} & & & \\ \cmidrule(l){5-10}
& & & & \multirow{3}{*}{M}
& Dev   & 10    & 21,699   & --- & --- &  \\
& & & & & Test  & 100   & 195,327  & --- & --- &  \\ \cmidrule(l){6-10}
& & & & & \textbf{Total} & \textbf{110} & \textbf{217,026} & & & \\ \cmidrule(l){4-11}
& & & \multirow{7}{*}{13B}
& \multirow{3}{*}{NM}
& Dev   & 10    & 21,730   & --- & --- & \multirow{7}{*}{0.12} \\
& & & & & Test  & 100   & 195,321  & --- & --- &  \\ \cmidrule(l){6-10}
& & & & & \textbf{Total} & \textbf{110} & \textbf{217,051} & & & \\ \cmidrule(l){5-10}
& & & & \multirow{3}{*}{M}
& Dev   & 10    & 21,720   & --- & --- &  \\
& & & & & Test  & 100   & 195,539  & --- & --- &  \\ \cmidrule(l){6-10}
& & & & & \textbf{Total} & \textbf{110} & \textbf{217,259} & & & \\ \cmidrule(l){4-11}
& & & \multirow{7}{*}{32B}
& \multirow{3}{*}{NM}
& Dev   & 10    & 21,838   & --- & --- & \multirow{7}{*}{0.08} \\
& & & & & Test  & 100   & 195,333  & --- & --- &  \\ \cmidrule(l){6-10}
& & & & & \textbf{Total} & \textbf{110} & \textbf{217,171} & & & \\ \cmidrule(l){5-10}
& & & & \multirow{3}{*}{M}
& Dev   & 10    & 21,779   & --- & --- &  \\
& & & & & Test  & 100   & 195,498  & --- & --- &  \\ \cmidrule(l){6-10}
& & & & & \textbf{Total} & \textbf{110} & \textbf{217,277} & & & \\ \cmidrule(l){2-11}
& \multirow{7}{*}{RLVR}
& \multirow{7}{*}{\makecell[l]{CoT-GSM8K,\\CoT-MATH}}
& \multirow{7}{*}{\makecell[l]{1B, 7B,\\13B, 32B}}
& \multirow{3}{*}{NM}
& Dev   & 29    & 21,582   & --- & --- & \multirow{7}{*}{0.01} \\
& & & & & Test  & 261   & 195,528  & --- & --- &  \\ \cmidrule(l){6-10}
& & & & & \textbf{Total} & \textbf{290} & \textbf{217,110} & & & \\ \cmidrule(l){5-10}
& & & & \multirow{3}{*}{M}
& Dev   & 29    & 21,543   & --- & --- &  \\
& & & & & Test  & 261   & 195,482  & --- & --- &  \\ \cmidrule(l){6-10}
& & & & & \textbf{Total} & \textbf{290} & \textbf{217,025} & & & \\
\bottomrule
\end{tabular}%
}
\caption{Detailed per-stage statistics for the post-training stage of \textsc{OLMo-Detect}. \textbf{NM}: non-member; \textbf{M}: member. \textbf{Dev}: development set; \textbf{Test}: test set. The \textbf{Subset} column lists constituent subsets; blank entries indicate single-subset domains whose subset shares the domain's name.}
\label{table:appendix_olmo_detect_statistics_3}
\end{table*}

\begin{table*}[t]
\centering
\resizebox{\textwidth}{!}{%
\begin{tabular}{@{}c l l l l r r l l l@{}}
\toprule
\textbf{Stage} & \textbf{Domain} & \textbf{Subset} & \textbf{Split} & \textbf{Partition} & \textbf{Samples} & \textbf{Tokens} & \textbf{Quality} & \textbf{Temporal} & \textbf{Lexical} \\ \midrule
\multirow{37}{*}{Pre-training}
& \multirow{37}{*}{StarCoder}
& \multirow{17}{*}{Python}
& \multirow{8}{*}{NM}
& Dev & 10 & 19,364 
  & \begin{tabular}[c]{@{}l@{\hspace{6pt}}r@{\hspace{3pt}}l@{}}
      GitHub Stars: & 29.50 & $\pm$ 61.41 \\
      Top-1 Freq:   & 0.31 & $\pm$ 0.01 \\
      Top-2 Freq:   & 0.36 & $\pm$ 0.04
    \end{tabular}
  & ---
  & \multirow{17}{*}{0.16} \\
& & & & Test & 90 & 175,594 
  & \begin{tabular}[c]{@{}l@{\hspace{6pt}}r@{\hspace{3pt}}l@{}}
      GitHub Stars: & 28.53 & $\pm$ 59.41 \\
      Top-1 Freq:   & 0.30 & $\pm$ 0.02 \\
      Top-2 Freq:   & 0.38 & $\pm$ 0.06
    \end{tabular}
  & ---
  &  \\ \cmidrule(l){5-9}
& & & & \textbf{Total} & \textbf{100} & \textbf{194,958} 
  & \begin{tabular}[c]{@{}l@{\hspace{6pt}}r@{\hspace{3pt}}l@{}}
      GitHub Stars: & 28.63 & $\pm$ 59.62 \\
      Top-1 Freq:   & 0.30 & $\pm$ 0.02 \\
      Top-2 Freq:   & 0.37 & $\pm$ 0.05
    \end{tabular}
  & ---
  &  \\ \cmidrule(l){4-9}
& & & \multirow{8}{*}{M}
& Dev & 10 & 19,551 
  & \begin{tabular}[c]{@{}l@{\hspace{6pt}}r@{\hspace{3pt}}l@{}}
      GitHub Stars: & 28.90 & $\pm$ 58.22 \\
      Top-1 Freq:   & 0.29 & $\pm$ 0.01 \\
      Top-2 Freq:   & 0.35 & $\pm$ 0.04
    \end{tabular}
  & ---
  &  \\
& & & & Test & 90 & 175,551 
  & \begin{tabular}[c]{@{}l@{\hspace{6pt}}r@{\hspace{3pt}}l@{}}
      GitHub Stars: & 29.23 & $\pm$ 57.87 \\
      Top-1 Freq:   & 0.29 & $\pm$ 0.01 \\
      Top-2 Freq:   & 0.36 & $\pm$ 0.06
    \end{tabular}
  & ---
  &  \\ \cmidrule(l){5-9}
& & & & \textbf{Total} & \textbf{100} & \textbf{195,102} 
  & \begin{tabular}[c]{@{}l@{\hspace{6pt}}r@{\hspace{3pt}}l@{}}
      GitHub Stars: & 29.20 & $\pm$ 57.90 \\
      Top-1 Freq:   & 0.29 & $\pm$ 0.01 \\
      Top-2 Freq:   & 0.36 & $\pm$ 0.06
    \end{tabular}
  & ---
  &  \\ \cmidrule(l){3-10}
& & \multirow{17}{*}{Java}
& \multirow{8}{*}{NM}
& Dev & 6 & 2,172 
  & \begin{tabular}[c]{@{}l@{\hspace{6pt}}r@{\hspace{3pt}}l@{}}
      GitHub Stars: & 47.33 & $\pm$ 74.29 \\
      Top-1 Freq:   & 0.30 & $\pm$ 0.01 \\
      Top-2 Freq:   & 0.38 & $\pm$ 0.03
    \end{tabular}
  & ---
  & \multirow{17}{*}{0.30} \\
& & & & Test & 50 & 20,026 
  & \begin{tabular}[c]{@{}l@{\hspace{6pt}}r@{\hspace{3pt}}l@{}}
      GitHub Stars: & 48.28 & $\pm$ 76.28 \\
      Top-1 Freq:   & 0.30 & $\pm$ 0.01 \\
      Top-2 Freq:   & 0.39 & $\pm$ 0.03
    \end{tabular}
  & ---
  &  \\ \cmidrule(l){5-9}
& & & & \textbf{Total} & \textbf{56} & \textbf{22,198} 
  & \begin{tabular}[c]{@{}l@{\hspace{6pt}}r@{\hspace{3pt}}l@{}}
      GitHub Stars: & 48.18 & $\pm$ 76.07 \\
      Top-1 Freq:   & 0.30 & $\pm$ 0.01 \\
      Top-2 Freq:   & 0.39 & $\pm$ 0.03
    \end{tabular}
  & ---
  &  \\ \cmidrule(l){4-9}
& & & \multirow{8}{*}{M}
& Dev & 6 & 2,036 
  & \begin{tabular}[c]{@{}l@{\hspace{6pt}}r@{\hspace{3pt}}l@{}}
      GitHub Stars: & 48.50 & $\pm$ 73.75 \\
      Top-1 Freq:   & 0.28 & $\pm$ 0.01 \\
      Top-2 Freq:   & 0.38 & $\pm$ 0.03
    \end{tabular}
  & ---
  &  \\
& & & & Test & 50 & 20,084 
  & \begin{tabular}[c]{@{}l@{\hspace{6pt}}r@{\hspace{3pt}}l@{}}
      GitHub Stars: & 48.56 & $\pm$ 73.64 \\
      Top-1 Freq:   & 0.28 & $\pm$ 0.02 \\
      Top-2 Freq:   & 0.36 & $\pm$ 0.04
    \end{tabular}
  & ---
  &  \\ \cmidrule(l){5-9}
& & & & \textbf{Total} & \textbf{56} & \textbf{22,120} 
  & \begin{tabular}[c]{@{}l@{\hspace{6pt}}r@{\hspace{3pt}}l@{}}
      GitHub Stars: & 48.55 & $\pm$ 73.65 \\
      Top-1 Freq:   & 0.28 & $\pm$ 0.02 \\
      Top-2 Freq:   & 0.36 & $\pm$ 0.04
    \end{tabular}
  & ---
  &  \\
\bottomrule
\end{tabular}%
}
\caption{Detailed per-subset statistics for StarCoder in the pre-training stage of \textsc{OLMo-Detect}. \textbf{NM}: non-member; \textbf{M}: member. \textbf{Dev}: development set; \textbf{Test}: test set.}
\label{table:appendix_olmo_detect_statistics_4}
\end{table*}

\begin{table*}[t]
\centering
\resizebox{\textwidth}{!}{%
\begin{tabular}{@{}c l l l l r r l l l@{}}
\toprule
\textbf{Stage} & \textbf{Domain} & \textbf{Subset} & \textbf{Split} & \textbf{Partition} & \textbf{Samples} & \textbf{Tokens} & \textbf{Quality} & \textbf{Temporal} & \textbf{Lexical} \\ \midrule
\multirow{26}{*}{Mid-training}
& \multirow{26}{*}{Stack Exchange}
& \multirow{12}{*}{Stack Overflow}
& \multirow{6}{*}{NM}
& Dev & 37 & 18,489 
  & \begin{tabular}[c]{@{}l@{\hspace{6pt}}r@{\hspace{3pt}}l@{}}
      Question Vote: & 3.49 & $\pm$ 2.26 \\
      Answer Vote:   & 5.14 & $\pm$ 2.04
    \end{tabular}
  & 2015-02-06 $\pm$ 1381.39d
  & \multirow{12}{*}{0.08} \\
& & & & Test & 331 & 165,394 
  & \begin{tabular}[c]{@{}l@{\hspace{6pt}}r@{\hspace{3pt}}l@{}}
      Question Vote: & 3.48 & $\pm$ 2.24 \\
      Answer Vote:   & 5.14 & $\pm$ 2.05
    \end{tabular}
  & 2015-02-05 $\pm$ 1379.01d
  & \\ \cmidrule(l){5-9}
& & & & \textbf{Total} & \textbf{368} & \textbf{183,883} 
  & \begin{tabular}[c]{@{}l@{\hspace{6pt}}r@{\hspace{3pt}}l@{}}
      Question Vote: & 3.48 & $\pm$ 2.25 \\
      Answer Vote:   & 5.14 & $\pm$ 2.05
    \end{tabular}
  & 2015-02-06 $\pm$ 1379.25d
  & \\ \cmidrule(l){4-9}
& & & \multirow{6}{*}{M}
& Dev & 37 & 18,489 
  & \begin{tabular}[c]{@{}l@{\hspace{6pt}}r@{\hspace{3pt}}l@{}}
      Question Vote: & 3.51 & $\pm$ 2.23 \\
      Answer Vote:   & 5.32 & $\pm$ 1.95
    \end{tabular}
  & 2015-02-06 $\pm$ 1329.25d
  & \\
& & & & Test & 331 & 165,471 
  & \begin{tabular}[c]{@{}l@{\hspace{6pt}}r@{\hspace{3pt}}l@{}}
      Question Vote: & 3.53 & $\pm$ 1.66 \\
      Answer Vote:   & 5.58 & $\pm$ 1.99
    \end{tabular}
  & 2015-02-06 $\pm$ 1378.98d
  & \\ \cmidrule(l){5-9}
& & & & \textbf{Total} & \textbf{368} & \textbf{183,960} 
  & \begin{tabular}[c]{@{}l@{\hspace{6pt}}r@{\hspace{3pt}}l@{}}
      Question Vote: & 3.52 & $\pm$ 1.72 \\
      Answer Vote:   & 5.55 & $\pm$ 1.98
    \end{tabular}
  & 2015-02-06 $\pm$ 1374.07d
  & \\ \cmidrule(l){3-10}
& & \multirow{12}{*}{Math Stack Exchange}
& \multirow{6}{*}{NM}
& Dev & 6 & 3,271 
  & \begin{tabular}[c]{@{}l@{\hspace{6pt}}r@{\hspace{3pt}}l@{}}
      Question Vote: & 3.33 & $\pm$ 1.80 \\
      Answer Vote:   & 5.00 & $\pm$ 1.15
    \end{tabular}
  & 2016-01-22 $\pm$ 1323.12d
  & \multirow{12}{*}{0.13} \\
& & & & Test & 56 & 30,054 
  & \begin{tabular}[c]{@{}l@{\hspace{6pt}}r@{\hspace{3pt}}l@{}}
      Question Vote: & 3.25 & $\pm$ 1.81 \\
      Answer Vote:   & 4.98 & $\pm$ 1.64
    \end{tabular}
  & 2016-02-12 $\pm$ 1329.29d
  & \\ \cmidrule(l){5-9}
& & & & \textbf{Total} & \textbf{62} & \textbf{33,325} 
  & \begin{tabular}[c]{@{}l@{\hspace{6pt}}r@{\hspace{3pt}}l@{}}
      Question Vote: & 3.26 & $\pm$ 1.80 \\
      Answer Vote:   & 4.98 & $\pm$ 1.60
    \end{tabular}
  & 2016-02-10 $\pm$ 1328.71d
  & \\ \cmidrule(l){4-9}
& & & \multirow{6}{*}{M}
& Dev & 6 & 3,356 
  & \begin{tabular}[c]{@{}l@{\hspace{6pt}}r@{\hspace{3pt}}l@{}}
      Question Vote: & 3.83 & $\pm$ 1.86 \\
      Answer Vote:   & 5.67 & $\pm$ 1.49
    \end{tabular}
  & 2016-02-03 $\pm$ 1074.89d
  & \\
& & & & Test & 56 & 29,886 
  & \begin{tabular}[c]{@{}l@{\hspace{6pt}}r@{\hspace{3pt}}l@{}}
      Question Vote: & 3.48 & $\pm$ 1.81 \\
      Answer Vote:   & 5.45 & $\pm$ 1.67
    \end{tabular}
  & 2016-01-28 $\pm$ 1302.79d
  & \\ \cmidrule(l){5-9}
& & & & \textbf{Total} & \textbf{62} & \textbf{33,242} 
  & \begin{tabular}[c]{@{}l@{\hspace{6pt}}r@{\hspace{3pt}}l@{}}
      Question Vote: & 3.52 & $\pm$ 1.82 \\
      Answer Vote:   & 5.47 & $\pm$ 1.65
    \end{tabular}
  & 2016-01-28 $\pm$ 1282.52d
  & \\ 
\bottomrule
\end{tabular}%
}
\caption{Detailed per-subset statistics for Stack Exchange in the mid-training stage of \textsc{OLMo-Detect}. \textbf{NM}: non-member; \textbf{M}: member. \textbf{Dev}: development set; \textbf{Test}: test set.}
\label{table:appendix_olmo_detect_statistics_5}
\end{table*}

\begin{table*}[t]
\centering
\resizebox{\textwidth}{!}{%
\begin{tabular}{@{}c l l l l l r r l l l@{}}
\toprule
\textbf{Stage} & \textbf{Domain} & \textbf{Model} & \textbf{Subset} & \textbf{Split} & \textbf{Partition} & \textbf{Samples} & \textbf{Tokens} & \textbf{Quality} & \textbf{Temporal} & \textbf{Lexical} \\ \midrule
\multirow{30.5}{*}{Post-training}
& \multirow{30.5}{*}{SFT}
& \multirow{15}{*}{1B, 32B}
& \multirow{7}{*}{Wildchat}
& \multirow{3}{*}{NM}
& Dev & 7 & 16,404 & --- & --- & \multirow{7}{*}{0.14} \\
& & & & & Test & 64 & 148,039 & --- & --- & \\ \cmidrule(l){6-10}
& & & & & \textbf{Total} & \textbf{71} & \textbf{164,443} & & & \\ \cmidrule(l){5-10}
& & & & \multirow{3}{*}{M}
& Dev & 7 & 16,438 & --- & --- & \\
& & & & & Test & 64 & 148,115 & --- & --- & \\ \cmidrule(l){6-10}
& & & & & \textbf{Total} & \textbf{71} & \textbf{164,553} & & & \\ \cmidrule(l){4-11}
& & & \multirow{7}{*}{Aya}
& \multirow{3}{*}{NM}
& Dev & 14 & 5,293 & --- & --- & \multirow{7}{*}{0.08} \\
& & & & & Test & 140 & 47,321 & --- & --- & \\ \cmidrule(l){6-10}
& & & & & \textbf{Total} & \textbf{154} & \textbf{52,614} & & & \\ \cmidrule(l){5-10}
& & & & \multirow{3}{*}{M}
& Dev & 14 & 5,258 & --- & --- & \\
& & & & & Test & 140 & 47,348 & --- & --- & \\ \cmidrule(l){6-10}
& & & & & \textbf{Total} & \textbf{154} & \textbf{52,606} & & & \\ \cmidrule(l){3-11}
& & \multirow{15}{*}{7B, 13B}
& \multirow{7}{*}{Wildchat}
& \multirow{3}{*}{NM}
& Dev & 7 & 16,715 & --- & --- & \multirow{7}{*}{0.18} \\
& & & & & Test & 66 & 148,305 & --- & --- & \\ \cmidrule(l){6-10}
& & & & & \textbf{Total} & \textbf{73} & \textbf{165,020} & & & \\ \cmidrule(l){5-10}
& & & & \multirow{3}{*}{M}
& Dev & 7 & 16,688 & --- & --- & \\
& & & & & Test & 66 & 148,362 & --- & --- & \\ \cmidrule(l){6-10}
& & & & & \textbf{Total} & \textbf{73} & \textbf{165,050} & & & \\ \cmidrule(l){4-11}
& & & \multirow{7}{*}{Aya}
& \multirow{3}{*}{NM}
& Dev & 9 & 5,244 & --- & --- & \multirow{7}{*}{0.06} \\
& & & & & Test & 82 & 46,860 & --- & --- & \\ \cmidrule(l){6-10}
& & & & & \textbf{Total} & \textbf{91} & \textbf{52,104} & & & \\ \cmidrule(l){5-10}
& & & & \multirow{3}{*}{M}
& Dev & 9 & 5,050 & --- & --- & \\
& & & & & Test & 82 & 46,818 & --- & --- & \\ \cmidrule(l){6-10}
& & & & & \textbf{Total} & \textbf{91} & \textbf{51,868} & & & \\
\bottomrule
\end{tabular}%
}
\caption{Detailed per-subset statistics for SFT in the post-training stage of \textsc{OLMo-Detect}. \textbf{NM}: non-member; \textbf{M}: member. \textbf{Dev}: development set; \textbf{Test}: test set.}
\label{table:appendix_olmo_detect_statistics_6}
\end{table*}

\begin{table*}[t]
\centering
\resizebox{\textwidth}{!}{%
\begin{tabular}{@{}c l l l l l r r l l l@{}}
\toprule
\textbf{Stage} & \textbf{Domain} & \textbf{Model} & \textbf{Subset} & \textbf{Split} & \textbf{Partition} & \textbf{Samples} & \textbf{Tokens} & \textbf{Quality} & \textbf{Temporal} & \textbf{Lexical} \\ \midrule
\multirow{15}{*}{Post-training}
& \multirow{15}{*}{RLVR}
& \multirow{15}{*}{\makecell[l]{1B, 7B,\\13B, 32B}}
& \multirow{7}{*}{CoT-GSM8K}
& \multirow{3}{*}{NM}
& Dev & 14 & 11,835 & --- & --- & \multirow{7}{*}{0.01} \\
& & & & & Test & 128 & 107,609 & --- & --- & \\ \cmidrule(l){6-10}
& & & & & \textbf{Total} & \textbf{142} & \textbf{119,444} & & & \\ \cmidrule(l){5-10}
& & & & \multirow{3}{*}{M}
& Dev & 14 & 11,796 & --- & --- & \\
& & & & & Test & 128 & 107,563 & --- & --- & \\ \cmidrule(l){6-10}
& & & & & \textbf{Total} & \textbf{142} & \textbf{119,359} & & & \\ \cmidrule(l){4-11}
& & & \multirow{7}{*}{CoT-MATH}
& \multirow{3}{*}{NM}
& Dev & 15 & 9,747 & --- & --- & \multirow{7}{*}{0.01} \\
& & & & & Test & 133 & 87,919 & --- & --- & \\ \cmidrule(l){6-10}
& & & & & \textbf{Total} & \textbf{148} & \textbf{97,666} & & & \\ \cmidrule(l){5-10}
& & & & \multirow{3}{*}{M}
& Dev & 15 & 9,747 & --- & --- & \\
& & & & & Test & 133 & 87,919 & --- & --- & \\ \cmidrule(l){6-10}
& & & & & \textbf{Total} & \textbf{148} & \textbf{97,666} & & & \\
\bottomrule
\end{tabular}%
}
\caption{Detailed per-subset statistics for RLVR in the post-training stage of \textsc{OLMo-Detect}. \textbf{NM}: non-member; \textbf{M}: member. \textbf{Dev}: development set; \textbf{Test}: test set.}
\label{table:appendix_olmo_detect_statistics_7}
\end{table*}

\section{Detailed Descriptions of Non-Member Construction in \textsc{OLMo-Detect}}
\label{appendix:non_member_construction}
This appendix details the construction of non-members for each domain in \textsc{OLMo-Detect}.

\subsection{Source-Matched Candidate Pools}

\paragraph{Domains with Held-Out Data.} These domains have a natural held-out split whose source already matches OLMo~2's training data. 

Pre-training: (i) \textbf{peS2o}, which has two held-out files available, \href{https://huggingface.co/datasets/allenai/peS2o/blob/main/data/v2/validation-00000-of-00002.json.gz}{\texttt{validation-00000-of-00002.json.gz}} and \href{https://huggingface.co/datasets/allenai/peS2o/blob/main/data/v2/validation-00001-of-00002.json.gz}{\texttt{validation-00001-of-00002.json.gz}}. (ii) \textbf{OpenWebMath}, which has two held-out files available, \href{https://huggingface.co/datasets/EleutherAI/proof-pile-2/blob/main/open-web-math/test/test.jsonl.zst}{\texttt{test.jsonl.zst}} and \href{https://huggingface.co/datasets/EleutherAI/proof-pile-2/blob/main/open-web-math/validation/val.jsonl.zst}{\texttt{val.jsonl.zst}}. 

Mid-training: \textbf{Math Mix}, whose GSM8K subset has one held-out file available, \href{https://huggingface.co/datasets/openai/gsm8k/blob/main/main/test-00000-of-00001.parquet}{\texttt{test-00000-of-00001.parquet}}. We additionally exclude the 200 instances designated by \citet{olmo20242} as a development set for evaluation. 

Post-training: \textbf{RLVR}, whose held-out files for CoT-GSM8K and CoT-MATH subsets can be constructed by prepending the same CoT exemplars used in OLMo~2's RLVR setup (8-shot for GSM8K and 3-shot for MATH) to GSM8K's \href{https://huggingface.co/datasets/openai/gsm8k/blob/main/main/test-00000-of-00001.parquet}{\texttt{test-00000-of-00001.parquet}} and the test split of MATH's \href{https://huggingface.co/datasets/qwedsacf/competition_math/blob/main/data/train-00000-of-00001-7320a6f3aba8ebd2.parquet}{\texttt{train-00000-of-00001-7320a6f3aba8ebd2.parquet}} (separated from the training portion during preprocessing), respectively.

\paragraph{Domains with the Remainder of Original Source Corpora.} These domains lack a natural held-out split but are sampled from larger corpora. Therefore, candidate non-member pools are drawn from the portions of those original corpora that OLMo~2 did not use for training. 

Pre-training: (i) \textbf{DCLM-Baseline.} DCLM-Baseline is the top 10\% quality subset of deduplicated DCLM-RefinedWeb. From the remainder of \href{https://data.commoncrawl.org/contrib/datacomp/DCLM-refinedweb/index.html}{DCLM-RefinedWeb}, we randomly select three shards as candidate pools (followed by deduplication): \texttt{global-shard\_01-local-shard\_0}, \texttt{global-shard\_05-local-shard\_6}, and \texttt{global-shard\_07-local-shard\_3}. (ii) \textbf{StarCoder}. OLMo~2's StarCoder is a quality-thresholded subset of the original StarCoder dataset \citep{li2023starcoder}, retaining only data from repositories with at least 2 GitHub stars, where the most-frequent word constitutes at most 30\% of tokens and the top-2 most-frequent words constitute at most 50\%. The top-5 StarCoder subsets by training provenance are GitHub issues, Git commits, Markdown, Java, and Python, from which we select two distinct programming languages, Java and Python. From the rejected remainder of these two subsets, we consider the following files as candidate pools: Java: \href{https://huggingface.co/datasets/bigcode/starcoderdata/blob/main/java/train-00000-of-00087.parquet}{\texttt{train-00000-of-00087.parquet}} through \href{https://huggingface.co/datasets/bigcode/starcoderdata/blob/main/java/train-00086-of-00087.parquet}{\texttt{train-00086-of-00087.parquet}}; Python: \href{https://huggingface.co/datasets/bigcode/starcoderdata/blob/main/python/train-00000-of-00059.parquet}{\texttt{train-00000-of-00059.parquet}} through \href{https://huggingface.co/datasets/bigcode/starcoderdata/blob/main/python/train-00058-of-00059.parquet}{\texttt{train-00058-of-00059.parquet}}. 

Mid-training: \textbf{Stack Exchange.} OLMo~2's Stack Exchange is a filtered version of \href{https://archive.org/details/stackexchange_20240930}{2024-09-30 dump}, retaining only data with an accepted answer, at least 3 question votes, and at least 5 answer votes. The candidate non-member pool is therefore the rejected remainder of the original 2024-09-30 dump, processed using the same pipeline as OLMo~2 to obtain matched Q\&A format. 

Post-training: (i) \textbf{SFT.} Six subsets of OLMo~2's SFT are randomly downsampled from larger upstream corpora: OpenAssistant, WildChat, SciRIFF, TableGPT, OpenMathInstruct~2, and Aya (see Table~\ref{table:app:olmo2_sft} for details). Among them, we select two distinct subsets, WildChat and Aya. The candidate pools for SFT are therefore the unused portions of original WildChat (\href{https://huggingface.co/datasets/allenai/WildChat-1M/blob/main/data/train-00000-of-00014.parquet}{\texttt{train-00000-of-00014.parquet}} to \href{https://huggingface.co/datasets/allenai/WildChat-1M/blob/main/data/train-00013-of-00014.parquet}{\texttt{train-00013-of-00014.parquet}}) and original Aya (\href{https://huggingface.co/datasets/CohereLabs/aya_dataset/blob/main/data/test-00000-of-00001.parquet}{\texttt{test-00000-of-00001.parquet}} and \href{https://huggingface.co/datasets/CohereLabs/aya_dataset/blob/main/data/train-00000-of-00001.parquet}{\texttt{train-00000-of-00001.parquet}}). (ii) \textbf{DPO.} Five of OLMo~2's DPO prompt subsets are randomly sampled from larger corpora: three WildChat variants (Unused, Reused, and IF), DaringAnteater, and Tülu~3 Personas IF (see Table~\ref{table:app:olmo2_dpo} for details). However, we only select WildChat (Unused) despite DPO's multi-subset nature. This is because neither DaringAnteater nor Tülu~3 Personas IF yields sufficient non-members after infini-gram-based 13-gram filtering; meanwhile, the three WildChat variants are not independently selectable either: since all three draw from the same upstream WildChat corpus, they share a single non-member remainder pool rather than three separate ones, and thus jointly count as one sampleable source. We therefore retain WildChat (Unused) alone, considering the remaining parts of original WildChat not used for OLMo~2 training (\href{https://huggingface.co/datasets/allenai/WildChat-1M/blob/main/data/train-00000-of-00014.parquet}{\texttt{train-00000-of-00014.parquet}} to \href{https://huggingface.co/datasets/allenai/WildChat-1M/blob/main/data/train-00013-of-00014.parquet}{\texttt{train-00013-of-00014.parquet}}) as candidate prompt pools. We then apply OLMo~2's UltraFeedback pipeline for response generation and preference annotation to obtain the source-matched candidate pools, yielding the final subset named UltraFeedback-WildChat (abbreviated as UF-Wildchat in the main paper). (iii) \textbf{RLVR.} The IFEval subset of OLMo~2's RLVR is randomly sampled from the \href{https://huggingface.co/datasets/allenai/tulu-v2-sft-mixture}{Tülu~2 SFT mix} \citep{ivison2023camels} and augmented with verifiable instruction-following constraints. Its candidate pool is therefore the remainder of the Tülu~2 SFT mix, augmented via the same constraint-construction pipeline.

\paragraph{RLVR Subset Selection.} As illustrated above, RLVR has three sampleable subsets: CoT-GSM8K and CoT-MATH, drawn from held-out data, and IFEval, drawn from the remainder of the original source corpus. IFEval would add the most within-domain diversity, but we find that its non-members cannot be aligned lexically with members. Since alignment takes precedence over diversity, we select CoT-GSM8K and CoT-MATH, accepting that the two share a similar CoT construction.

\paragraph{Excluded Domains.} Pretraining's Wikipedia/Wikibooks is excluded, as no source-matched candidate non-member pool is available.

\subsection{Domain-Dependent Sampling Strategies}

For domains whose candidate pools originate from either held-out files or the remainder of their source corpora without quality thresholding, sampling follows the random-sampling procedure described in Section~\ref{sec:sampling_strategies}. For quality-thresholded domains, we use boundary sampling to obtain candidate non-members whose quality is just below OLMo~2's selection threshold. Detailed descriptions are provided below.

\paragraph{DCLM-Baseline.} We first identify the minimum fastText score in OLMo~2's DCLM-Baseline: 0.01811. Within the candidate pool, we then sample instances closest to this threshold, yielding over 350 candidate non-members (420K tokens) with an average fastText score of around 0.01809.

\paragraph{StarCoder.} From the candidate pool, we sample instances whose top-1 word frequency is closest to 0.3 and top-2 word frequency is closest to 0.5, while enforcing the same GitHub-star constraint as OLMo~2, requiring each instance to have at least 2 stars. This yields 56 candidate non-members (22K tokens) for Java, with an average GitHub-star count of 74 and top-1/top-2 word frequencies of 0.30 and 0.39, and 250 candidate non-members (300K tokens) for Python, with an average GitHub-star count of 28 and top-1/top-2 word frequencies of 0.30 and 0.37. StarCoder is the only domain where the proportional allocation specified in Section~\ref{token_budget} cannot be met: although Java constitutes a larger share of OLMo~2's training data than Python and would therefore be expected to contribute more tokens, too few Java candidates satisfy our filtering criteria to meet that target. We therefore retain all available Java non-members (56 instances, 22K tokens) and use Python to fill the remainder of the domain's $\sim$217K-token budget.

\paragraph{Stack Exchange.} From the candidate pool, we sample instances satisfying either (a) question vote $= \text{2}$ and answer vote $\geq \text{5}$, or (b) question vote $\geq \text{3}$ and answer vote $= \text{4}$. This yields over 400 candidate non-members (200K tokens) for Stack Overflow, with an average question vote of 3.48 and answer vote of 5.14, and around 100 candidate non-members (50K tokens) for Math Stack Exchange, with an average question vote of 3.26 and answer vote of 4.98.

\subsection{Infini-Gram-Based 13-Gram Filtering}
After obtaining candidate non-members for each domain, we perform 13-gram filtering via infini-gram to retain those with minimal training-data overlap, as described in Section~\ref{13_gram_filtering}. Here, we detail the index construction: we build an infini-gram index over each OLMo~2 model's full training corpus, defined as the union of all training-stage documents with multiplicities reflecting epoch and model-soup repetition. The composition for each model size is summarized in Table~\ref{table:app:infini_gram_corpora}.

\paragraph{Excluded Domains.} Although source-matched candidate pools are identified for arXiv, using held-out files \href{https://huggingface.co/datasets/EleutherAI/proof-pile-2/blob/main/arxiv/test/arXiv_000.jsonl.zst}{\texttt{arXiv\_000.jsonl.zst}} through \href{https://huggingface.co/datasets/EleutherAI/proof-pile-2/blob/main/arxiv/test/arXiv_099.jsonl.zst}{\texttt{arXiv\_099.jsonl.zst}} (test) and \href{https://huggingface.co/datasets/EleutherAI/proof-pile-2/blob/main/arxiv/validation/arXiv_000.jsonl.zst}{\texttt{arXiv\_000.jsonl.zst}} through \href{https://huggingface.co/datasets/EleutherAI/proof-pile-2/blob/main/arxiv/validation/arXiv_099.jsonl.zst}{\texttt{arXiv\_099.jsonl.zst}} (validation), and for Algebraic Stack, using held-out files in its \href{https://huggingface.co/datasets/EleutherAI/proof-pile-2/tree/main/algebraic-stack/test}{test} and \href{https://huggingface.co/datasets/EleutherAI/proof-pile-2/tree/main/algebraic-stack/validation}{validation} directories, no candidates from either pool survive 13-gram filtering. Therefore, both arXiv and Algebraic Stack are excluded.

\subsection{Token Budget Allocation}
We sample from non-members that survive 13-gram filtering, following the rules in Section~\ref{token_budget}. To ensure meaningful statistical analysis, each domain contains at least 100 non-members, with at least 50 instances per subset in multi-subset domains, to ensure meaningful statistical analysis.

\section{Detailed Descriptions of Member Construction in \textsc{OLMo-Detect}} 
\label{appendix:member_construction}
This appendix details the construction of members for each domain in \textsc{OLMo-Detect}. 

\subsection{Candidate Pools} 
\subsubsection{Domains without Quality Thresholding} 

For the following domains, candidate pools are the corresponding training data used by OLMo~2. 

Pre-training: (i) \textbf{peS2o,} whose candidate member pool is OLMo~2's peS2o training data, \href{https://huggingface.co/datasets/allenai/olmo-mix-1124/blob/main/data/pes2o/pes2o-0000.json.gz}{\texttt{pes2o-0000.json.gz}} through \href{https://huggingface.co/datasets/allenai/olmo-mix-1124/blob/main/data/pes2o/pes2o-0025.json.gz}{\texttt{pes2o-0025.json.gz}}. (ii) \textbf{OpenWebMath,} whose candidate member pool is OLMo~2's OpenWebMath training data, \href{https://huggingface.co/datasets/allenai/olmo-mix-1124/blob/main/data/open-web-math/train/open-web-math-train-0000.json.gz}{\texttt{open-web-math-train-0000.json.gz}} through \href{https://huggingface.co/datasets/allenai/olmo-mix-1124/blob/main/data/open-web-math/train/open-web-math-train-0012.json.gz}{\texttt{open-web-math-train-0012.json.gz}}. 

Mid-training: \textbf{Math Mix,} whose only subset, GSM8K, has a candidate member pool drawn from its corresponding OLMo~2 training data, \href{https://huggingface.co/datasets/allenai/dolmino-mix-1124/blob/main/data/math/gsm8k/main/train/0.jsonl.zst}{\texttt{0.jsonl.zst}}.

Post-training: (i) \textbf{SFT.} Both of SFT's subsets, WildChat and Aya, draw their candidate member pool from the corresponding OLMo~2 training data, filtered by source keyword: \texttt{tulu\_v3.9\_wildchat\_100k} for WildChat and \texttt{tulu\_v3.9\_aya\_100k} for Aya. The 1B and 32B models draw shards 0--5 of \href{https://huggingface.co/datasets/allenai/tulu-3-sft-olmo-2-mixture-0225}{\texttt{tulu-3-sft-olmo-2-mixture-0225}}; the 7B and 13B models draw shards 0--5 of \href{https://huggingface.co/datasets/allenai/tulu-3-sft-olmo-2-mixture}{\texttt{tulu-3-sft-olmo-2-mixture}}. (ii) \textbf{DPO.} For each model size, candidate pools are drawn from OLMo~2's corresponding DPO training data for DPO's only subset, UltraFeedback-WildChat, filtered using a model-specific keyword: 1B draws shards 0--4 of \href{https://huggingface.co/datasets/allenai/olmo-2-0425-1b-preference-mix}{\texttt{olmo-2-0425-1b-preference-mix}} (keyword: \texttt{olmo2-1b-wildchat-unused-on-policy}); 7B draws shards 0--5 of \href{https://huggingface.co/datasets/allenai/olmo-2-1124-7b-preference-mix}{\texttt{olmo-2-1124-7b-preference-mix}} (keyword: \texttt{wildchat\_v3.9\_unused\_on\_policy\_olmo2\_7b}); 13B draws shards 0--5 of \href{https://huggingface.co/datasets/allenai/olmo-2-1124-13b-preference-mix}{\texttt{olmo-2-1124-13b-preference-mix}} (keyword: \texttt{wildchat\_v3.9\_unused\_on\_policy\_olmo2\_13b}); and 32B draws shards 0--4 of \href{https://huggingface.co/datasets/allenai/olmo-2-0325-32b-preference-mix}{\texttt{olmo-2-0325-32b-preference-mix}} (keyword: \texttt{wildchat\_v3.9\_unused\_off\_policy\_prompts-olmo32}). (iii) \textbf{RLVR.} The candidate pools for both the CoT-GSM8K and CoT-MATH subsets of RLVR originate from the single shard of \href{https://huggingface.co/datasets/allenai/RLVR-GSM-MATH-IF-Mixed-Constraints}{\texttt{RLVR-GSM-MATH-IF-Mixed-Constraints}}, filtered using the keywords \texttt{gsm8k} and \texttt{MATH}, respectively.

\subsubsection{Quality-Thresholded Domains} 
OLMo~2's training data in the following domains has systematically higher quality than the corresponding non-members, since only instances that passed OLMo~2's quality threshold were selected for training. To align text quality with non-members, we select training data closest to the threshold as the candidate pool.

\paragraph{DCLM-Baseline.} We restrict the candidate pool to the lowest-scoring instances, those at the 10\% quality threshold with a fastText score of 0.01811, drawn from three shards of OLMo~2's DCLM-Baseline: \texttt{global-shard\_01-local-shard\_0}, \texttt{global-shard\_05-local-shard\_6}, and \texttt{global-shard\_07-local-shard\_3}. This yields a candidate member pool of over 300 instances (580K tokens). 

\textbf{StarCoder.} We restrict the candidate pool to instances whose top-1 frequency lies within 0.28--0.3 or top-2 frequency lies within 0.48--0.5. Regarding GitHub stars, since candidates are drawn from OLMo~2's StarCoder training data, the $\geq 2$ GitHub-star constraint, enforced for non-members, is already satisfied by construction. This yields a candidate member pool of over 8,000 instances (7.3M tokens) for Java and over 5,500 instances (15.9M tokens) for Python.

\textbf{Stack Exchange.} We restrict the candidate pool to instances with question vote $= \text{3}$ or answer vote $= \text{5}$. This yields a candidate member pool of over 300K instances (200M tokens) for Stack Overflow and over 30K instances (19M tokens) for Math Stack Exchange.

\subsection{Alignment Axes}
We align members and non-members on three axes: text quality, temporal range, and lexical similarity, as discussed in Section~\ref{sec4.5}. Each axis is either a signal that existing MIAs are known to exploit or a by-product of OLMo~2's filtering that must therefore be controlled. These three are not exhaustive. Adding more axes is possible but carries a cost: each further constraint makes the simulated-annealing search substantially harder and shrinks the feasible candidate pool, reducing subset sizes and therefore the reliability of the resulting statistical analysis. We thus stop at the three most important axes, prioritizing subset size and statistical reliability over additional constraints.

\subsection{Constrained Sampling via Simulated Annealing}
\label{appendix:simulated_annealing}
This section provides a detailed description of the simulated-annealing procedure used to sample members, introduced in Section~\ref{sec4.5}. The procedure is applied independently to each domain, or to each subset within multi-subset domains, against its own corresponding non-member set.

\paragraph{Setup.} Given a candidate member pool $A$ and a fixed non-member split with token distribution $p_B$, the goal is to select a subset $A' \subseteq A$ satisfying three constraints: (i) $|A'|$ equals the non-member instance count $|B|$; (ii) the total token count of $A'$ is within $\delta = 250$ of the non-member token count; and (iii) the distributional loss $\mathcal{L}$ against the non-member reference is minimized:
\begin{equation}
\mathcal{L} = \mathrm{JSD}(p_{A'}, p_B) + \mathds{1}[\text{temporal metadata available}] \cdot \mathcal{L}_{\text{temp}},
\end{equation}
where $p_{A'}$ is the token distribution of $A'$, and $\mathrm{JSD}$ denotes the Jensen--Shannon divergence, computed with base-2 logarithms so that $\mathrm{JSD}(p_{A'}, p_B) \in [0, 1]$, enforcing lexical alignment. The temporal alignment term takes the form:
\begin{equation}
\mathcal{L}_{\text{temp}} = \frac{|\mu_{A'} - \mu_B| + |\sigma_{A'} - \sigma_B|}{\sigma_B},
\end{equation}
with $\mu$ and $\sigma$ denoting the mean and standard deviation of the temporal axis (in days), normalized by the non-member standard deviation $\sigma_B$.

\paragraph{Algorithm.} We initialize by randomly partitioning $A$ into $(A', \text{leftover})$ such that $|A'|$ matches the target instance count $|B|$. The algorithm then runs for $T_{\max} = 10^7$ steps, with each step proposing a swap between one instance in $A'$ and one in the leftover pool. The procedure operates in two phases. In the \textbf{greedy token-matching phase}, used when the current token count is outside the $\delta$ margin, the swap is accepted only if it brings the token count closer to target. In the \textbf{alignment optimization phase}, used once within margin, the swap is accepted if the post-swap token count remains within margin and the swap satisfies the Metropolis criterion: accepted if $\Delta\mathcal{L} < 0$, otherwise accepted with probability $\exp(-\Delta\mathcal{L}/T)$.

\paragraph{Temperature Schedule.} The temperature $T$ is initialized at $T_0 = 10^{-3}$ and decays geometrically: $T \leftarrow 0.999995 \cdot T$ after each accepted swap in the alignment optimization phase. This schedule provides broad exploration early in the search and progressive convergence as the temperature decreases.

\paragraph{Reproducibility.} All simulated-annealing runs use a fixed random seed of 42, and the resulting subset $A^*$ is retained as the final member split.

\paragraph{Excluded Domains.} Mid-training's FLAN has sufficient candidate members and non-members, but no member and non-member pair can be matched on both instance count and total tokens, even with the token margin relaxed to 1{,}000. We therefore exclude it from \textsc{OLMo-Detect}.

\begin{table*}[t]
\centering
\resizebox{\textwidth}{!}{%
\begin{tabular}{@{}l cccc cc ccc c@{}}
\toprule
& \multicolumn{4}{c}{\textbf{Pre-training}} & \multicolumn{2}{c}{\textbf{Mid-training}} & \multicolumn{3}{c}{\textbf{Post-training}} & \\
\cmidrule(lr){2-5} \cmidrule(lr){6-7} \cmidrule(lr){8-10}
& DCLM-Baseline & OpenWebMath & peS2o & StarCoder & Math Mix & Stack Exchange & SFT & DPO & RLVR & \textbf{Avg} \\
\midrule
\textbf{1B} & 4.07$_{\pm4.87}$ & 8.33$_{\pm5.69}$ & 3.11$_{\pm4.01}$ & 9.31$_{\pm5.24}$ & 6.65$_{\pm4.77}$ & 5.71$_{\pm5.34}$ & 6.08$_{\pm5.24}$ & 2.12$_{\pm3.01}$ & 2.06$_{\pm4.55}$ & 5.27 \\
\textbf{7B} & 4.07$_{\pm4.87}$ & 8.33$_{\pm5.69}$ & 3.11$_{\pm4.01}$ & 9.31$_{\pm5.24}$ & 6.66$_{\pm4.75}$ & 5.71$_{\pm5.34}$ & 6.54$_{\pm5.29}$ & 1.98$_{\pm3.05}$ & 2.06$_{\pm4.55}$ & 5.31 \\
\textbf{13B} & 4.07$_{\pm4.87}$ & 8.33$_{\pm5.69}$ & 3.11$_{\pm4.01}$ & 9.31$_{\pm5.24}$ & 6.65$_{\pm4.75}$ & 5.71$_{\pm5.34}$ & 6.55$_{\pm5.30}$ & 2.52$_{\pm3.37}$ & 2.06$_{\pm4.55}$ & 5.37 \\
\textbf{32B} & 4.07$_{\pm4.87}$ & 8.33$_{\pm5.69}$ & 3.11$_{\pm4.01}$ & 9.31$_{\pm5.24}$ & 6.64$_{\pm4.75}$ & 5.71$_{\pm5.35}$ & 6.12$_{\pm5.31}$ & 2.20$_{\pm3.15}$ & 2.06$_{\pm4.55}$ & 5.28 \\
\midrule
\textbf{Avg} & 4.07 & 8.33 & 3.11 & 9.31 & 6.65 & 5.71 & 6.32 & 2.20 & 2.06 & \textbf{5.31} \\
\bottomrule
\end{tabular}%
}
\caption{Per-model-size 13-gram overlap $\eta$ (\%; mean$_{\pm\text{std}}$ over non-members) for each domain against the training corpus of the corresponding OLMo~2 model size. Multi-subset domains are macro-averaged across subsets (Table~\ref{table:13_gram_overlap_subsets}), with the domain-level standard deviation computed as the root mean square of the subset-level standard deviations.}
\label{table:13_gram_overlap_model_size}
\end{table*}

\begin{table*}[t]
\centering
\resizebox{0.85\textwidth}{!}{%
\begin{tabular}{@{}l l cccc c@{}}
\toprule
\textbf{Domain} & \textbf{Subset} & \textbf{1B} & \textbf{7B} & \textbf{13B} & \textbf{32B} & \textbf{Avg} \\
\midrule
\multirow{2}{*}{StarCoder} & Python & 7.17$_{\pm5.28}$ & 7.17$_{\pm5.28}$ & 7.17$_{\pm5.28}$ & 7.17$_{\pm5.28}$ & 7.17 \\
 & Java & 11.45$_{\pm5.20}$ & 11.45$_{\pm5.20}$ & 11.45$_{\pm5.20}$ & 11.45$_{\pm5.20}$ & 11.45 \\
\cmidrule(l){1-7}
\multirow{2}{*}{Stack Exchange} & Math Stack Exchange & 5.67$_{\pm4.91}$ & 5.67$_{\pm4.91}$ & 5.67$_{\pm4.91}$ & 5.68$_{\pm4.92}$ & 5.67 \\
 & Stack Overflow & 5.74$_{\pm5.74}$ & 5.74$_{\pm5.74}$ & 5.74$_{\pm5.75}$ & 5.75$_{\pm5.74}$ & 5.74 \\
\cmidrule(l){1-7}
\multirow{2}{*}{SFT} & WildChat & 5.16$_{\pm4.49}$ & 4.53$_{\pm4.53}$ & 4.55$_{\pm4.53}$ & 5.14$_{\pm4.59}$ & 4.84 \\
 & Aya & 7.01$_{\pm5.89}$ & 8.55$_{\pm5.95}$ & 8.55$_{\pm5.96}$ & 7.10$_{\pm5.95}$ & 7.80 \\
\cmidrule(l){1-7}
\multirow{3}{*}{DPO} & Prompt & 0.00$_{\pm0.00}$ & 0.13$_{\pm1.04}$ & 0.00$_{\pm0.00}$ & 0.01$_{\pm0.11}$ & 0.04 \\
 & Chosen & 3.55$_{\pm3.58}$ & 3.06$_{\pm3.46}$ & 4.20$_{\pm4.47}$ & 3.73$_{\pm4.07}$ & 3.63 \\
 & Rejected & 2.80$_{\pm3.79}$ & 2.75$_{\pm3.84}$ & 3.37$_{\pm3.75}$ & 2.85$_{\pm3.63}$ & 2.94 \\
\cmidrule(l){1-7}
\multirow{2}{*}{RLVR} & CoT-GSM8K & 1.45$_{\pm3.77}$ & 1.45$_{\pm3.77}$ & 1.45$_{\pm3.77}$ & 1.45$_{\pm3.77}$ & 1.45 \\
 & CoT-MATH & 2.67$_{\pm5.20}$ & 2.67$_{\pm5.20}$ & 2.67$_{\pm5.20}$ & 2.67$_{\pm5.20}$ & 2.67 \\
\bottomrule
\end{tabular}%
}
\caption{Per-subset 13-gram overlap $\eta$ (\%; mean$_{\pm\text{std}}$ over non-members) for multi-subset domains. Domain-level values are macro-averaged across subsets; for DPO, they are averaged across the prompt, chosen, and rejected fields.}
\label{table:13_gram_overlap_subsets}
\end{table*}

\begin{table*}[t]
\centering
\resizebox{\textwidth}{!}{%
\begin{tabular}{@{}l cccc cc ccc c@{}}
\toprule
& \multicolumn{4}{c}{\textbf{Pre-training}} & \multicolumn{2}{c}{\textbf{Mid-training}} & \multicolumn{3}{c}{\textbf{Post-training}} & \\
\cmidrule(lr){2-5} \cmidrule(lr){6-7} \cmidrule(lr){8-10}
& DCLM-Baseline & OpenWebMath & peS2o & StarCoder & Math Mix & Stack Exchange & SFT & DPO & RLVR & \textbf{Avg} \\
\midrule
\textbf{1B} & 0.49 & 0.52 & 0.48 & 0.50 & 0.52 & 0.48 & 0.51 & 0.47 & 0.51 & 0.50 \\
\textbf{7B} & 0.49 & 0.52 & 0.48 & 0.50 & 0.52 & 0.48 & 0.52 & 0.51 & 0.51 & 0.50 \\
\textbf{13B} & 0.49 & 0.52 & 0.48 & 0.50 & 0.52 & 0.48 & 0.52 & 0.49 & 0.51 & 0.50 \\
\textbf{32B} & 0.49 & 0.52 & 0.48 & 0.50 & 0.52 & 0.48 & 0.51 & 0.48 & 0.51 & 0.50 \\
\midrule
\textbf{Avg} & 0.49 & 0.52 & 0.48 & 0.50 & 0.52 & 0.48 & 0.51 & 0.49 & 0.51 & \textbf{0.50} \\
\bottomrule
\end{tabular}%
}
\caption{Per-model-size blind detectability, measured by BoW classifier AUC. Multi-subset domains are macro-averaged across subsets, with per-subset AUCs reported in Table~\ref{table:blind_auc_subsets}.}
\label{table:blind_auc_model_size}
\end{table*}

\begin{table*}[t]
\centering
\resizebox{0.8\textwidth}{!}{%
\begin{tabular}{@{}l l cccc c@{}}
\toprule
\textbf{Domain} & \textbf{Subset} & \textbf{1B} & \textbf{7B} & \textbf{13B} & \textbf{32B} & \textbf{Avg} \\
\midrule
\multirow{2}{*}{StarCoder} & Python & 0.48 & 0.48 & 0.48 & 0.48 & 0.48 \\
 & Java & 0.51 & 0.51 & 0.51 & 0.51 & 0.51 \\
\cmidrule(l){2-7}
\multirow{2}{*}{Stack Exchange} & Math Stack Exchange & 0.47 & 0.47 & 0.47 & 0.47 & 0.47 \\
 & Stack Overflow & 0.49 & 0.49 & 0.49 & 0.49 & 0.49 \\
\cmidrule(l){2-7}
\multirow{2}{*}{SFT} & WildChat & 0.52 & 0.54 & 0.54 & 0.52 & 0.53 \\
 & Aya & 0.50 & 0.51 & 0.51 & 0.50 & 0.50 \\
\cmidrule(l){2-7}
\multirow{2}{*}{RLVR} & CoT-GSM8K & 0.51 & 0.51 & 0.51 & 0.51 & 0.51 \\
 & CoT-MATH & 0.51 & 0.51 & 0.51 & 0.51 & 0.51 \\
\bottomrule
\end{tabular}%
}
\caption{Per-subset blind detectability for multi-subset domains, measured by BoW classifier AUC.}
\label{table:blind_auc_subsets}
\end{table*}

\section{Detailed Per-Domain Statistics of OLMo-Detect (Shifted)}
\label{appendix:olmo_detect_shifted}

This appendix provides detailed statistics of \textsc{OLMo-Detect (Shifted)}. Overall statistics are reported in Table~\ref{appendix:table:olmo_detect_shifted:overall_statistics}; detailed per-domain statistics are reported in Tables~\ref{table:app:shifted:pre}--\ref{table:app:shifted:post}; per-subset statistics for multi-subset domains, StarCoder, Stack Exchange, and SFT, are reported in Tables~\ref{table:app:shifted:starcoder}--\ref{table:app:shifted:sft}. Math Mix and RLVR are excluded, since random sampling does not yield members with meaningful distribution shifts in these domains.

\paragraph{Analysis of Lexical Shift Across Domains.} In \textsc{OLMo-Detect (Shifted)}, members are drawn at random rather than perturbed toward a target, so each domain shifts only along the axes on which its source corpus actually varies. Web pages, academic papers, and Q\&A posts remain lexically uniform across quality tiers and publication years, so DCLM-Baseline, peS2o, and Stack Exchange shift on quality and temporal range instead: DCLM-Baseline's quality score rises from 0.018 to 0.131, peS2o's mean date moves back by nearly nine years, and Stack Exchange's question and answer votes roughly double, while their lexical values change by at most 0.03. By contrast, StarCoder, OpenWebMath, SFT, and DPO draw on corpora that vary lexically, through repository-specific identifiers, heterogeneous web pages, and multi-task multilingual data, and their lexical values rise two- to fourfold, from 0.09--0.15 to 0.32--0.36. In DPO, for instance, the dominant non-English script changes from Chinese to Cyrillic. Every domain is therefore shifted on at least one axis, and the shift is deliberately not maximized: \textbf{it emulates the unaligned construction of existing benchmarks rather than a worst case.}

\section{Detailed Descriptions of Experimental Settings}
\label{appendix:experiment_setting}

\begin{table*}[t]
\centering
\resizebox{\textwidth}{!}{%
\begin{tabular}{@{}lll@{}}
\toprule
Method & Reference & Repository \\
\midrule
PPL & \citet{yeom2018privacy} & --- (standard baseline) \\
Zlib        & \citet{carlini2021extracting} & --- (standard baseline) \\
Lower   & \citet{carlini2021extracting} & --- (standard baseline) \\
MinK      & \citet{shi2023detecting}      & \url{https://github.com/swj0419/detect-pretrain-code} \\
MinK++         & \citet{zhang2024min}          & \url{https://github.com/zjysteven/mink-plus-plus} \\
DCPDD            & \citet{zhang2024pretraining}  & \url{https://github.com/zhang-wei-chao/DC-PDD} \\
ReCaLL            & \citet{xie2024recall}         & \url{https://github.com/ruoyuxie/recall} \\
CAMIA             & \citet{chang2025context}      & \url{https://github.com/changhongyan123/context_aware_mia} \\
PAC               & \citet{ye2024data}            & \url{https://github.com/yyy01/PAC} \\
EMMIA             & \citet{kim2026detecting}      & \url{https://github.com/gyuwankim/em-mia} \\
DCQ               & \citet{golchin2025data}       & \url{https://github.com/shahriargolchin/DCQ} \\
Neigh. & \citet{mattern2023membership} & \url{https://github.com/mireshghallah/neighborhood-curvature-mia} \\
CDD               & \citet{dong2024generalization} & \url{https://github.com/YihongDong/CDD-TED4LLMs} \\
Guided & \citet{golchin2023time}       & \url{https://github.com/shahriargolchin/time-travel-in-llms} \\
SelfCrit     & \citet{tao2025detecting}      & \url{https://github.com/yongding-tao/RL-Data-Contamination} \\
\midrule
FSD      & \citet{zhang2025fine}   & \url{https://github.com/ml-stat-Sustech/Fine-tuned-Score-Deviation} \\
MIA-Tuner & \citet{fu2025mia} & \url{https://github.com/tsinghua-fib-lab/AAAI2025_MIA-Tuner} \\
Sup-CAMIA  & \citet{chang2025context} & \url{https://github.com/changhongyan123/context_aware_mia} \\
\bottomrule
\end{tabular}%
}
\caption{Public repositories followed by our implementations of the compared MIAs. The final three rows are the supervised methods; all others are unsupervised.}
\label{tab:method_repos}
\end{table*}

\begin{table*}[t]
\centering
\resizebox{\textwidth}{!}{%
\begin{tabular}{@{}lll cccc@{}}
\toprule
Method & Hyperparameter & Search space & \multicolumn{4}{c}{Selected value} \\
\cmidrule(lr){4-7}
 & & & 1B & 7B & 13B & 32B \\
\midrule
MinK & $k$ & $\{0.05,\dots,1.0\}$ & \multicolumn{4}{c}{$0.2$} \\
MinK++    & $k$ & $\{0.05,\dots,1.0\}$ & \multicolumn{4}{c}{$0.2$} \\
DCPDD       & $\alpha$ & $\{0.001,\dots,0.5\}$ & $0.001$ & $0.2$ & $0.05$ & $0.2$ \\
ReCaLL       & \# prefix shots $n$ & $\{1,\dots,12\}$ & $10$ & $3$ & $7$ & $12$ \\
CAMIA        & $p$-value composition & \{Edgington, Fisher, George\} & George & Fisher & George & George \\
PAC          & $(k_1,k_2)$ & $k_1\!\in\!\{.05,.1,.2\},\ k_2\!\in\!\{.1,.2,.3,.4\}$ & \multicolumn{4}{c}{$(0.05,\,0.10)$} \\
DCQ          & score type & \{mean, min, $\kappa$, $\kappa_{\mathrm{np}}$\}$P_{\text{correct}}$ & min & min & $\kappa_{\mathrm{np}}$ & $\kappa$ \\
Neigh. & score type & \{$d$, $z$\} & \multicolumn{4}{c}{$d$} \\
CDD          & $(\alpha, L, r)$ & $\alpha\!\in\!\{0,\dots,.3\}, L\!\in\!\{50,100,200\}, r\!\in\!\{.25,.5,.75\}$ & $(.025,50,.5)$ & $(.075,50,.5)$ & $(.15,50,.25)$ & $(.3,50,.25)$ \\
Guided & score type & \{ROUGE-guided, ROUGE-$\Delta$\} & ROUGE-$\Delta$ & ROUGE-$\Delta$ & ROUGE-$\Delta$ & ROUGE-guided \\
\bottomrule
\end{tabular}%
}
\caption{Tunable hyperparameters, search spaces, and the values selected by development-set AUC on \textsc{OLMo-Detect}.}
\label{tab:method_hparams}
\end{table*}

\paragraph{Implementation Details.}
We implement every compared method from its original public repository (Table~\ref{tab:method_repos}). To ensure a fair comparison, every method is evaluated under an identical protocol with a batch size of $1$ and a context length of $4{,}096$ tokens for the generative methods that require text generation (Neigh., Guided, DCQ, CDD, and SelfCrit) and $1{,}024$ tokens for the remaining non-generative methods.
All experiments are run on NVIDIA A100 (80\,GB) GPUs, using two GPUs for the 32B model
and a single GPU for 1B to 13B models. Method-specific hyperparameters, where applicable, are tuned on the held-out development set; the search spaces and selected values are given in Table~\ref{tab:method_hparams}. For reproducibility, all experiments use a fixed random seed of $42$.

\paragraph{Determinism and Statistical Reporting.}
We use a single random seed of $42$ because the majority of our evaluated methods are deterministic. All non-generative unsupervised methods (PPL, Zlib, Lower, MinK, MinK++, DCPDD, ReCaLL, CAMIA, EMMIA) compute their statistics directly from token log-probabilities, without any sampling. Among the remaining methods: the two generative unsupervised methods, Guided and SelfCrit, decode greedily (\texttt{do\_sample}=\texttt{False}, temperature $0$). SelfCrit's two responses are not sampled but produced from two different prompts, an initial answer and a follow-up requesting an alternative solution, so greedy decoding yields two distinct yet deterministic outputs. DCQ scores the target model with multiple-choice log-probabilities, which are likewise deterministic; its quiz options are generated once by an external \texttt{gpt-4o-mini} judge and cached on disk, shared across all model sizes and runs, so the identical quiz is reused on every run. Sup-CAMIA fits a convex logistic-regression classifier on deterministic features. For all of these methods, the seed has no effect on the reported scores.

The seed matters only for the methods that draw stochastic samples or train with random data ordering: Neigh.\ (neighbors sampled from a T5 mask-filling model), CDD (completions sampled from the target model at temperature $0.8$), PAC (word-level random-swap augmentation), and the two supervised fine-tuning methods FSD (a LoRA adapter) and MIA-Tuner (a soft prompt). Due to the computational cost of repeated runs, none of these works reports results across multiple seeds in their papers; we follow this convention and fix both the framework seed ($42$) and each method's own internal seed parameter where one exists (e.g., PAC's and Neigh.'s), holding their random draws, augmentations, and training runs fixed across runs.

Since per-instance scores are fixed under a single seed, we quantify the uncertainty from the finite evaluation set with a nonparametric bootstrap. We resample members and non-members with replacement within every subset, recompute the full subset-to-domain-to-stage aggregation on each of $2{,}000$ resamples, and report the $2.5$th and $97.5$th percentiles as the $95\%$ confidence interval, so the interval applies to the overall AUC rather than to any single evaluation. This captures uncertainty from the finite evaluation set but not seed-to-seed variance for the five stochastic methods above; we leave a full multi-seed analysis to future work. Intervals are given in Tables~\ref{table:ci_overall_unsup} and~\ref{table:ci_overall_sup}.

\begin{table}[t]
\centering
\begin{tabular}{lcc}
\toprule
\textbf{Method} & \textbf{AUC} & \textbf{95\% CI} \\
\midrule
PPL & 0.65 & [0.64, 0.67] \\
Zlib & 0.68 & [0.67, 0.70] \\
Lower & 0.63 & [0.62, 0.65] \\
MinK & 0.66 & [0.64, 0.67] \\
MinK++ & 0.56 & [0.54, 0.57] \\
DCPDD & 0.67 & [0.65, 0.68] \\
PAC & 0.66 & [0.64, 0.67] \\
ReCaLL & 0.46 & [0.45, 0.47] \\
EMMIA & 0.57 & [0.56, 0.58] \\
CAMIA & 0.65 & [0.64, 0.67] \\
Neigh. & 0.59 & [0.58, 0.61] \\
Guided & 0.49 & [0.48, 0.50] \\
DCQ & 0.49 & [0.48, 0.51] \\
CDD & 0.52 & [0.52, 0.53] \\
\bottomrule
\end{tabular}
\caption{Bootstrap $95\%$ confidence intervals for the overall AUC of each unsupervised MIA on \textsc{OLMo-Detect}, i.e.\ the \textbf{\textit{All}} column of Table~\ref{table:olmo_detect_overall_unsupervised} averaged over model sizes. Members and non-members are resampled with replacement within every subset and the full subset-to-domain-to-stage aggregation is recomputed on each of $2{,}000$ resamples, so the interval applies to the aggregate rather than to any single evaluation. SelfCrit is omitted because it covers only post-training. Point estimates match Table~\ref{table:olmo_detect_overall_unsupervised}.}
\label{table:ci_overall_unsup}
\end{table}

\begin{table}[t]
\centering
\begin{tabular}{llcc}
\toprule
\textbf{Method} & \textbf{Setting} & \textbf{AUC} & \textbf{95\% CI} \\
\midrule
Sup-CAMIA & In-domain & 0.68 & [0.67, 0.68] \\
Sup-CAMIA & LODO & 0.61 & [0.60, 0.62] \\
MIA-Tuner & In-domain & 0.55 & [0.54, 0.56] \\
MIA-Tuner & LODO & 0.49 & [0.48, 0.50] \\
\bottomrule
\end{tabular}
\caption{Bootstrap $95\%$ confidence intervals for the overall AUC of the supervised MIAs on \textsc{OLMo-Detect}, i.e.\ the \textbf{\textit{All}} column of Table~\ref{table:olmo_detect_overall_supervised} averaged over model sizes, computed as in Table~\ref{table:ci_overall_unsup}. FSD is omitted because it is reported as a gain over its base detector rather than as an absolute AUC. Point estimates match Table~\ref{table:olmo_detect_overall_supervised}.}
\label{table:ci_overall_sup}
\end{table}

\paragraph{Continuous Scoring for Categorical Methods.} 
DCQ \citep{golchin2025data} and Guided \citep{golchin2023time} originally produce categorical outputs that are not directly compatible with AUC computation. For DCQ, we replace the original single-letter output with the first-token logprobs over $\{A, B, C, D, E\}$ from a single forward pass on the model under evaluation. Each instance is presented through four bias compensator quiz (BCQ) prompts, with the original placed at slots $L \in \{A, B, C, D\}$ respectively; the per-slot score $P(L \mid \mathrm{BCQ}_L)$ is the model's recognition probability for that placement. The final per-instance score is the mean (or minimum) of these four probabilities, with the choice tuned per (model, dataset) on the dev set. For Guided, we generate two greedy completions per instance: a \emph{guided} completion (prompted with the source corpus name) and a \emph{general} completion (no source cited). The per-instance score is the ROUGE-L F1 between the guided completion and the held-out second piece of the original paragraph, or its difference against the general completion; the variant is selected per (model, dataset) on the dev set. Both adaptations preserve the original methods' computations and use the same underlying signals (next-token logprobs for DCQ; ROUGE-L F1, the paper's own metric, for Guided); only the post-processing step that maps these signals to a categorical decision is replaced with continuous score extraction.

\paragraph{Excluded MIAs.} Among existing MIAs that we have surveyed, four are excluded due to setting incompatibility: (i) methods that require access to a model other than $\mathcal{M}$, whether an existing smaller model or one separately fine-tuned, including Smaller Ref \citep{carlini2021extracting} and SPV-MIA \citep{fu2024membership}, since our setting assumes only grey-box access to $\mathcal{M}$, optionally extended to fine-tuning $\mathcal{M}$ itself, not the construction of an additional model; (ii) narrowly focused methods, such as Name-cloze \citep{chang2023speak}, since it requires text containing proper nouns to mask; (iii) methods with no open-source implementation, such as TS-Guessing \citep{deng2024investigating}.

\begin{table*}[t]
\centering
\resizebox{0.98\textwidth}{!}{%
\begin{tabular}{@{}cllllrrlll@{}}
\toprule
\textbf{Stage} & \textbf{Domain} & \textbf{Subsets} & \textbf{Split} & \textbf{Samples} & \textbf{Tokens} & \textbf{Quality} & \textbf{Temporal} & \textbf{Lexical} \\
\midrule
\multirow{10}{*}{Pre-training}
& \multirow{2}{*}{DCLM-Baseline} & 
  & Non-member & 150 & 217{,}117 & FastText: 0.01809 $\pm$ 0.00007 & 2018-01-19 $\pm$ 930d & \multirow{2}{*}{0.08} \\
& & & Member & 150 & 217{,}295 & FastText: 0.13129 $\pm$ 0.18279 & 2017-02-21 $\pm$ 1208d & \\
\cmidrule(l){2-9}
& \multirow{2}{*}{peS2o} & 
  & Non-member & 268 & 216{,}981 & --- & 2022-12-14 $\pm$ 11d & \multirow{2}{*}{0.13} \\
& & & Member & 268 & 217{,}039 & --- & 2014-02-03 $\pm$ 2657d & \\
\cmidrule(l){2-9}
& \multirow{2}{*}{OpenWebMath} & 
  & Non-member & 151 & 217{,}183 & --- & 2019-10-13 $\pm$ 853d & \multirow{2}{*}{0.32} \\
& & & Member & 151 & 217{,}379 & --- & 2011-10-01 $\pm$ 2936d & \\
\cmidrule(l){2-9}
& \multirow{4}{*}{StarCoder} & \multirow{4}{*}{\makecell[l]{Python,\\Java}}
  & Non-member & 156 & 217{,}156 & \begin{tabular}[c]{@{}l@{\hspace{4pt}}r@{\hspace{2pt}}l@{}}
      GitHub Stars: & 35.65 & $\pm$ 66.66 \\
      Top-1 Freq:   & 0.30 & $\pm$ 0.02 \\
      Top-2 Freq:   & 0.38 & $\pm$ 0.05
    \end{tabular} & --- & \multirow{4}{*}{0.36} \\
& & & Member & 156 & 217{,}284 & \begin{tabular}[c]{@{}l@{\hspace{4pt}}r@{\hspace{2pt}}l@{}}
      GitHub Stars: & 10.77 & $\pm$ 17.94 \\
      Top-1 Freq:   & 0.09 & $\pm$ 0.03 \\
      Top-2 Freq:   & 0.15 & $\pm$ 0.04
    \end{tabular} & --- & \\
\midrule
\multirow{3}{*}{Mid-training}
& \multirow{3}{*}{Stack Exchange} & \multirow{3}{*}{\makecell[l]{Stack Overflow,\\Math Stack Exchange}}
  & Non-member & 430 & 217{,}208 & \begin{tabular}[c]{@{}l@{\hspace{4pt}}l@{}}
      Question Vote: & 3.45 $\pm$ 2.19 \\
      Answer Vote:   & 5.12 $\pm$ 1.99
    \end{tabular} & 2015-03-31 $\pm$ 1378d & \multirow{3}{*}{0.08} \\
& & & Member & 430 & 217{,}209 & \begin{tabular}[c]{@{}l@{\hspace{4pt}}l@{}}
      Question Vote: & 8.12 $\pm$ 7.08 \\
      Answer Vote:   & 10.02 $\pm$ 6.37
    \end{tabular} & 2013-05-31 $\pm$ 1213d & \\
\midrule
\multirow{4}{*}{Post-training}
& \multirow{2}{*}{SFT$^{\dagger}$} & \multirow{2}{*}{\makecell[l]{WildChat,\\Aya}}
  & Non-member & 225 & 217{,}057 & --- & --- & \multirow{2}{*}{0.34} \\
& & & Member & 225 & 217{,}238 & --- & --- & \\
\cmidrule(l){2-9}
& \multirow{2}{*}{DPO$^{\dagger}$} & \multirow{2}{*}{\makecell[l]{UF-WildChat}}
  & Non-member & 113 & 217{,}045 & --- & --- & \multirow{2}{*}{0.32} \\
& & & Member & 113 & 217{,}154 & --- & --- & \\
\bottomrule
\end{tabular}%
}
\caption{Overall statistics of \textsc{OLMo-Detect (Shifted)}, spanning seven of the nine domains of \textsc{OLMo-Detect} (GSM8K and RLVR are excluded, as neither has any shiftable axis). Non-members are unchanged from \textsc{OLMo-Detect}, while members are randomly sampled subject only to matched instance and token counts, leaving Text \textbf{Quality}, \textbf{Temporal} Range, and \textbf{Lexical} Similarity unaligned. ``---'' denotes an axis that does not apply to a given domain, i.e., no quality thresholding is applied, or no timestamp metadata exists. \textbf{NM}: non-member; \textbf{M}: member. The \textbf{Subsets} column lists constituent subsets; blank entries indicate single-subset domains whose subset shares the domain's name. $^{\dagger}$SFT and DPO are constructed separately for each model size because their training data differ across sizes; the reported values correspond to the 1B model.}
\label{appendix:table:olmo_detect_shifted:overall_statistics}
\end{table*}

\begin{table*}[h]
\centering
\resizebox{\textwidth}{!}{%
%
}
\caption{Detailed per-domain statistics for the pre-training stage of \textsc{OLMo-Detect (Shifted)}. \textbf{NM}: non-member; \textbf{M}: member. \textbf{Dev}: development set; \textbf{Test}: test set. The \textbf{Subset} column lists constituent subsets; blank entries indicate single-subset domains whose subset shares the domain's name.}
\label{table:app:shifted:pre}
\end{table*}

\begin{table*}[h]
\centering
\resizebox{\textwidth}{!}{%
%
%
}
\caption{Detailed per-domain statistics for the mid-training stage of \textsc{OLMo-Detect (Shifted)}. \textbf{NM}: non-member; \textbf{M}: member. \textbf{Dev}: development set; \textbf{Test}: test set. The \textbf{Subset} column lists constituent subsets.}
\label{table:app:shifted:mid}
\end{table*}

\begin{table*}[t]
\centering
\resizebox{0.93\textwidth}{!}{%
\begin{tabular}{@{}c l l l l l r r l l l@{}}
\toprule
\textbf{Stage} & \textbf{Domain} & \textbf{Subset} & \textbf{Model} & \textbf{Split} & \textbf{Partition} & \textbf{Samples} & \textbf{Tokens} & \textbf{Quality} & \textbf{Temporal} & \textbf{Lexical} \\ \midrule
\multirow{44}{*}{Post-training}
& \multirow{14.5}{*}{SFT}
& \multirow{14.5}{*}{\makecell[l]{WildChat,\\Aya}}
& \multirow{7}{*}{1B, 32B}
& \multirow{4}{*}{NM}
& Dev   & 21    & 21,697   & --- & --- & \multirow{7}{*}{0.34} \\
& & & & & Test  & 204   & 195,360  & --- & --- &  \\ \cmidrule(l){6-10}
& & & & & \textbf{Total} & \textbf{225} & \textbf{217,057} & & & \\ \cmidrule(l){5-10}
& & & & \multirow{3}{*}{M}
& Dev   & 21    & 21,727   & --- & --- &  \\
& & & & & Test  & 204   & 195,511  & --- & --- &  \\ \cmidrule(l){6-10}
& & & & & \textbf{Total} & \textbf{225} & \textbf{217,238} & & & \\ \cmidrule(l){4-11}
& & & \multirow{7}{*}{7B, 13B}
& \multirow{4}{*}{NM}
& Dev   & 16    & 21,959   & --- & --- & \multirow{7}{*}{0.34} \\
\bottomrule
\end{tabular}%
}
\caption{Detailed per-domain statistics for the post-training stage of \textsc{OLMo-Detect (Shifted)}. \textbf{NM}: non-member; \textbf{M}: member. \textbf{Dev}: development set; \textbf{Test}: test set. The \textbf{Subset} column lists constituent subsets.}
\label{table:app:shifted:post}
\end{table*}

\begin{table*}[h]
\centering
\resizebox{\textwidth}{!}{%
%
%
}
\caption{Detailed per-subset statistics for StarCoder in \textsc{OLMo-Detect (Shifted)}. \textbf{NM}: non-member; \textbf{M}: member. \textbf{Dev}: development set; \textbf{Test}: test set.}
\label{table:app:shifted:starcoder}
\end{table*}

\begin{table*}[h]
\centering
\resizebox{\textwidth}{!}{%
%
%
}
\caption{Detailed per-subset statistics for Stack Exchange in \textsc{OLMo-Detect (Shifted)}. \textbf{NM}: non-member; \textbf{M}: member. \textbf{Dev}: development set; \textbf{Test}: test set.}
\label{table:app:shifted:stackexchange}
\end{table*}

\begin{table*}[t]
\centering
\resizebox{\textwidth}{!}{%
\begin{tabular}{@{}c l l l l l r r l l l@{}}
\toprule
\textbf{Stage} & \textbf{Domain} & \textbf{Model} & \textbf{Subset} & \textbf{Split} & \textbf{Partition} & \textbf{Samples} & \textbf{Tokens} & \textbf{Quality} & \textbf{Temporal} & \textbf{Lexical} \\ \midrule
\multirow{30.5}{*}{Post-training}
& \multirow{30.5}{*}{SFT}
& \multirow{15}{*}{1B, 32B}
& \multirow{7}{*}{Wildchat}
& \multirow{3}{*}{NM}
& Dev & 7 & 16,404 & --- & --- & \multirow{7}{*}{0.43} \\
& & & & & Test & 64 & 148,039 & --- & --- & \\ \cmidrule(l){6-10}
& & & & & \textbf{Total} & \textbf{71} & \textbf{164,443} & & & \\ \cmidrule(l){5-10}
& & & & \multirow{3}{*}{M}
& Dev & 7 & 16,388 & --- & --- & \\
& & & & & Test & 64 & 148,120 & --- & --- & \\ \cmidrule(l){6-10}
& & & & & \textbf{Total} & \textbf{71} & \textbf{164,508} & & & \\ \cmidrule(l){4-11}
& & & \multirow{7}{*}{Aya}
& \multirow{3}{*}{NM}
& Dev & 14 & 5,293 & --- & --- & \multirow{7}{*}{0.39} \\
\bottomrule
\end{tabular}%
}
\caption{Detailed per-subset statistics for SFT in \textsc{OLMo-Detect (Shifted)}. \textbf{NM}: non-member; \textbf{M}: member. \textbf{Dev}: development set; \textbf{Test}: test set.}
\label{table:app:shifted:sft}
\end{table*}

\section{Per-Domain AUC Results for Unsupervised MIAs on \textsc{OLMo-Detect}}
\label{appendix:auc_results_unsup}

This appendix provides detailed AUC results for unsupervised MIAs on \textsc{OLMo-Detect} for each domain, complementing the stage-level aggregates in Table~\ref{table:olmo_detect_overall_unsupervised}. Per-domain AUC results are shown in Tables~\ref{table:auc_dclm}--\ref{table:auc_rlvr_subsets}.

\begin{table*}[t]
\centering
\resizebox{1\textwidth}{!}{%
\begin{tabular}{@{}l@{\;\;\;}c@{\;\;}c@{\;\;}c@{\;\;}c@{\;\;}c@{\;\;}c@{\;\;}c@{\;\;}c@{\;\;}c@{\;\;}c@{\;\;}c@{\;\;}c@{\;\;}c@{\;\;}c@{\;\;\;}c@{}}
\toprule
\multirow{2}{*}{\textbf{Model}} & \multicolumn{10}{c}{\textbf{Likelihood-based}} & \textbf{Pert.} & \multicolumn{2}{c}{\textbf{Prompt}} & \textbf{Out.-Dist.} & \multirow{2}{*}{\makecell{\textbf{Method-}\\\textbf{Avg}}} \\
\cmidrule(lr){2-11} \cmidrule(lr){12-12} \cmidrule(lr){13-14} \cmidrule(lr){15-15}
 & \textbf{PPL} & \textbf{Zlib} & \textbf{Lower} & \textbf{MinK} & \textbf{MinK++} & \textbf{DCPDD} & \textbf{PAC} & \textbf{ReCaLL} & \textbf{EMMIA} & \textbf{CAMIA} & \textbf{Neigh.} & \textbf{Guided} & \textbf{DCQ} & \textbf{CDD} & \\
\midrule
\textbf{1B} & 0.60 & 0.59 & 0.60 & 0.57 & 0.46 & 0.57 & 0.61 & 0.46 & 0.52 & 0.60 & \textbf{0.64} & 0.54 & 0.50 & 0.50 & 0.55 \\
\textbf{7B} & 0.64 & 0.64 & 0.62 & 0.61 & 0.53 & 0.64 & 0.65 & 0.50 & 0.54 & \textbf{0.65} & 0.61 & 0.49 & 0.43 & 0.49 & 0.57 \\
\textbf{13B} & 0.65 & 0.65 & 0.63 & 0.63 & 0.52 & 0.64 & \textbf{0.68} & 0.62 & 0.53 & 0.66 & 0.62 & 0.47 & 0.47 & 0.51 & 0.59 \\
\textbf{32B} & 0.66 & 0.67 & 0.66 & 0.64 & 0.61 & 0.65 & \textbf{0.69} & 0.53 & 0.50 & 0.64 & 0.61 & 0.47 & 0.64 & 0.51 & 0.61 \\
\cmidrule(l){2-16}
\textbf{Model-Avg} & 0.64 & 0.64 & 0.63 & 0.61 & 0.53 & 0.62 & \textbf{0.66} & 0.53 & 0.52 & 0.64 & 0.62 & 0.49 & 0.51 & 0.50 & 0.58 \\
\bottomrule
\end{tabular}%
}
\caption{Comparison of unsupervised MIAs on the DCLM-Baseline domain of \textsc{OLMo-Detect}, evaluated via AUC. Methods are grouped by family as in Table~\ref{table:olmo_detect_overall_unsupervised}. \textbf{Model-Avg} averages across model sizes, while \textbf{Method-Avg} averages the methods with a value in that row. The best-performing method in each row is highlighted in bold.}
\label{table:auc_dclm}
\end{table*}

\begin{table*}[t]
\centering
\resizebox{1\textwidth}{!}{%
\begin{tabular}{@{}l@{\;\;\;}c@{\;\;}c@{\;\;}c@{\;\;}c@{\;\;}c@{\;\;}c@{\;\;}c@{\;\;}c@{\;\;}c@{\;\;}c@{\;\;}c@{\;\;}c@{\;\;}c@{\;\;}c@{\;\;\;}c@{}}
\toprule
\multirow{2}{*}{\textbf{Model}} & \multicolumn{10}{c}{\textbf{Likelihood-based}} & \textbf{Pert.} & \multicolumn{2}{c}{\textbf{Prompt}} & \textbf{Out.-Dist.} & \multirow{2}{*}{\makecell{\textbf{Method-}\\\textbf{Avg}}} \\
\cmidrule(lr){2-11} \cmidrule(lr){12-12} \cmidrule(lr){13-14} \cmidrule(lr){15-15}
 & \textbf{PPL} & \textbf{Zlib} & \textbf{Lower} & \textbf{MinK} & \textbf{MinK++} & \textbf{DCPDD} & \textbf{PAC} & \textbf{ReCaLL} & \textbf{EMMIA} & \textbf{CAMIA} & \textbf{Neigh.} & \textbf{Guided} & \textbf{DCQ} & \textbf{CDD} & \\
\midrule
\textbf{1B} & 0.47 & 0.49 & 0.48 & 0.50 & 0.51 & 0.52 & 0.48 & 0.51 & \textbf{0.63} & 0.50 & 0.49 & 0.44 & 0.49 & 0.50 & 0.50 \\
\textbf{7B} & 0.47 & 0.49 & 0.48 & 0.48 & 0.49 & 0.51 & 0.47 & 0.55 & \textbf{0.70} & 0.47 & 0.51 & 0.52 & 0.52 & 0.50 & 0.51 \\
\textbf{13B} & 0.48 & 0.50 & 0.50 & 0.49 & 0.50 & 0.51 & 0.51 & 0.51 & \textbf{0.70} & 0.47 & 0.51 & 0.51 & 0.54 & 0.50 & 0.52 \\
\textbf{32B} & 0.49 & 0.51 & 0.50 & 0.51 & 0.48 & 0.52 & 0.51 & 0.52 & \textbf{0.71} & 0.47 & 0.53 & 0.44 & 0.64 & 0.49 & 0.52 \\
\cmidrule(l){2-16}
\textbf{Model-Avg} & 0.48 & 0.50 & 0.49 & 0.50 & 0.50 & 0.51 & 0.49 & 0.52 & \textbf{0.69} & 0.48 & 0.51 & 0.48 & 0.55 & 0.50 & 0.51 \\
\bottomrule
\end{tabular}%
}
\caption{Comparison of unsupervised MIAs on the peS2o domain of \textsc{OLMo-Detect}, evaluated via AUC. Methods are grouped by family as in Table~\ref{table:olmo_detect_overall_unsupervised}. \textbf{Model-Avg} averages across model sizes, while \textbf{Method-Avg} averages the methods with a value in that row. The best-performing method in each row is highlighted in bold.}
\label{table:auc_pes2o}
\end{table*}

\begin{table*}[t]
\centering
\resizebox{1\textwidth}{!}{%
\begin{tabular}{@{}l@{\;\;\;}c@{\;\;}c@{\;\;}c@{\;\;}c@{\;\;}c@{\;\;}c@{\;\;}c@{\;\;}c@{\;\;}c@{\;\;}c@{\;\;}c@{\;\;}c@{\;\;}c@{\;\;}c@{\;\;\;}c@{}}
\toprule
\multirow{2}{*}{\textbf{Model}} & \multicolumn{10}{c}{\textbf{Likelihood-based}} & \textbf{Pert.} & \multicolumn{2}{c}{\textbf{Prompt}} & \textbf{Out.-Dist.} & \multirow{2}{*}{\makecell{\textbf{Method-}\\\textbf{Avg}}} \\
\cmidrule(lr){2-11} \cmidrule(lr){12-12} \cmidrule(lr){13-14} \cmidrule(lr){15-15}
 & \textbf{PPL} & \textbf{Zlib} & \textbf{Lower} & \textbf{MinK} & \textbf{MinK++} & \textbf{DCPDD} & \textbf{PAC} & \textbf{ReCaLL} & \textbf{EMMIA} & \textbf{CAMIA} & \textbf{Neigh.} & \textbf{Guided} & \textbf{DCQ} & \textbf{CDD} & \\
\midrule
\textbf{1B} & 0.58 & 0.65 & 0.60 & 0.60 & 0.49 & 0.65 & \textbf{0.69} & 0.60 & 0.53 & 0.55 & 0.53 & 0.43 & 0.59 & 0.52 & 0.57 \\
\textbf{7B} & 0.60 & 0.65 & 0.62 & 0.61 & 0.50 & 0.67 & \textbf{0.68} & 0.60 & 0.53 & 0.54 & 0.53 & 0.55 & 0.46 & 0.54 & 0.58 \\
\textbf{13B} & 0.60 & 0.64 & 0.59 & 0.60 & 0.50 & 0.68 & \textbf{0.70} & 0.60 & 0.52 & 0.60 & 0.52 & 0.45 & 0.54 & 0.56 & 0.58 \\
\textbf{32B} & 0.60 & 0.63 & 0.58 & 0.62 & 0.48 & 0.69 & \textbf{0.70} & 0.56 & 0.64 & 0.61 & 0.52 & 0.58 & 0.45 & 0.54 & 0.59 \\
\cmidrule(l){2-16}
\textbf{Model-Avg} & 0.59 & 0.64 & 0.60 & 0.61 & 0.49 & 0.67 & \textbf{0.69} & 0.59 & 0.56 & 0.58 & 0.53 & 0.50 & 0.51 & 0.54 & 0.58 \\
\bottomrule
\end{tabular}%
}
\caption{Comparison of unsupervised MIAs on the StarCoder domain of \textsc{OLMo-Detect}, evaluated via AUC (macro-averaged over its subsets). Methods are grouped by family as in Table~\ref{table:olmo_detect_overall_unsupervised}. \textbf{Model-Avg} averages across model sizes, while \textbf{Method-Avg} averages the methods with a value in that row. The best-performing method in each row is highlighted in bold.}
\label{table:auc_starcoder}
\end{table*}

\begin{table*}[t]
\centering
\resizebox{1\textwidth}{!}{%
\begin{tabular}{@{}l@{\;\;\;}c@{\;\;}c@{\;\;}c@{\;\;}c@{\;\;}c@{\;\;}c@{\;\;}c@{\;\;}c@{\;\;}c@{\;\;}c@{\;\;}c@{\;\;}c@{\;\;}c@{\;\;}c@{\;\;\;}c@{}}
\toprule
\multirow{2}{*}{\textbf{Model}} & \multicolumn{10}{c}{\textbf{Likelihood-based}} & \textbf{Pert.} & \multicolumn{2}{c}{\textbf{Prompt}} & \textbf{Out.-Dist.} & \multirow{2}{*}{\makecell{\textbf{Method-}\\\textbf{Avg}}} \\
\cmidrule(lr){2-11} \cmidrule(lr){12-12} \cmidrule(lr){13-14} \cmidrule(lr){15-15}
 & \textbf{PPL} & \textbf{Zlib} & \textbf{Lower} & \textbf{MinK} & \textbf{MinK++} & \textbf{DCPDD} & \textbf{PAC} & \textbf{ReCaLL} & \textbf{EMMIA} & \textbf{CAMIA} & \textbf{Neigh.} & \textbf{Guided} & \textbf{DCQ} & \textbf{CDD} & \\
\midrule
\textbf{1B} & 0.51 & 0.56 & 0.61 & 0.55 & 0.48 & \textbf{0.61} & 0.54 & 0.44 & 0.54 & 0.56 & 0.54 & 0.47 & 0.52 & 0.51 & 0.53 \\
\textbf{7B} & 0.51 & 0.57 & \textbf{0.59} & 0.55 & 0.50 & 0.57 & 0.56 & 0.49 & 0.53 & 0.46 & 0.52 & 0.42 & 0.24 & 0.48 & 0.50 \\
\textbf{13B} & 0.52 & 0.58 & 0.61 & 0.56 & 0.53 & \textbf{0.63} & 0.58 & 0.48 & 0.52 & 0.45 & 0.55 & 0.44 & 0.42 & 0.49 & 0.53 \\
\textbf{32B} & 0.57 & 0.64 & 0.59 & 0.60 & 0.59 & \textbf{0.65} & 0.60 & 0.44 & 0.60 & 0.43 & 0.55 & 0.43 & 0.51 & 0.48 & 0.55 \\
\cmidrule(l){2-16}
\textbf{Model-Avg} & 0.53 & 0.59 & 0.60 & 0.56 & 0.52 & \textbf{0.61} & 0.57 & 0.46 & 0.55 & 0.47 & 0.54 & 0.44 & 0.42 & 0.49 & 0.53 \\
\bottomrule
\end{tabular}%
}
\caption{Comparison of unsupervised MIAs on the OpenWebMath domain of \textsc{OLMo-Detect}, evaluated via AUC. Methods are grouped by family as in Table~\ref{table:olmo_detect_overall_unsupervised}. \textbf{Model-Avg} averages across model sizes, while \textbf{Method-Avg} averages the methods with a value in that row. The best-performing method in each row is highlighted in bold.}
\label{table:auc_openwebmath}
\end{table*}

\begin{table*}[t]
\centering
\resizebox{1\textwidth}{!}{%
\begin{tabular}{@{}l@{\;\;\;}c@{\;\;}c@{\;\;}c@{\;\;}c@{\;\;}c@{\;\;}c@{\;\;}c@{\;\;}c@{\;\;}c@{\;\;}c@{\;\;}c@{\;\;}c@{\;\;}c@{\;\;}c@{\;\;\;}c@{}}
\toprule
\multirow{2}{*}{\textbf{Model}} & \multicolumn{10}{c}{\textbf{Likelihood-based}} & \textbf{Pert.} & \multicolumn{2}{c}{\textbf{Prompt}} & \textbf{Out.-Dist.} & \multirow{2}{*}{\makecell{\textbf{Method-}\\\textbf{Avg}}} \\
\cmidrule(lr){2-11} \cmidrule(lr){12-12} \cmidrule(lr){13-14} \cmidrule(lr){15-15}
 & \textbf{PPL} & \textbf{Zlib} & \textbf{Lower} & \textbf{MinK} & \textbf{MinK++} & \textbf{DCPDD} & \textbf{PAC} & \textbf{ReCaLL} & \textbf{EMMIA} & \textbf{CAMIA} & \textbf{Neigh.} & \textbf{Guided} & \textbf{DCQ} & \textbf{CDD} & \\
\midrule
\textbf{1B} & 0.73 & 0.76 & 0.72 & \textbf{0.76} & 0.57 & 0.69 & 0.72 & 0.32 & 0.50 & 0.75 & 0.64 & 0.49 & 0.46 & 0.52 & 0.62 \\
\textbf{7B} & 0.85 & 0.89 & 0.83 & 0.86 & 0.52 & \textbf{0.91} & 0.83 & 0.05 & 0.51 & 0.88 & 0.72 & 0.50 & 0.51 & 0.55 & 0.67 \\
\textbf{13B} & 0.98 & \textbf{0.99} & 0.93 & 0.99 & 0.58 & 0.99 & 0.96 & 0.01 & 0.48 & 0.99 & 0.89 & 0.48 & 0.53 & 0.68 & 0.75 \\
\textbf{32B} & 0.98 & 0.99 & 0.93 & 0.97 & 0.51 & 0.98 & 0.92 & 0.02 & 0.50 & \textbf{0.99} & 0.91 & 0.55 & 0.51 & 0.74 & 0.75 \\
\cmidrule(l){2-16}
\textbf{Model-Avg} & 0.89 & \textbf{0.91} & 0.85 & 0.89 & 0.55 & 0.89 & 0.86 & 0.10 & 0.50 & 0.90 & 0.79 & 0.51 & 0.50 & 0.62 & 0.70 \\
\bottomrule
\end{tabular}%
}
\caption{Comparison of unsupervised MIAs on the GSM8K subset of \textsc{OLMo-Detect}'s Math Mix domain, evaluated via AUC. Methods are grouped by family as in Table~\ref{table:olmo_detect_overall_unsupervised}. \textbf{Model-Avg} averages across model sizes, while \textbf{Method-Avg} averages the methods with a value in that row. The best-performing method in each row is highlighted in bold.}
\label{table:auc_gsm8k}
\end{table*}

\begin{table*}[t]
\centering
\resizebox{1\textwidth}{!}{%
\begin{tabular}{@{}l@{\;\;\;}c@{\;\;}c@{\;\;}c@{\;\;}c@{\;\;}c@{\;\;}c@{\;\;}c@{\;\;}c@{\;\;}c@{\;\;}c@{\;\;}c@{\;\;}c@{\;\;}c@{\;\;}c@{\;\;\;}c@{}}
\toprule
\multirow{2}{*}{\textbf{Model}} & \multicolumn{10}{c}{\textbf{Likelihood-based}} & \textbf{Pert.} & \multicolumn{2}{c}{\textbf{Prompt}} & \textbf{Out.-Dist.} & \multirow{2}{*}{\makecell{\textbf{Method-}\\\textbf{Avg}}} \\
\cmidrule(lr){2-11} \cmidrule(lr){12-12} \cmidrule(lr){13-14} \cmidrule(lr){15-15}
 & \textbf{PPL} & \textbf{Zlib} & \textbf{Lower} & \textbf{MinK} & \textbf{MinK++} & \textbf{DCPDD} & \textbf{PAC} & \textbf{ReCaLL} & \textbf{EMMIA} & \textbf{CAMIA} & \textbf{Neigh.} & \textbf{Guided} & \textbf{DCQ} & \textbf{CDD} & \\
\midrule
\textbf{1B} & 0.55 & \textbf{0.59} & 0.50 & 0.54 & 0.49 & 0.53 & 0.59 & 0.54 & 0.56 & 0.58 & 0.55 & 0.54 & 0.54 & 0.51 & 0.54 \\
\textbf{7B} & 0.58 & \textbf{0.64} & 0.55 & 0.60 & 0.56 & 0.61 & 0.63 & 0.53 & 0.57 & 0.59 & 0.56 & 0.45 & 0.40 & 0.53 & 0.56 \\
\textbf{13B} & 0.60 & \textbf{0.67} & 0.56 & 0.62 & 0.61 & 0.67 & 0.66 & 0.66 & 0.65 & 0.61 & 0.58 & 0.49 & 0.64 & 0.52 & 0.61 \\
\textbf{32B} & 0.63 & 0.70 & 0.53 & 0.65 & 0.61 & 0.68 & 0.69 & \textbf{0.73} & 0.69 & 0.64 & 0.60 & 0.54 & 0.64 & 0.50 & 0.63 \\
\cmidrule(l){2-16}
\textbf{Model-Avg} & 0.59 & \textbf{0.65} & 0.53 & 0.60 & 0.57 & 0.62 & 0.64 & 0.61 & 0.62 & 0.61 & 0.57 & 0.51 & 0.55 & 0.51 & 0.59 \\
\bottomrule
\end{tabular}%
}
\caption{Comparison of unsupervised MIAs on the Stack Exchange domain of \textsc{OLMo-Detect}, evaluated via AUC (macro-averaged over its subsets). Methods are grouped by family as in Table~\ref{table:olmo_detect_overall_unsupervised}. \textbf{Model-Avg} averages across model sizes, while \textbf{Method-Avg} averages the methods with a value in that row. The best-performing method in each row is highlighted in bold.}
\label{table:auc_stackexchange}
\end{table*}

\begin{table*}[t]
\centering
\resizebox{1\textwidth}{!}{%
\begin{tabular}{@{}l@{\;\;\;}c@{\;\;}c@{\;\;}c@{\;\;}c@{\;\;}c@{\;\;}c@{\;\;}c@{\;\;}c@{\;\;}c@{\;\;}c@{\;\;}c@{\;\;}c@{\;\;}c@{\;\;}c@{\;\;\;}c@{}}
\toprule
\multirow{2}{*}{\textbf{Model}} & \multicolumn{10}{c}{\textbf{Likelihood-based}} & \textbf{Pert.} & \multicolumn{2}{c}{\textbf{Prompt}} & \textbf{Out.-Dist.} & \multirow{2}{*}{\makecell{\textbf{Method-}\\\textbf{Avg}}} \\
\cmidrule(lr){2-11} \cmidrule(lr){12-12} \cmidrule(lr){13-14} \cmidrule(lr){15-15}
 & \textbf{PPL} & \textbf{Zlib} & \textbf{Lower} & \textbf{MinK} & \textbf{MinK++} & \textbf{DCPDD} & \textbf{PAC} & \textbf{ReCaLL} & \textbf{EMMIA} & \textbf{CAMIA} & \textbf{Neigh.} & \textbf{Guided} & \textbf{DCQ} & \textbf{CDD} & \\
\midrule
\textbf{1B} & 0.62 & 0.64 & 0.49 & 0.63 & \textbf{0.73} & 0.56 & 0.51 & 0.61 & 0.57 & 0.49 & 0.54 & 0.48 & 0.51 & 0.51 & 0.56 \\
\textbf{7B} & 0.73 & 0.70 & 0.55 & 0.71 & \textbf{0.77} & 0.70 & 0.59 & 0.74 & 0.59 & 0.74 & 0.60 & 0.51 & 0.43 & 0.50 & 0.63 \\
\textbf{13B} & 0.65 & 0.61 & 0.50 & 0.64 & \textbf{0.70} & 0.60 & 0.55 & 0.65 & 0.55 & 0.57 & 0.51 & 0.54 & 0.44 & 0.51 & 0.57 \\
\textbf{32B} & 0.64 & 0.64 & 0.50 & 0.65 & \textbf{0.69} & 0.60 & 0.51 & 0.69 & 0.55 & 0.54 & 0.52 & 0.44 & 0.45 & 0.50 & 0.57 \\
\cmidrule(l){2-16}
\textbf{Model-Avg} & 0.66 & 0.65 & 0.51 & 0.66 & \textbf{0.72} & 0.61 & 0.54 & 0.67 & 0.56 & 0.58 & 0.54 & 0.49 & 0.46 & 0.51 & 0.58 \\
\bottomrule
\end{tabular}%
}
\caption{Comparison of unsupervised MIAs on the SFT domain of \textsc{OLMo-Detect}, evaluated via AUC (macro-averaged over its subsets). Methods are grouped by family as in Table~\ref{table:olmo_detect_overall_unsupervised}. \textbf{Model-Avg} averages across model sizes, while \textbf{Method-Avg} averages the methods with a value in that row. The best-performing method in each row is highlighted in bold.}
\label{table:auc_sft}
\end{table*}

\begin{table*}[t]
\centering
\resizebox{1\textwidth}{!}{%
\begin{tabular}{@{}l@{\;\;\;}c@{\;\;}c@{\;\;}c@{\;\;}c@{\;\;}c@{\;\;}c@{\;\;}c@{\;\;}c@{\;\;}c@{\;\;}c@{\;\;}c@{\;\;}c@{\;\;}c@{\;\;}c@{\;\;}c@{\;\;\;}c@{}}
\toprule
\multirow{2}{*}{\textbf{Model}} & \multicolumn{10}{c}{\textbf{Likelihood-based}} & \textbf{Pert.} & \multicolumn{2}{c}{\textbf{Prompt}} & \multicolumn{2}{c}{\textbf{Out.-Dist.}} & \multirow{2}{*}{\makecell{\textbf{Method-}\\\textbf{Avg}}} \\
\cmidrule(lr){2-11} \cmidrule(lr){12-12} \cmidrule(lr){13-14} \cmidrule(lr){15-16}
 & \textbf{PPL} & \textbf{Zlib} & \textbf{Lower} & \textbf{MinK} & \textbf{MinK++} & \textbf{DCPDD} & \textbf{PAC} & \textbf{ReCaLL} & \textbf{EMMIA} & \textbf{CAMIA} & \textbf{Neigh.} & \textbf{Guided} & \textbf{DCQ} & \textbf{CDD} & \textbf{SelfCrit}$^\dagger$ & \\
\midrule
\textbf{1B} & 0.44 & 0.55 & \textbf{0.64} & 0.44 & 0.45 & 0.47 & 0.62 & 0.54 & 0.60 & 0.59 & 0.54 & 0.53 & 0.45 & 0.53 & 0.43 & 0.52 \\
\textbf{7B} & 0.41 & 0.39 & 0.50 & 0.40 & 0.40 & 0.40 & 0.43 & 0.48 & 0.50 & \textbf{0.60} & 0.46 & 0.48 & 0.30 & 0.53 & 0.53 & 0.45 \\
\textbf{13B} & 0.38 & 0.47 & \textbf{0.64} & 0.38 & 0.45 & 0.39 & 0.53 & 0.59 & 0.49 & 0.57 & 0.49 & 0.45 & 0.50 & 0.53 & 0.43 & 0.49 \\
\textbf{32B} & 0.46 & 0.53 & 0.62 & 0.48 & 0.47 & 0.48 & 0.62 & \textbf{0.65} & 0.39 & 0.55 & 0.60 & 0.47 & 0.47 & 0.48 & 0.55 & 0.52 \\
\cmidrule(l){2-17}
\textbf{Model-Avg} & 0.42 & 0.48 & \textbf{0.60} & 0.43 & 0.44 & 0.43 & 0.55 & 0.57 & 0.49 & 0.58 & 0.52 & 0.48 & 0.43 & 0.52 & 0.48 & 0.50 \\
\bottomrule
\end{tabular}%
}
\caption{Comparison of unsupervised MIAs on the UF-Wildchat subset of \textsc{OLMo-Detect}'s DPO domain, evaluated via AUC. Methods are grouped by family as in Table~\ref{table:olmo_detect_overall_unsupervised}. \textbf{Model-Avg} averages across model sizes, while \textbf{Method-Avg} averages the methods with a value in that row. The best-performing method in each row is highlighted in bold. $^\dagger$SelfCrit is evaluated only on DPO and RLVR.}
\label{table:auc_dpo}
\end{table*}

\begin{table*}[t]
\centering
\resizebox{1\textwidth}{!}{%
\begin{tabular}{@{}l@{\;\;\;}c@{\;\;}c@{\;\;}c@{\;\;}c@{\;\;}c@{\;\;}c@{\;\;}c@{\;\;}c@{\;\;}c@{\;\;}c@{\;\;}c@{\;\;}c@{\;\;}c@{\;\;}c@{\;\;}c@{\;\;\;}c@{}}
\toprule
\multirow{2}{*}{\textbf{Model}} & \multicolumn{10}{c}{\textbf{Likelihood-based}} & \textbf{Pert.} & \multicolumn{2}{c}{\textbf{Prompt}} & \multicolumn{2}{c}{\textbf{Out.-Dist.}} & \multirow{2}{*}{\makecell{\textbf{Method-}\\\textbf{Avg}}} \\
\cmidrule(lr){2-11} \cmidrule(lr){12-12} \cmidrule(lr){13-14} \cmidrule(lr){15-16}
 & \textbf{PPL} & \textbf{Zlib} & \textbf{Lower} & \textbf{MinK} & \textbf{MinK++} & \textbf{DCPDD} & \textbf{PAC} & \textbf{ReCaLL} & \textbf{EMMIA} & \textbf{CAMIA} & \textbf{Neigh.} & \textbf{Guided} & \textbf{DCQ} & \textbf{CDD} & \textbf{SelfCrit}$^\dagger$ & \\
\midrule
\textbf{1B} & 0.77 & 0.79 & 0.64 & \textbf{0.79} & 0.60 & 0.74 & 0.69 & 0.36 & 0.66 & 0.68 & 0.51 & 0.50 & 0.43 & 0.48 & 0.50 & 0.61 \\
\textbf{7B} & 0.89 & 0.88 & 0.80 & \textbf{0.90} & 0.69 & 0.87 & 0.77 & 0.25 & 0.73 & 0.82 & 0.60 & 0.53 & 0.48 & 0.48 & 0.48 & 0.68 \\
\textbf{13B} & 0.95 & 0.94 & 0.81 & \textbf{0.95} & 0.62 & 0.92 & 0.75 & 0.18 & 0.76 & 0.85 & 0.58 & 0.47 & 0.50 & 0.46 & 0.46 & 0.68 \\
\textbf{32B} & 0.94 & 0.91 & 0.90 & \textbf{0.94} & 0.72 & 0.92 & 0.77 & 0.28 & 0.47 & 0.89 & 0.60 & 0.47 & 0.50 & 0.49 & 0.41 & 0.68 \\
\cmidrule(l){2-17}
\textbf{Model-Avg} & 0.89 & 0.88 & 0.79 & \textbf{0.90} & 0.66 & 0.86 & 0.75 & 0.27 & 0.65 & 0.81 & 0.57 & 0.49 & 0.48 & 0.48 & 0.46 & 0.66 \\
\bottomrule
\end{tabular}%
}
\caption{Comparison of unsupervised MIAs on the RLVR domain of \textsc{OLMo-Detect}, evaluated via AUC (macro-averaged over its subsets). Methods are grouped by family as in Table~\ref{table:olmo_detect_overall_unsupervised}. \textbf{Model-Avg} averages across model sizes, while \textbf{Method-Avg} averages the methods with a value in that row. The best-performing method in each row is highlighted in bold. $^\dagger$SelfCrit is evaluated only on DPO and RLVR.}
\label{table:auc_rlvr}
\end{table*}

\begin{table*}[t]
\centering
\resizebox{1\textwidth}{!}{%
\begin{tabular}{@{}l@{\;\;}l@{\;\;\;}c@{\;\;}c@{\;\;}c@{\;\;}c@{\;\;}c@{\;\;}c@{\;\;}c@{\;\;}c@{\;\;}c@{\;\;}c@{\;\;}c@{\;\;}c@{\;\;}c@{\;\;}c@{\;\;\;}c@{}}
\toprule
\multirow{2}{*}{\textbf{Subset}} & \multirow{2}{*}{\textbf{Model}} & \multicolumn{10}{c}{\textbf{Likelihood-based}} & \textbf{Pert.} & \multicolumn{2}{c}{\textbf{Prompt}} & \textbf{Out.-Dist.} & \multirow{2}{*}{\makecell{\textbf{Method-}\\\textbf{Avg}}} \\
\cmidrule(lr){3-12} \cmidrule(lr){13-13} \cmidrule(lr){14-15} \cmidrule(lr){16-16}
 &  & \textbf{PPL} & \textbf{Zlib} & \textbf{Lower} & \textbf{MinK} & \textbf{MinK++} & \textbf{DCPDD} & \textbf{PAC} & \textbf{ReCaLL} & \textbf{EMMIA} & \textbf{CAMIA} & \textbf{Neigh.} & \textbf{Guided} & \textbf{DCQ} & \textbf{CDD} & \\
\midrule
\multirow{5}{*}{\textbf{Python}} & 1B & 0.54 & 0.61 & 0.58 & 0.55 & 0.50 & 0.61 & \textbf{0.69} & 0.52 & 0.49 & 0.58 & 0.52 & 0.38 & 0.57 & 0.52 & 0.55 \\
 & 7B & 0.55 & 0.62 & 0.60 & 0.56 & 0.53 & 0.61 & \textbf{0.68} & 0.51 & 0.50 & 0.58 & 0.52 & 0.48 & 0.51 & 0.51 & 0.56 \\
 & 13B & 0.55 & 0.62 & 0.58 & 0.56 & 0.52 & 0.62 & \textbf{0.67} & 0.56 & 0.51 & 0.62 & 0.51 & 0.48 & 0.54 & 0.52 & 0.56 \\
 & 32B & 0.54 & 0.60 & 0.55 & 0.55 & 0.53 & 0.61 & \textbf{0.66} & 0.53 & 0.62 & 0.60 & 0.53 & 0.56 & 0.49 & 0.56 & 0.57 \\
\cmidrule(l){2-17}
 & \textit{Model-Avg} & 0.54 & 0.62 & 0.58 & 0.55 & 0.52 & 0.61 & \textbf{0.68} & 0.53 & 0.53 & 0.60 & 0.52 & 0.48 & 0.53 & 0.53 & 0.56 \\
\cmidrule(l){1-17}
\multirow{5}{*}{\textbf{Java}} & 1B & 0.62 & 0.68 & 0.61 & 0.65 & 0.47 & 0.69 & \textbf{0.70} & 0.68 & 0.57 & 0.51 & 0.54 & 0.47 & 0.61 & 0.51 & 0.59 \\
\bottomrule
\end{tabular}%
}
\caption{Comparison of unsupervised MIAs on the subsets of the StarCoder domain of \textsc{OLMo-Detect}, evaluated via AUC. Methods are grouped by family as in Table~\ref{table:olmo_detect_overall_unsupervised}. \textbf{Model-Avg} averages across model sizes, while \textbf{Method-Avg} averages the methods with a value in that row. The best-performing method in each row is highlighted in bold.}
\label{table:auc_starcoder_subsets}
\end{table*}

\begin{table*}[t]
\centering
\resizebox{1\textwidth}{!}{%
\begin{tabular}{@{}l@{\;\;}l@{\;\;\;}c@{\;\;}c@{\;\;}c@{\;\;}c@{\;\;}c@{\;\;}c@{\;\;}c@{\;\;}c@{\;\;}c@{\;\;}c@{\;\;}c@{\;\;}c@{\;\;}c@{\;\;}c@{\;\;\;}c@{}}
\toprule
\multirow{2}{*}{\textbf{Subset}} & \multirow{2}{*}{\textbf{Model}} & \multicolumn{10}{c}{\textbf{Likelihood-based}} & \textbf{Pert.} & \multicolumn{2}{c}{\textbf{Prompt}} & \textbf{Out.-Dist.} & \multirow{2}{*}{\makecell{\textbf{Method-}\\\textbf{Avg}}} \\
\cmidrule(lr){3-12} \cmidrule(lr){13-13} \cmidrule(lr){14-15} \cmidrule(lr){16-16}
 &  & \textbf{PPL} & \textbf{Zlib} & \textbf{Lower} & \textbf{MinK} & \textbf{MinK++} & \textbf{DCPDD} & \textbf{PAC} & \textbf{ReCaLL} & \textbf{EMMIA} & \textbf{CAMIA} & \textbf{Neigh.} & \textbf{Guided} & \textbf{DCQ} & \textbf{CDD} & \\
\midrule
\multirow{5}{*}{\textbf{Math Stack Exchange}} & 1B & 0.55 & 0.60 & 0.49 & 0.54 & 0.49 & 0.55 & \textbf{0.62} & 0.57 & 0.58 & 0.60 & 0.57 & 0.60 & 0.59 & 0.52 & 0.56 \\
 & 7B & 0.57 & 0.63 & 0.53 & 0.58 & 0.55 & 0.59 & \textbf{0.67} & 0.51 & 0.62 & 0.61 & 0.57 & 0.41 & 0.36 & 0.54 & 0.55 \\
 & 13B & 0.60 & 0.67 & 0.55 & 0.62 & 0.62 & 0.68 & \textbf{0.71} & 0.63 & 0.69 & 0.64 & 0.60 & 0.45 & 0.63 & 0.51 & 0.61 \\
 & 32B & 0.62 & 0.71 & 0.51 & 0.65 & 0.61 & 0.70 & 0.71 & \textbf{0.79} & 0.74 & 0.66 & 0.61 & 0.57 & 0.59 & 0.50 & 0.64 \\
\cmidrule(l){2-17}
 & \textit{Model-Avg} & 0.59 & 0.65 & 0.52 & 0.60 & 0.57 & 0.63 & \textbf{0.68} & 0.63 & 0.66 & 0.63 & 0.59 & 0.51 & 0.54 & 0.52 & 0.59 \\
\cmidrule(l){1-17}
\multirow{5}{*}{\textbf{Stack Overflow}} & 1B & 0.56 & \textbf{0.58} & 0.50 & 0.54 & 0.49 & 0.52 & 0.55 & 0.51 & 0.55 & 0.57 & 0.54 & 0.49 & 0.49 & 0.49 & 0.53 \\
\bottomrule
\end{tabular}%
}
\caption{Comparison of unsupervised MIAs on the subsets of the Stack Exchange domain of \textsc{OLMo-Detect}, evaluated via AUC. Methods are grouped by family as in Table~\ref{table:olmo_detect_overall_unsupervised}. \textbf{Model-Avg} averages across model sizes, while \textbf{Method-Avg} averages the methods with a value in that row. The best-performing method in each row is highlighted in bold.}
\label{table:auc_stackexchange_subsets}
\end{table*}

\begin{table*}[t]
\centering
\resizebox{1\textwidth}{!}{%
\begin{tabular}{@{}l@{\;\;}l@{\;\;\;}c@{\;\;}c@{\;\;}c@{\;\;}c@{\;\;}c@{\;\;}c@{\;\;}c@{\;\;}c@{\;\;}c@{\;\;}c@{\;\;}c@{\;\;}c@{\;\;}c@{\;\;}c@{\;\;\;}c@{}}
\toprule
\multirow{2}{*}{\textbf{Subset}} & \multirow{2}{*}{\textbf{Model}} & \multicolumn{10}{c}{\textbf{Likelihood-based}} & \textbf{Pert.} & \multicolumn{2}{c}{\textbf{Prompt}} & \textbf{Out.-Dist.} & \multirow{2}{*}{\makecell{\textbf{Method-}\\\textbf{Avg}}} \\
\cmidrule(lr){3-12} \cmidrule(lr){13-13} \cmidrule(lr){14-15} \cmidrule(lr){16-16}
 &  & \textbf{PPL} & \textbf{Zlib} & \textbf{Lower} & \textbf{MinK} & \textbf{MinK++} & \textbf{DCPDD} & \textbf{PAC} & \textbf{ReCaLL} & \textbf{EMMIA} & \textbf{CAMIA} & \textbf{Neigh.} & \textbf{Guided} & \textbf{DCQ} & \textbf{CDD} & \\
\midrule
\multirow{5}{*}{\textbf{Aya}} & 1B & 0.56 & 0.58 & 0.50 & 0.54 & \textbf{0.67} & 0.54 & 0.50 & 0.48 & 0.58 & 0.54 & 0.54 & 0.49 & 0.48 & 0.50 & 0.53 \\
 & 7B & \textbf{0.77} & 0.70 & 0.53 & 0.72 & 0.76 & 0.72 & 0.67 & 0.69 & 0.59 & 0.76 & 0.69 & 0.52 & 0.47 & 0.53 & 0.65 \\
 & 13B & \textbf{0.69} & 0.60 & 0.49 & 0.66 & 0.64 & 0.62 & 0.60 & 0.56 & 0.51 & 0.67 & 0.59 & 0.51 & 0.49 & 0.55 & 0.59 \\
 & 32B & 0.55 & 0.56 & 0.49 & 0.54 & 0.59 & 0.55 & 0.49 & \textbf{0.60} & 0.44 & 0.50 & 0.51 & 0.38 & 0.51 & 0.51 & 0.52 \\
\cmidrule(l){2-17}
 & \textit{Model-Avg} & 0.64 & 0.61 & 0.51 & 0.61 & \textbf{0.66} & 0.61 & 0.57 & 0.58 & 0.53 & 0.62 & 0.58 & 0.47 & 0.49 & 0.52 & 0.57 \\
\cmidrule(l){1-17}
\multirow{5}{*}{\textbf{WildChat}} & 1B & 0.69 & 0.70 & 0.48 & 0.71 & \textbf{0.79} & 0.58 & 0.51 & 0.75 & 0.55 & 0.44 & 0.54 & 0.47 & 0.54 & 0.52 & 0.59 \\
\bottomrule
\end{tabular}%
}
\caption{Comparison of unsupervised MIAs on the subsets of the SFT domain of \textsc{OLMo-Detect}, evaluated via AUC. Methods are grouped by family as in Table~\ref{table:olmo_detect_overall_unsupervised}. \textbf{Model-Avg} averages across model sizes, while \textbf{Method-Avg} averages the methods with a value in that row. The best-performing method in each row is highlighted in bold.}
\label{table:auc_sft_subsets}
\end{table*}

\begin{table*}[t]
\centering
\resizebox{1\textwidth}{!}{%
\begin{tabular}{@{}l@{\;\;}l@{\;\;\;}c@{\;\;}c@{\;\;}c@{\;\;}c@{\;\;}c@{\;\;}c@{\;\;}c@{\;\;}c@{\;\;}c@{\;\;}c@{\;\;}c@{\;\;}c@{\;\;}c@{\;\;}c@{\;\;}c@{\;\;\;}c@{}}
\toprule
\multirow{2}{*}{\textbf{Subset}} & \multirow{2}{*}{\textbf{Model}} & \multicolumn{10}{c}{\textbf{Likelihood-based}} & \textbf{Pert.} & \multicolumn{2}{c}{\textbf{Prompt}} & \multicolumn{2}{c}{\textbf{Out.-Dist.}} & \multirow{2}{*}{\makecell{\textbf{Method-}\\\textbf{Avg}}} \\
\cmidrule(lr){3-12} \cmidrule(lr){13-13} \cmidrule(lr){14-15} \cmidrule(lr){16-17}
 &  & \textbf{PPL} & \textbf{Zlib} & \textbf{Lower} & \textbf{MinK} & \textbf{MinK++} & \textbf{DCPDD} & \textbf{PAC} & \textbf{ReCaLL} & \textbf{EMMIA} & \textbf{CAMIA} & \textbf{Neigh.} & \textbf{Guided} & \textbf{DCQ} & \textbf{CDD} & \textbf{SelfCrit}$^\dagger$ & \\
\midrule
\multirow{5}{*}{\textbf{CoT-GSM8K}} & 1B & 0.85 & \textbf{0.89} & 0.73 & 0.88 & 0.62 & 0.80 & 0.78 & 0.21 & 0.80 & 0.75 & 0.49 & 0.51 & 0.42 & 0.51 & 0.47 & 0.65 \\
 & 7B & 0.98 & 0.99 & 0.87 & \textbf{0.99} & 0.78 & 0.98 & 0.89 & 0.10 & 0.97 & 0.98 & 0.62 & 0.51 & 0.50 & 0.53 & 0.50 & 0.75 \\
 & 13B & \textbf{1.00} & 1.00 & 0.90 & 1.00 & 0.64 & 0.99 & 0.78 & 0.01 & 1.00 & 0.98 & 0.56 & 0.48 & 0.41 & 0.49 & 0.39 & 0.71 \\
 & 32B & \textbf{1.00} & 1.00 & 0.95 & 1.00 & 0.79 & 1.00 & 0.87 & 0.04 & 0.50 & 0.99 & 0.61 & 0.49 & 0.51 & 0.50 & 0.35 & 0.71 \\
\cmidrule(l){2-18}
 & \textit{Model-Avg} & 0.96 & \textbf{0.97} & 0.86 & 0.97 & 0.71 & 0.94 & 0.83 & 0.09 & 0.82 & 0.93 & 0.57 & 0.50 & 0.46 & 0.51 & 0.43 & 0.70 \\
\cmidrule(l){1-18}
\multirow{5}{*}{\textbf{CoT-MATH}} & 1B & 0.70 & 0.68 & 0.54 & \textbf{0.71} & 0.58 & 0.69 & 0.60 & 0.51 & 0.52 & 0.61 & 0.52 & 0.50 & 0.43 & 0.46 & 0.53 & 0.57 \\
\bottomrule
\end{tabular}%
}
\caption{Comparison of unsupervised MIAs on the subsets of the RLVR domain of \textsc{OLMo-Detect}, evaluated via AUC. Methods are grouped by family as in Table~\ref{table:olmo_detect_overall_unsupervised}. \textbf{Model-Avg} averages across model sizes, while \textbf{Method-Avg} averages the methods with a value in that row. The best-performing method in each row is highlighted in bold. $^\dagger$SelfCrit is evaluated only on DPO and RLVR.}
\label{table:auc_rlvr_subsets}
\end{table*}

\section{TPR@5\% FPR Results for Unsupervised MIAs on \textsc{OLMo-Detect}}
\label{appendix:tpr_results_unsup}

This appendix provides detailed per-domain TPR@5\% FPR results for unsupervised MIAs on \textsc{OLMo-Detect}. Stage-level aggregate results are reported in Table~\ref{table:tpr_unsup:all}, while per-domain results are shown in Tables~\ref{table:tpr_unsup:dclm}--\ref{table:tpr_unsup:rlvr_subsets}.

\begin{table*}[t]
\centering
\resizebox{0.99\textwidth}{!}{%
\begin{tabular}{@{}l@{\;\;}l@{\;\;\;}c@{\;\;}c@{\;\;}c@{\;\;}c@{\;\;}c@{\;\;}c@{\;\;}c@{\;\;}c@{\;\;}c@{\;\;}c@{\;\;}c@{\;\;}c@{\;\;}c@{\;\;}c@{\;\;}c@{\;\;\;}c@{}}
\toprule
\multirow{2}{*}{\textbf{Model}} & \multirow{2}{*}{\textbf{Stage}} & \multicolumn{10}{c}{\textbf{Likelihood-based}} & \textbf{Pert.} & \multicolumn{2}{c}{\textbf{Prompt}} & \multicolumn{2}{c}{\textbf{Out.-Dist.}} & \multirow{2}{*}{\makecell{\textbf{Method-}\\\textbf{Avg}}} \\
\cmidrule(lr){3-12} \cmidrule(lr){13-13} \cmidrule(lr){14-15} \cmidrule(lr){16-17}
 & & \textbf{PPL} & \textbf{Zlib} & \textbf{Lower} & \textbf{MinK} & \textbf{MinK++} & \textbf{DCPDD} & \textbf{PAC} & \textbf{ReCaLL} & \textbf{EMMIA} & \textbf{CAMIA} & \textbf{Neigh.} & \textbf{Guided} & \textbf{DCQ} & \textbf{CDD} & \textbf{SelfCrit}$^\dagger$ & \\
\midrule
\multirow{4}{*}{1B}
 & Pre  & 0.07 & 0.09 & 0.08 & 0.09 & 0.07 & \textbf{0.12} & 0.12 & 0.07 & 0.10 & 0.09 & 0.05 & 0.03 & 0.07 & 0.06 & --- & 0.08 \\
 & Mid  & 0.14 & \textbf{0.17} & 0.15 & 0.15 & 0.09 & 0.10 & 0.14 & 0.07 & 0.09 & 0.15 & 0.10 & 0.06 & 0.07 & 0.06 & --- & 0.11 \\
 & Post & \textbf{0.20} & 0.18 & 0.12 & 0.20 & 0.15 & 0.12 & 0.13 & 0.10 & 0.09 & 0.17 & 0.05 & 0.05 & 0.04 & 0.07 & 0.07 & 0.12 \\ \cmidrule(l){2-18}
 & All  & 0.13 & \textbf{0.15} & 0.12 & 0.15 & 0.10 & 0.11 & 0.13 & 0.08 & 0.09 & 0.14 & 0.07 & 0.05 & 0.06 & 0.06 & 0.07 & 0.10 \\
\cmidrule(l){1-18}
\multirow{4}{*}{7B}
 & Pre  & 0.08 & 0.13 & 0.10 & 0.11 & 0.09 & 0.15 & 0.13 & 0.08 & \textbf{0.18} & 0.12 & 0.06 & 0.05 & 0.05 & 0.06 & --- & 0.10 \\
 & Mid  & 0.24 & 0.33 & 0.15 & 0.28 & 0.10 & \textbf{0.39} & 0.24 & 0.02 & 0.07 & 0.27 & 0.16 & 0.05 & 0.05 & 0.07 & --- & 0.17 \\
 & Post & \textbf{0.36} & 0.32 & 0.18 & 0.34 & 0.23 & 0.30 & 0.15 & 0.09 & 0.22 & 0.27 & 0.07 & 0.05 & 0.03 & 0.07 & 0.04 & 0.18 \\ \cmidrule(l){2-18}
 & All  & 0.23 & 0.26 & 0.14 & 0.24 & 0.14 & \textbf{0.28} & 0.17 & 0.06 & 0.16 & 0.22 & 0.10 & 0.05 & 0.04 & 0.06 & 0.04 & 0.15 \\
\cmidrule(l){1-18}
\multirow{4}{*}{13B}
 & Pre  & 0.10 & 0.13 & 0.09 & 0.11 & 0.07 & 0.13 & 0.17 & 0.06 & \textbf{0.19} & 0.10 & 0.08 & 0.04 & 0.05 & 0.06 & --- & 0.10 \\
 & Mid  & 0.49 & \textbf{0.57} & 0.33 & 0.51 & 0.14 & 0.55 & 0.49 & 0.05 & 0.07 & 0.50 & 0.31 & 0.06 & 0.05 & 0.11 & --- & 0.30 \\
 & Post & \textbf{0.34} & 0.34 & 0.26 & 0.33 & 0.12 & 0.28 & 0.14 & 0.15 & 0.22 & 0.28 & 0.07 & 0.05 & 0.06 & 0.08 & 0.06 & 0.19 \\ \cmidrule(l){2-18}
 & All  & 0.31 & \textbf{0.35} & 0.22 & 0.31 & 0.11 & 0.32 & 0.26 & 0.09 & 0.16 & 0.30 & 0.15 & 0.05 & 0.05 & 0.08 & 0.06 & 0.19 \\
\cmidrule(l){1-18}
\multirow{4}{*}{32B}
 & Pre  & 0.12 & 0.16 & 0.07 & 0.12 & 0.10 & 0.16 & \textbf{0.20} & 0.06 & 0.20 & 0.12 & 0.08 & 0.04 & 0.07 & 0.06 & --- & 0.11 \\
 & Mid  & 0.49 & \textbf{0.58} & 0.39 & 0.49 & 0.13 & 0.56 & 0.46 & 0.08 & 0.12 & 0.52 & 0.36 & 0.11 & 0.10 & 0.14 & --- & 0.32 \\
 & Post & 0.35 & 0.31 & 0.19 & \textbf{0.37} & 0.18 & 0.32 & 0.17 & 0.13 & 0.04 & 0.31 & 0.09 & 0.04 & 0.05 & 0.07 & 0.07 & 0.18 \\ \cmidrule(l){2-18}
 & All  & 0.32 & \textbf{0.35} & 0.22 & 0.32 & 0.14 & 0.35 & 0.28 & 0.09 & 0.12 & 0.31 & 0.18 & 0.06 & 0.07 & 0.09 & 0.07 & 0.20 \\
\midrule
\multirow{4}{*}{\makecell{\textbf{Model-}\\\textbf{Avg}}}
 & Pre  & 0.09 & 0.13 & 0.09 & 0.10 & 0.08 & 0.14 & 0.15 & 0.07 & \textbf{0.17} & 0.11 & 0.07 & 0.04 & 0.06 & 0.06 & --- & 0.10 \\
 & Mid  & 0.34 & \textbf{0.41} & 0.26 & 0.36 & 0.11 & 0.40 & 0.33 & 0.05 & 0.09 & 0.36 & 0.24 & 0.07 & 0.07 & 0.10 & --- & 0.23 \\
 & Post & \textbf{0.31} & 0.29 & 0.19 & 0.31 & 0.17 & 0.25 & 0.15 & 0.12 & 0.14 & 0.26 & 0.07 & 0.05 & 0.04 & 0.07 & 0.06 & 0.17 \\ \cmidrule(l){2-18}
 & All  & 0.25 & \textbf{0.28} & 0.18 & 0.26 & 0.12 & 0.26 & 0.21 & 0.08 & 0.13 & 0.24 & 0.12 & 0.05 & 0.06 & 0.07 & 0.06 & 0.16 \\
\bottomrule
\end{tabular}%
}
\caption{Comparison of unsupervised MIAs on \textsc{OLMo-Detect}, evaluated via TPR@5\% FPR. Methods are grouped by family: likelihood-based, perturbation-based (\textbf{Pert.}), prompt-based and output-distribution-based (\textbf{Out.-Dist.}). \textbf{Pre}/\textbf{Mid}/\textbf{Post} report macro-averaged TPRs over the domains of each stage, while \textbf{All} macro-averages the three stages. \textbf{Model-Avg} averages across model sizes, while \textbf{Method-Avg} averages the methods with a value in that row. The best-performing method in each row is highlighted in bold. $^\dagger$SelfCrit is evaluated only on DPO and RLVR and thus has only \textbf{Post} and \textbf{All} values, averaged over these two domains. Per-domain results are provided in Tables~\ref{table:tpr_unsup:dclm}--\ref{table:tpr_unsup:rlvr_subsets}.}
\label{table:tpr_unsup:all}
\end{table*}

\begin{table*}[t]
\centering
\resizebox{0.99\textwidth}{!}{%
\begin{tabular}{@{}l@{\;\;\;}c@{\;\;}c@{\;\;}c@{\;\;}c@{\;\;}c@{\;\;}c@{\;\;}c@{\;\;}c@{\;\;}c@{\;\;}c@{\;\;}c@{\;\;}c@{\;\;}c@{\;\;}c@{\;\;\;}c@{}}
\toprule
\multirow{2}{*}{\textbf{Model}} & \multicolumn{10}{c}{\textbf{Likelihood-based}} & \textbf{Pert.} & \multicolumn{2}{c}{\textbf{Prompt}} & \textbf{Out.-Dist.} & \multirow{2}{*}{\makecell{\textbf{Method-}\\\textbf{Avg}}} \\
\cmidrule(lr){2-11} \cmidrule(lr){12-12} \cmidrule(lr){13-14} \cmidrule(lr){15-15}
 & \textbf{PPL} & \textbf{Zlib} & \textbf{Lower} & \textbf{MinK} & \textbf{MinK++} & \textbf{DCPDD} & \textbf{PAC} & \textbf{ReCaLL} & \textbf{EMMIA} & \textbf{CAMIA} & \textbf{Neigh.} & \textbf{Guided} & \textbf{DCQ} & \textbf{CDD} & \\
\midrule
\textbf{1B} & 0.09 & 0.06 & 0.14 & 0.13 & 0.01 & 0.12 & \textbf{0.16} & 0.03 & 0.06 & 0.13 & 0.09 & 0.04 & 0.03 & 0.06 & 0.08 \\
\textbf{7B} & 0.15 & 0.11 & 0.16 & 0.21 & 0.06 & 0.18 & 0.20 & 0.03 & 0.07 & \textbf{0.22} & 0.08 & 0.06 & 0.04 & 0.04 & 0.12 \\
\textbf{13B} & 0.16 & 0.13 & 0.14 & 0.20 & 0.05 & 0.15 & \textbf{0.26} & 0.06 & 0.08 & 0.23 & 0.10 & 0.04 & 0.03 & 0.06 & 0.12 \\
\textbf{32B} & 0.21 & 0.18 & 0.13 & 0.22 & 0.11 & 0.18 & \textbf{0.34} & 0.06 & 0.12 & 0.26 & 0.11 & 0.04 & 0.07 & 0.07 & 0.15 \\
\cmidrule(l){2-16}
\textbf{Model-Avg} & 0.15 & 0.12 & 0.14 & 0.19 & 0.06 & 0.16 & \textbf{0.24} & 0.04 & 0.08 & 0.21 & 0.10 & 0.05 & 0.04 & 0.06 & 0.12 \\
\bottomrule
\end{tabular}%
}
\caption{Comparison of unsupervised MIAs on the DCLM-Baseline domain of \textsc{OLMo-Detect}, evaluated via TPR@5\% FPR. Methods are grouped by family as in Table~\ref{table:tpr_unsup:all}. \textbf{Model-Avg} averages across model sizes, while \textbf{Method-Avg} averages the methods with a value in that row. The best-performing method in each row is highlighted in bold.}
\label{table:tpr_unsup:dclm}
\end{table*}

\begin{table*}[t]
\centering
\resizebox{0.99\textwidth}{!}{%
\begin{tabular}{@{}l@{\;\;\;}c@{\;\;}c@{\;\;}c@{\;\;}c@{\;\;}c@{\;\;}c@{\;\;}c@{\;\;}c@{\;\;}c@{\;\;}c@{\;\;}c@{\;\;}c@{\;\;}c@{\;\;}c@{\;\;\;}c@{}}
\toprule
\multirow{2}{*}{\textbf{Model}} & \multicolumn{10}{c}{\textbf{Likelihood-based}} & \textbf{Pert.} & \multicolumn{2}{c}{\textbf{Prompt}} & \textbf{Out.-Dist.} & \multirow{2}{*}{\makecell{\textbf{Method-}\\\textbf{Avg}}} \\
\cmidrule(lr){2-11} \cmidrule(lr){12-12} \cmidrule(lr){13-14} \cmidrule(lr){15-15}
 & \textbf{PPL} & \textbf{Zlib} & \textbf{Lower} & \textbf{MinK} & \textbf{MinK++} & \textbf{DCPDD} & \textbf{PAC} & \textbf{ReCaLL} & \textbf{EMMIA} & \textbf{CAMIA} & \textbf{Neigh.} & \textbf{Guided} & \textbf{DCQ} & \textbf{CDD} & \\
\midrule
\textbf{1B} & 0.06 & 0.07 & 0.05 & 0.09 & 0.07 & 0.07 & 0.06 & 0.05 & \textbf{0.18} & 0.06 & 0.04 & 0.05 & 0.04 & 0.05 & 0.07 \\
\textbf{7B} & 0.07 & 0.07 & 0.06 & 0.07 & 0.07 & 0.07 & 0.07 & 0.05 & \textbf{0.47} & 0.09 & 0.05 & 0.03 & 0.10 & 0.05 & 0.09 \\
\textbf{13B} & 0.07 & 0.07 & 0.08 & 0.07 & 0.05 & 0.09 & 0.10 & 0.05 & \textbf{0.44} & 0.05 & 0.07 & 0.05 & 0.07 & 0.05 & 0.09 \\
\textbf{32B} & 0.08 & 0.09 & 0.08 & 0.10 & 0.08 & 0.10 & 0.09 & 0.07 & \textbf{0.48} & 0.05 & 0.06 & 0.05 & 0.11 & 0.04 & 0.10 \\
\cmidrule(l){2-16}
\textbf{Model-Avg} & 0.07 & 0.07 & 0.07 & 0.08 & 0.07 & 0.08 & 0.08 & 0.05 & \textbf{0.39} & 0.06 & 0.05 & 0.05 & 0.08 & 0.05 & 0.09 \\
\bottomrule
\end{tabular}%
}
\caption{Comparison of unsupervised MIAs on the peS2o domain of \textsc{OLMo-Detect}, evaluated via TPR@5\% FPR. Methods are grouped by family as in Table~\ref{table:tpr_unsup:all}. \textbf{Model-Avg} averages across model sizes, while \textbf{Method-Avg} averages the methods with a value in that row. The best-performing method in each row is highlighted in bold.}
\label{table:tpr_unsup:pes2o}
\end{table*}

\begin{table*}[t]
\centering
\resizebox{0.99\textwidth}{!}{%
\begin{tabular}{@{}l@{\;\;\;}c@{\;\;}c@{\;\;}c@{\;\;}c@{\;\;}c@{\;\;}c@{\;\;}c@{\;\;}c@{\;\;}c@{\;\;}c@{\;\;}c@{\;\;}c@{\;\;}c@{\;\;}c@{\;\;\;}c@{}}
\toprule
\multirow{2}{*}{\textbf{Model}} & \multicolumn{10}{c}{\textbf{Likelihood-based}} & \textbf{Pert.} & \multicolumn{2}{c}{\textbf{Prompt}} & \textbf{Out.-Dist.} & \multirow{2}{*}{\makecell{\textbf{Method-}\\\textbf{Avg}}} \\
\cmidrule(lr){2-11} \cmidrule(lr){12-12} \cmidrule(lr){13-14} \cmidrule(lr){15-15}
 & \textbf{PPL} & \textbf{Zlib} & \textbf{Lower} & \textbf{MinK} & \textbf{MinK++} & \textbf{DCPDD} & \textbf{PAC} & \textbf{ReCaLL} & \textbf{EMMIA} & \textbf{CAMIA} & \textbf{Neigh.} & \textbf{Guided} & \textbf{DCQ} & \textbf{CDD} & \\
\midrule
\textbf{1B} & 0.08 & 0.17 & 0.07 & 0.08 & 0.13 & 0.16 & 0.17 & \textbf{0.17} & 0.09 & 0.11 & 0.05 & 0.01 & 0.16 & 0.07 & 0.11 \\
\textbf{7B} & 0.08 & \textbf{0.27} & 0.10 & 0.08 & 0.08 & 0.23 & 0.22 & 0.22 & 0.10 & 0.11 & 0.04 & 0.05 & 0.06 & 0.12 & 0.13 \\
\textbf{13B} & 0.09 & 0.24 & 0.08 & 0.09 & 0.07 & 0.20 & \textbf{0.27} & 0.09 & 0.11 & 0.10 & 0.07 & 0.03 & 0.01 & 0.07 & 0.11 \\
\textbf{32B} & 0.08 & 0.19 & 0.04 & 0.08 & 0.03 & 0.24 & \textbf{0.26} & 0.08 & 0.10 & 0.12 & 0.06 & 0.04 & 0.06 & 0.10 & 0.11 \\
\cmidrule(l){2-16}
\textbf{Model-Avg} & 0.08 & 0.22 & 0.07 & 0.08 & 0.08 & 0.21 & \textbf{0.23} & 0.14 & 0.10 & 0.11 & 0.06 & 0.03 & 0.07 & 0.09 & 0.11 \\
\bottomrule
\end{tabular}%
}
\caption{Comparison of unsupervised MIAs on the StarCoder domain of \textsc{OLMo-Detect}, evaluated via TPR@5\% FPR (macro-averaged over its subsets). Methods are grouped by family as in Table~\ref{table:tpr_unsup:all}. \textbf{Model-Avg} averages across model sizes, while \textbf{Method-Avg} averages the methods with a value in that row. The best-performing method in each row is highlighted in bold.}
\label{table:tpr_unsup:starcoder}
\end{table*}

\begin{table*}[t]
\centering
\resizebox{0.99\textwidth}{!}{%
\begin{tabular}{@{}l@{\;\;\;}c@{\;\;}c@{\;\;}c@{\;\;}c@{\;\;}c@{\;\;}c@{\;\;}c@{\;\;}c@{\;\;}c@{\;\;}c@{\;\;}c@{\;\;}c@{\;\;}c@{\;\;}c@{\;\;\;}c@{}}
\toprule
\multirow{2}{*}{\textbf{Model}} & \multicolumn{10}{c}{\textbf{Likelihood-based}} & \textbf{Pert.} & \multicolumn{2}{c}{\textbf{Prompt}} & \textbf{Out.-Dist.} & \multirow{2}{*}{\makecell{\textbf{Method-}\\\textbf{Avg}}} \\
\cmidrule(lr){2-11} \cmidrule(lr){12-12} \cmidrule(lr){13-14} \cmidrule(lr){15-15}
 & \textbf{PPL} & \textbf{Zlib} & \textbf{Lower} & \textbf{MinK} & \textbf{MinK++} & \textbf{DCPDD} & \textbf{PAC} & \textbf{ReCaLL} & \textbf{EMMIA} & \textbf{CAMIA} & \textbf{Neigh.} & \textbf{Guided} & \textbf{DCQ} & \textbf{CDD} & \\
\midrule
\textbf{1B} & 0.04 & 0.07 & 0.08 & 0.06 & 0.08 & \textbf{0.12} & 0.07 & 0.01 & 0.09 & 0.07 & 0.04 & 0.03 & 0.05 & 0.07 & 0.06 \\
\textbf{7B} & 0.04 & 0.07 & 0.07 & 0.06 & \textbf{0.14} & 0.12 & 0.02 & 0.01 & 0.09 & 0.06 & 0.07 & 0.04 & 0.00 & 0.04 & 0.06 \\
\textbf{13B} & 0.06 & 0.10 & 0.04 & 0.06 & 0.11 & 0.10 & 0.04 & 0.04 & \textbf{0.15} & 0.04 & 0.07 & 0.03 & 0.08 & 0.03 & 0.07 \\
\textbf{32B} & 0.10 & \textbf{0.17} & 0.04 & 0.07 & 0.15 & 0.13 & 0.11 & 0.03 & 0.10 & 0.04 & 0.07 & 0.02 & 0.04 & 0.02 & 0.08 \\
\cmidrule(l){2-16}
\textbf{Model-Avg} & 0.06 & 0.10 & 0.06 & 0.06 & \textbf{0.12} & 0.12 & 0.06 & 0.03 & 0.11 & 0.05 & 0.06 & 0.03 & 0.04 & 0.04 & 0.07 \\
\bottomrule
\end{tabular}%
}
\caption{Comparison of unsupervised MIAs on the OpenWebMath domain of \textsc{OLMo-Detect}, evaluated via TPR@5\% FPR. Methods are grouped by family as in Table~\ref{table:tpr_unsup:all}. \textbf{Model-Avg} averages across model sizes, while \textbf{Method-Avg} averages the methods with a value in that row. The best-performing method in each row is highlighted in bold.}
\label{table:tpr_unsup:openwebmath}
\end{table*}

\begin{table*}[t]
\centering
\resizebox{0.99\textwidth}{!}{%
\begin{tabular}{@{}l@{\;\;\;}c@{\;\;}c@{\;\;}c@{\;\;}c@{\;\;}c@{\;\;}c@{\;\;}c@{\;\;}c@{\;\;}c@{\;\;}c@{\;\;}c@{\;\;}c@{\;\;}c@{\;\;}c@{\;\;\;}c@{}}
\toprule
\multirow{2}{*}{\textbf{Model}} & \multicolumn{10}{c}{\textbf{Likelihood-based}} & \textbf{Pert.} & \multicolumn{2}{c}{\textbf{Prompt}} & \textbf{Out.-Dist.} & \multirow{2}{*}{\makecell{\textbf{Method-}\\\textbf{Avg}}} \\
\cmidrule(lr){2-11} \cmidrule(lr){12-12} \cmidrule(lr){13-14} \cmidrule(lr){15-15}
 & \textbf{PPL} & \textbf{Zlib} & \textbf{Lower} & \textbf{MinK} & \textbf{MinK++} & \textbf{DCPDD} & \textbf{PAC} & \textbf{ReCaLL} & \textbf{EMMIA} & \textbf{CAMIA} & \textbf{Neigh.} & \textbf{Guided} & \textbf{DCQ} & \textbf{CDD} & \\
\midrule
\textbf{1B} & 0.21 & \textbf{0.25} & 0.17 & 0.25 & 0.14 & 0.12 & 0.16 & 0.03 & 0.04 & 0.22 & 0.11 & 0.04 & 0.04 & 0.07 & 0.13 \\
\textbf{7B} & 0.43 & 0.54 & 0.21 & 0.48 & 0.13 & \textbf{0.60} & 0.37 & 0.00 & 0.05 & 0.49 & 0.22 & 0.05 & 0.06 & 0.07 & 0.26 \\
\textbf{13B} & 0.92 & \textbf{0.97} & 0.54 & 0.93 & 0.21 & 0.94 & 0.78 & 0.00 & 0.03 & 0.95 & 0.53 & 0.05 & 0.04 & 0.17 & 0.50 \\
\textbf{32B} & 0.89 & 0.96 & 0.61 & 0.84 & 0.14 & 0.92 & 0.67 & 0.00 & 0.06 & \textbf{0.96} & 0.63 & 0.12 & 0.06 & 0.24 & 0.51 \\
\cmidrule(l){2-16}
\textbf{Model-Avg} & 0.61 & \textbf{0.68} & 0.38 & 0.63 & 0.16 & 0.65 & 0.49 & 0.01 & 0.05 & 0.66 & 0.37 & 0.06 & 0.05 & 0.14 & 0.35 \\
\bottomrule
\end{tabular}%
}
\caption{Comparison of unsupervised MIAs on the GSM8K subset of \textsc{OLMo-Detect}'s Math Mix domain, evaluated via TPR@5\% FPR. Methods are grouped by family as in Table~\ref{table:tpr_unsup:all}. \textbf{Model-Avg} averages across model sizes, while \textbf{Method-Avg} averages the methods with a value in that row. The best-performing method in each row is highlighted in bold.}
\label{table:tpr_unsup:gsm8k}
\end{table*}

\begin{table*}[t]
\centering
\resizebox{0.99\textwidth}{!}{%
\begin{tabular}{@{}l@{\;\;\;}c@{\;\;}c@{\;\;}c@{\;\;}c@{\;\;}c@{\;\;}c@{\;\;}c@{\;\;}c@{\;\;}c@{\;\;}c@{\;\;}c@{\;\;}c@{\;\;}c@{\;\;}c@{\;\;\;}c@{}}
\toprule
\multirow{2}{*}{\textbf{Model}} & \multicolumn{10}{c}{\textbf{Likelihood-based}} & \textbf{Pert.} & \multicolumn{2}{c}{\textbf{Prompt}} & \textbf{Out.-Dist.} & \multirow{2}{*}{\makecell{\textbf{Method-}\\\textbf{Avg}}} \\
\cmidrule(lr){2-11} \cmidrule(lr){12-12} \cmidrule(lr){13-14} \cmidrule(lr){15-15}
 & \textbf{PPL} & \textbf{Zlib} & \textbf{Lower} & \textbf{MinK} & \textbf{MinK++} & \textbf{DCPDD} & \textbf{PAC} & \textbf{ReCaLL} & \textbf{EMMIA} & \textbf{CAMIA} & \textbf{Neigh.} & \textbf{Guided} & \textbf{DCQ} & \textbf{CDD} & \\
\midrule
\textbf{1B} & 0.06 & 0.08 & 0.12 & 0.05 & 0.03 & 0.08 & 0.12 & 0.10 & \textbf{0.13} & 0.07 & 0.10 & 0.07 & 0.10 & 0.06 & 0.08 \\
\textbf{7B} & 0.06 & 0.12 & 0.09 & 0.08 & 0.06 & \textbf{0.18} & 0.12 & 0.03 & 0.09 & 0.06 & 0.10 & 0.05 & 0.05 & 0.06 & 0.08 \\
\textbf{13B} & 0.07 & 0.17 & 0.13 & 0.08 & 0.07 & 0.16 & \textbf{0.19} & 0.10 & 0.10 & 0.05 & 0.10 & 0.07 & 0.06 & 0.05 & 0.10 \\
\textbf{32B} & 0.09 & 0.20 & 0.17 & 0.13 & 0.12 & 0.21 & \textbf{0.25} & 0.16 & 0.17 & 0.07 & 0.10 & 0.10 & 0.14 & 0.05 & 0.14 \\
\cmidrule(l){2-16}
\textbf{Model-Avg} & 0.07 & 0.14 & 0.13 & 0.08 & 0.07 & 0.16 & \textbf{0.17} & 0.10 & 0.12 & 0.06 & 0.10 & 0.07 & 0.09 & 0.05 & 0.10 \\
\bottomrule
\end{tabular}%
}
\caption{Comparison of unsupervised MIAs on the Stack Exchange domain of \textsc{OLMo-Detect}, evaluated via TPR@5\% FPR (macro-averaged over its subsets). Methods are grouped by family as in Table~\ref{table:tpr_unsup:all}. \textbf{Model-Avg} averages across model sizes, while \textbf{Method-Avg} averages the methods with a value in that row. The best-performing method in each row is highlighted in bold.}
\label{table:tpr_unsup:stackexchange}
\end{table*}

\begin{table*}[t]
\centering
\resizebox{0.99\textwidth}{!}{%
\begin{tabular}{@{}l@{\;\;\;}c@{\;\;}c@{\;\;}c@{\;\;}c@{\;\;}c@{\;\;}c@{\;\;}c@{\;\;}c@{\;\;}c@{\;\;}c@{\;\;}c@{\;\;}c@{\;\;}c@{\;\;}c@{\;\;\;}c@{}}
\toprule
\multirow{2}{*}{\textbf{Model}} & \multicolumn{10}{c}{\textbf{Likelihood-based}} & \textbf{Pert.} & \multicolumn{2}{c}{\textbf{Prompt}} & \textbf{Out.-Dist.} & \multirow{2}{*}{\makecell{\textbf{Method-}\\\textbf{Avg}}} \\
\cmidrule(lr){2-11} \cmidrule(lr){12-12} \cmidrule(lr){13-14} \cmidrule(lr){15-15}
 & \textbf{PPL} & \textbf{Zlib} & \textbf{Lower} & \textbf{MinK} & \textbf{MinK++} & \textbf{DCPDD} & \textbf{PAC} & \textbf{ReCaLL} & \textbf{EMMIA} & \textbf{CAMIA} & \textbf{Neigh.} & \textbf{Guided} & \textbf{DCQ} & \textbf{CDD} & \\
\midrule
\textbf{1B} & 0.22 & 0.14 & 0.09 & 0.21 & \textbf{0.33} & 0.12 & 0.08 & 0.12 & 0.04 & 0.15 & 0.07 & 0.05 & 0.05 & 0.05 & 0.12 \\
\textbf{7B} & 0.40 & 0.29 & 0.13 & 0.36 & \textbf{0.50} & 0.28 & 0.17 & 0.17 & 0.07 & 0.22 & 0.06 & 0.06 & 0.02 & 0.06 & 0.20 \\
\textbf{13B} & 0.19 & 0.14 & 0.07 & 0.18 & \textbf{0.26} & 0.10 & 0.06 & 0.11 & 0.06 & 0.14 & 0.09 & 0.10 & 0.07 & 0.08 & 0.12 \\
\textbf{32B} & 0.28 & 0.15 & 0.07 & 0.32 & \textbf{0.33} & 0.19 & 0.05 & 0.17 & 0.07 & 0.21 & 0.07 & 0.08 & 0.00 & 0.08 & 0.15 \\
\cmidrule(l){2-16}
\textbf{Model-Avg} & 0.27 & 0.18 & 0.09 & 0.27 & \textbf{0.36} & 0.17 & 0.09 & 0.14 & 0.06 & 0.18 & 0.07 & 0.07 & 0.04 & 0.07 & 0.15 \\
\bottomrule
\end{tabular}%
}
\caption{Comparison of unsupervised MIAs on the SFT domain of \textsc{OLMo-Detect}, evaluated via TPR@5\% FPR (macro-averaged over its subsets). Methods are grouped by family as in Table~\ref{table:tpr_unsup:all}. \textbf{Model-Avg} averages across model sizes, while \textbf{Method-Avg} averages the methods with a value in that row. The best-performing method in each row is highlighted in bold.}
\label{table:tpr_unsup:sft}
\end{table*}

\begin{table*}[t]
\centering
\resizebox{0.99\textwidth}{!}{%
\begin{tabular}{@{}l@{\;\;\;}c@{\;\;}c@{\;\;}c@{\;\;}c@{\;\;}c@{\;\;}c@{\;\;}c@{\;\;}c@{\;\;}c@{\;\;}c@{\;\;}c@{\;\;}c@{\;\;}c@{\;\;}c@{\;\;}c@{\;\;\;}c@{}}
\toprule
\multirow{2}{*}{\textbf{Model}} & \multicolumn{10}{c}{\textbf{Likelihood-based}} & \textbf{Pert.} & \multicolumn{2}{c}{\textbf{Prompt}} & \multicolumn{2}{c}{\textbf{Out.-Dist.}} & \multirow{2}{*}{\makecell{\textbf{Method-}\\\textbf{Avg}}} \\
\cmidrule(lr){2-11} \cmidrule(lr){12-12} \cmidrule(lr){13-14} \cmidrule(lr){15-16}
 & \textbf{PPL} & \textbf{Zlib} & \textbf{Lower} & \textbf{MinK} & \textbf{MinK++} & \textbf{DCPDD} & \textbf{PAC} & \textbf{ReCaLL} & \textbf{EMMIA} & \textbf{CAMIA} & \textbf{Neigh.} & \textbf{Guided} & \textbf{DCQ} & \textbf{CDD} & \textbf{SelfCrit}$^\dagger$ & \\
\midrule
\textbf{1B} & 0.07 & 0.06 & 0.14 & 0.08 & 0.04 & 0.07 & 0.12 & 0.16 & 0.08 & \textbf{0.22} & 0.04 & 0.04 & 0.03 & 0.10 & 0.12 & 0.09 \\
\textbf{7B} & 0.02 & 0.03 & 0.07 & 0.01 & 0.02 & 0.01 & 0.01 & 0.07 & 0.09 & 0.07 & 0.04 & 0.05 & 0.00 & \textbf{0.10} & 0.04 & 0.04 \\
\textbf{13B} & 0.03 & 0.06 & 0.28 & 0.04 & 0.04 & 0.03 & 0.06 & \textbf{0.33} & 0.10 & 0.08 & 0.03 & 0.03 & 0.06 & 0.11 & 0.06 & 0.09 \\
\textbf{32B} & 0.04 & 0.09 & 0.09 & 0.06 & 0.07 & 0.04 & 0.13 & \textbf{0.19} & 0.01 & 0.06 & 0.03 & 0.03 & 0.07 & 0.07 & 0.11 & 0.07 \\
\cmidrule(l){2-17}
\textbf{Model-Avg} & 0.04 & 0.06 & 0.15 & 0.05 & 0.04 & 0.04 & 0.08 & \textbf{0.19} & 0.07 & 0.11 & 0.03 & 0.04 & 0.04 & 0.09 & 0.08 & 0.07 \\
\bottomrule
\end{tabular}%
}
\caption{Comparison of unsupervised MIAs on the UF-Wildchat subset of \textsc{OLMo-Detect}'s DPO domain, evaluated via TPR@5\% FPR. Methods are grouped by family as in Table~\ref{table:tpr_unsup:all}. \textbf{Model-Avg} averages across model sizes, while \textbf{Method-Avg} averages the methods with a value in that row. The best-performing method in each row is highlighted in bold. $^\dagger$SelfCrit is evaluated only on DPO and RLVR.}
\label{table:tpr_unsup:dpo}
\end{table*}

\begin{table*}[t]
\centering
\resizebox{0.99\textwidth}{!}{%
\begin{tabular}{@{}l@{\;\;\;}c@{\;\;}c@{\;\;}c@{\;\;}c@{\;\;}c@{\;\;}c@{\;\;}c@{\;\;}c@{\;\;}c@{\;\;}c@{\;\;}c@{\;\;}c@{\;\;}c@{\;\;}c@{\;\;}c@{\;\;\;}c@{}}
\toprule
\multirow{2}{*}{\textbf{Model}} & \multicolumn{10}{c}{\textbf{Likelihood-based}} & \textbf{Pert.} & \multicolumn{2}{c}{\textbf{Prompt}} & \multicolumn{2}{c}{\textbf{Out.-Dist.}} & \multirow{2}{*}{\makecell{\textbf{Method-}\\\textbf{Avg}}} \\
\cmidrule(lr){2-11} \cmidrule(lr){12-12} \cmidrule(lr){13-14} \cmidrule(lr){15-16}
 & \textbf{PPL} & \textbf{Zlib} & \textbf{Lower} & \textbf{MinK} & \textbf{MinK++} & \textbf{DCPDD} & \textbf{PAC} & \textbf{ReCaLL} & \textbf{EMMIA} & \textbf{CAMIA} & \textbf{Neigh.} & \textbf{Guided} & \textbf{DCQ} & \textbf{CDD} & \textbf{SelfCrit}$^\dagger$ & \\
\midrule
\textbf{1B} & 0.29 & \textbf{0.35} & 0.12 & 0.33 & 0.08 & 0.16 & 0.19 & 0.02 & 0.15 & 0.14 & 0.03 & 0.06 & 0.04 & 0.04 & 0.02 & 0.14 \\
\textbf{7B} & \textbf{0.66} & 0.65 & 0.33 & 0.66 & 0.16 & 0.60 & 0.25 & 0.02 & 0.51 & 0.53 & 0.11 & 0.04 & 0.05 & 0.03 & 0.05 & 0.31 \\
\textbf{13B} & \textbf{0.82} & 0.81 & 0.41 & 0.78 & 0.06 & 0.71 & 0.30 & 0.00 & 0.52 & 0.62 & 0.11 & 0.02 & 0.05 & 0.05 & 0.05 & 0.36 \\
\textbf{32B} & 0.71 & 0.68 & 0.43 & \textbf{0.74} & 0.15 & 0.71 & 0.34 & 0.03 & 0.04 & 0.67 & 0.17 & 0.03 & 0.06 & 0.05 & 0.02 & 0.32 \\
\cmidrule(l){2-17}
\textbf{Model-Avg} & 0.62 & 0.62 & 0.32 & \textbf{0.63} & 0.11 & 0.54 & 0.27 & 0.02 & 0.30 & 0.49 & 0.10 & 0.04 & 0.05 & 0.05 & 0.04 & 0.28 \\
\bottomrule
\end{tabular}%
}
\caption{Comparison of unsupervised MIAs on the RLVR domain of \textsc{OLMo-Detect}, evaluated via TPR@5\% FPR (macro-averaged over its subsets). Methods are grouped by family as in Table~\ref{table:tpr_unsup:all}. \textbf{Model-Avg} averages across model sizes, while \textbf{Method-Avg} averages the methods with a value in that row. The best-performing method in each row is highlighted in bold. $^\dagger$SelfCrit is evaluated only on DPO and RLVR.}
\label{table:tpr_unsup:rlvr}
\end{table*}

\begin{table*}[t]
\centering
\resizebox{0.99\textwidth}{!}{%
\begin{tabular}{@{}l@{\;\;}l@{\;\;\;}c@{\;\;}c@{\;\;}c@{\;\;}c@{\;\;}c@{\;\;}c@{\;\;}c@{\;\;}c@{\;\;}c@{\;\;}c@{\;\;}c@{\;\;}c@{\;\;}c@{\;\;}c@{\;\;\;}c@{}}
\toprule
\multirow{2}{*}{\textbf{Subset}} & \multirow{2}{*}{\textbf{Model}} & \multicolumn{10}{c}{\textbf{Likelihood-based}} & \textbf{Pert.} & \multicolumn{2}{c}{\textbf{Prompt}} & \textbf{Out.-Dist.} & \multirow{2}{*}{\makecell{\textbf{Method-}\\\textbf{Avg}}} \\
\cmidrule(lr){3-12} \cmidrule(lr){13-13} \cmidrule(lr){14-15} \cmidrule(lr){16-16}
 &  & \textbf{PPL} & \textbf{Zlib} & \textbf{Lower} & \textbf{MinK} & \textbf{MinK++} & \textbf{DCPDD} & \textbf{PAC} & \textbf{ReCaLL} & \textbf{EMMIA} & \textbf{CAMIA} & \textbf{Neigh.} & \textbf{Guided} & \textbf{DCQ} & \textbf{CDD} & \\
\midrule
\multirow{5}{*}{\textbf{Python}} & 1B & 0.10 & \textbf{0.23} & 0.10 & 0.10 & 0.12 & 0.20 & 0.16 & 0.09 & 0.06 & 0.17 & 0.06 & 0.00 & 0.14 & 0.08 & 0.11 \\
 & 7B & 0.10 & 0.22 & 0.12 & 0.10 & 0.12 & 0.19 & \textbf{0.23} & 0.09 & 0.08 & 0.13 & 0.06 & 0.07 & 0.05 & 0.02 & 0.11 \\
 & 13B & 0.11 & 0.22 & 0.10 & 0.12 & 0.07 & 0.18 & \textbf{0.26} & 0.06 & 0.08 & 0.16 & 0.04 & 0.06 & 0.01 & 0.07 & 0.11 \\
 & 32B & 0.10 & 0.14 & 0.03 & 0.10 & 0.07 & 0.17 & \textbf{0.22} & 0.09 & 0.10 & 0.16 & 0.04 & 0.04 & 0.06 & 0.13 & 0.10 \\
\cmidrule(l){2-17}
 & \textit{Model-Avg} & 0.10 & 0.21 & 0.09 & 0.10 & 0.09 & 0.18 & \textbf{0.22} & 0.08 & 0.08 & 0.15 & 0.05 & 0.04 & 0.06 & 0.07 & 0.11 \\
\cmidrule(l){1-17}
\multirow{5}{*}{\textbf{Java}} & 1B & 0.06 & 0.10 & 0.04 & 0.06 & 0.14 & 0.12 & 0.18 & \textbf{0.26} & 0.12 & 0.06 & 0.04 & 0.02 & 0.18 & 0.07 & 0.10 \\
\bottomrule
\end{tabular}%
}
\caption{Comparison of unsupervised MIAs on the subsets of the StarCoder domain of \textsc{OLMo-Detect}, evaluated via TPR@5\% FPR. Methods are grouped by family as in Table~\ref{table:tpr_unsup:all}. \textbf{Model-Avg} averages across model sizes, while \textbf{Method-Avg} averages the methods with a value in that row. The best-performing method in each row is highlighted in bold.}
\label{table:tpr_unsup:starcoder_subsets}
\end{table*}

\begin{table*}[t]
\centering
\resizebox{0.99\textwidth}{!}{%
\begin{tabular}{@{}l@{\;\;}l@{\;\;\;}c@{\;\;}c@{\;\;}c@{\;\;}c@{\;\;}c@{\;\;}c@{\;\;}c@{\;\;}c@{\;\;}c@{\;\;}c@{\;\;}c@{\;\;}c@{\;\;}c@{\;\;}c@{\;\;\;}c@{}}
\toprule
\multirow{2}{*}{\textbf{Subset}} & \multirow{2}{*}{\textbf{Model}} & \multicolumn{10}{c}{\textbf{Likelihood-based}} & \textbf{Pert.} & \multicolumn{2}{c}{\textbf{Prompt}} & \textbf{Out.-Dist.} & \multirow{2}{*}{\makecell{\textbf{Method-}\\\textbf{Avg}}} \\
\cmidrule(lr){3-12} \cmidrule(lr){13-13} \cmidrule(lr){14-15} \cmidrule(lr){16-16}
 &  & \textbf{PPL} & \textbf{Zlib} & \textbf{Lower} & \textbf{MinK} & \textbf{MinK++} & \textbf{DCPDD} & \textbf{PAC} & \textbf{ReCaLL} & \textbf{EMMIA} & \textbf{CAMIA} & \textbf{Neigh.} & \textbf{Guided} & \textbf{DCQ} & \textbf{CDD} & \\
\midrule
\multirow{5}{*}{\textbf{Math Stack Exchange}} & 1B & 0.07 & 0.07 & 0.16 & 0.04 & 0.04 & 0.11 & \textbf{0.20} & 0.14 & 0.20 & 0.11 & 0.11 & 0.09 & 0.16 & 0.08 & 0.11 \\
 & 7B & 0.05 & 0.07 & 0.07 & 0.07 & 0.02 & \textbf{0.25} & 0.16 & 0.04 & 0.14 & 0.08 & 0.11 & 0.02 & 0.02 & 0.05 & 0.08 \\
 & 13B & 0.07 & 0.16 & 0.15 & 0.07 & 0.04 & 0.20 & \textbf{0.27} & 0.11 & 0.11 & 0.02 & 0.12 & 0.09 & 0.07 & 0.04 & 0.11 \\
 & 32B & 0.11 & 0.25 & 0.21 & 0.14 & 0.11 & 0.25 & \textbf{0.39} & 0.23 & 0.20 & 0.05 & 0.11 & 0.11 & 0.12 & 0.05 & 0.17 \\
\cmidrule(l){2-17}
 & \textit{Model-Avg} & 0.08 & 0.14 & 0.15 & 0.08 & 0.05 & 0.20 & \textbf{0.25} & 0.13 & 0.16 & 0.06 & 0.11 & 0.08 & 0.09 & 0.06 & 0.12 \\
\cmidrule(l){1-17}
\multirow{5}{*}{\textbf{Stack Overflow}} & 1B & 0.06 & \textbf{0.10} & 0.09 & 0.07 & 0.03 & 0.05 & 0.05 & 0.06 & 0.06 & 0.04 & 0.08 & 0.05 & 0.05 & 0.03 & 0.06 \\
\bottomrule
\end{tabular}%
}
\caption{Comparison of unsupervised MIAs on the subsets of the Stack Exchange domain of \textsc{OLMo-Detect}, evaluated via TPR@5\% FPR. Methods are grouped by family as in Table~\ref{table:tpr_unsup:all}. \textbf{Model-Avg} averages across model sizes, while \textbf{Method-Avg} averages the methods with a value in that row. The best-performing method in each row is highlighted in bold.}
\label{table:tpr_unsup:stackexchange_subsets}
\end{table*}

\begin{table*}[t]
\centering
\resizebox{0.99\textwidth}{!}{%
\begin{tabular}{@{}l@{\;\;}l@{\;\;\;}c@{\;\;}c@{\;\;}c@{\;\;}c@{\;\;}c@{\;\;}c@{\;\;}c@{\;\;}c@{\;\;}c@{\;\;}c@{\;\;}c@{\;\;}c@{\;\;}c@{\;\;}c@{\;\;\;}c@{}}
\toprule
\multirow{2}{*}{\textbf{Subset}} & \multirow{2}{*}{\textbf{Model}} & \multicolumn{10}{c}{\textbf{Likelihood-based}} & \textbf{Pert.} & \multicolumn{2}{c}{\textbf{Prompt}} & \textbf{Out.-Dist.} & \multirow{2}{*}{\makecell{\textbf{Method-}\\\textbf{Avg}}} \\
\cmidrule(lr){3-12} \cmidrule(lr){13-13} \cmidrule(lr){14-15} \cmidrule(lr){16-16}
 &  & \textbf{PPL} & \textbf{Zlib} & \textbf{Lower} & \textbf{MinK} & \textbf{MinK++} & \textbf{DCPDD} & \textbf{PAC} & \textbf{ReCaLL} & \textbf{EMMIA} & \textbf{CAMIA} & \textbf{Neigh.} & \textbf{Guided} & \textbf{DCQ} & \textbf{CDD} & \\
\midrule
\multirow{5}{*}{\textbf{Aya}} & 1B & 0.12 & 0.02 & 0.16 & 0.16 & \textbf{0.24} & 0.09 & 0.04 & 0.09 & 0.06 & 0.14 & 0.08 & 0.06 & 0.04 & 0.05 & 0.10 \\
 & 7B & \textbf{0.48} & 0.27 & 0.12 & 0.40 & 0.35 & 0.26 & 0.20 & 0.18 & 0.09 & 0.41 & 0.07 & 0.10 & 0.05 & 0.11 & 0.22 \\
 & 13B & \textbf{0.28} & 0.15 & 0.06 & 0.28 & 0.23 & 0.15 & 0.10 & 0.09 & 0.04 & 0.26 & 0.09 & 0.13 & 0.11 & 0.13 & 0.15 \\
 & 32B & 0.17 & 0.04 & 0.11 & 0.14 & 0.10 & 0.06 & 0.04 & \textbf{0.22} & 0.07 & 0.08 & 0.10 & 0.04 & 0.01 & 0.14 & 0.09 \\
\cmidrule(l){2-17}
 & \textit{Model-Avg} & \textbf{0.26} & 0.12 & 0.11 & 0.25 & 0.23 & 0.14 & 0.09 & 0.15 & 0.06 & 0.22 & 0.08 & 0.08 & 0.05 & 0.11 & 0.14 \\
\cmidrule(l){1-17}
\multirow{5}{*}{\textbf{WildChat}} & 1B & 0.31 & 0.25 & 0.02 & 0.26 & \textbf{0.42} & 0.14 & 0.12 & 0.14 & 0.03 & 0.16 & 0.06 & 0.05 & 0.05 & 0.05 & 0.15 \\
\bottomrule
\end{tabular}%
}
\caption{Comparison of unsupervised MIAs on the subsets of the SFT domain of \textsc{OLMo-Detect}, evaluated via TPR@5\% FPR. Methods are grouped by family as in Table~\ref{table:tpr_unsup:all}. \textbf{Model-Avg} averages across model sizes, while \textbf{Method-Avg} averages the methods with a value in that row. The best-performing method in each row is highlighted in bold.}
\label{table:tpr_unsup:sft_subsets}
\end{table*}

\begin{table*}[t]
\centering
\resizebox{0.99\textwidth}{!}{%
\begin{tabular}{@{}l@{\;\;}l@{\;\;\;}c@{\;\;}c@{\;\;}c@{\;\;}c@{\;\;}c@{\;\;}c@{\;\;}c@{\;\;}c@{\;\;}c@{\;\;}c@{\;\;}c@{\;\;}c@{\;\;}c@{\;\;}c@{\;\;}c@{\;\;\;}c@{}}
\toprule
\multirow{2}{*}{\textbf{Subset}} & \multirow{2}{*}{\textbf{Model}} & \multicolumn{10}{c}{\textbf{Likelihood-based}} & \textbf{Pert.} & \multicolumn{2}{c}{\textbf{Prompt}} & \multicolumn{2}{c}{\textbf{Out.-Dist.}} & \multirow{2}{*}{\makecell{\textbf{Method-}\\\textbf{Avg}}} \\
\cmidrule(lr){3-12} \cmidrule(lr){13-13} \cmidrule(lr){14-15} \cmidrule(lr){16-17}
 &  & \textbf{PPL} & \textbf{Zlib} & \textbf{Lower} & \textbf{MinK} & \textbf{MinK++} & \textbf{DCPDD} & \textbf{PAC} & \textbf{ReCaLL} & \textbf{EMMIA} & \textbf{CAMIA} & \textbf{Neigh.} & \textbf{Guided} & \textbf{DCQ} & \textbf{CDD} & \textbf{SelfCrit}$^\dagger$ & \\
\midrule
\multirow{5}{*}{\textbf{CoT-GSM8K}} & 1B & 0.45 & \textbf{0.55} & 0.22 & 0.43 & 0.10 & 0.19 & 0.30 & 0.00 & 0.25 & 0.14 & 0.01 & 0.06 & 0.04 & 0.03 & 0.02 & 0.19 \\
 & 7B & 0.94 & 0.96 & 0.46 & \textbf{0.97} & 0.24 & 0.91 & 0.38 & 0.00 & 0.97 & 0.95 & 0.11 & 0.02 & 0.07 & 0.04 & 0.03 & 0.47 \\
 & 13B & \textbf{1.00} & 1.00 & 0.59 & 1.00 & 0.09 & 0.96 & 0.41 & 0.00 & 1.00 & 0.97 & 0.06 & 0.03 & 0.02 & 0.05 & 0.02 & 0.48 \\
 & 32B & \textbf{1.00} & 1.00 & 0.63 & 1.00 & 0.21 & 0.98 & 0.52 & 0.00 & 0.05 & 0.95 & 0.16 & 0.03 & 0.09 & 0.06 & 0.01 & 0.45 \\
\cmidrule(l){2-18}
 & \textit{Model-Avg} & 0.85 & \textbf{0.88} & 0.48 & 0.85 & 0.16 & 0.76 & 0.41 & 0.00 & 0.57 & 0.75 & 0.09 & 0.03 & 0.05 & 0.04 & 0.02 & 0.40 \\
\cmidrule(l){1-18}
\multirow{5}{*}{\textbf{CoT-MATH}} & 1B & 0.14 & 0.15 & 0.03 & \textbf{0.22} & 0.05 & 0.13 & 0.08 & 0.05 & 0.05 & 0.14 & 0.06 & 0.05 & 0.04 & 0.06 & 0.02 & 0.08 \\
\bottomrule
\end{tabular}%
}
\caption{Comparison of unsupervised MIAs on the subsets of the RLVR domain of \textsc{OLMo-Detect}, evaluated via TPR@5\% FPR. Methods are grouped by family as in Table~\ref{table:tpr_unsup:all}. \textbf{Model-Avg} averages across model sizes, while \textbf{Method-Avg} averages the methods with a value in that row. The best-performing method in each row is highlighted in bold. $^\dagger$SelfCrit is evaluated only on DPO and RLVR.}
\label{table:tpr_unsup:rlvr_subsets}
\end{table*}

\section{Per-Domain AUC Results for Supervised MIAs on \textsc{OLMo-Detect}}
\label{appendix:auc_results_sup}

We report per-domain AUCs for the supervised MIAs in Tables~\ref{table:auc_sup_dclm}--\ref{table:auc_sup_rlvr}, one table per domain, complementing the per-stage results of Table~\ref{table:olmo_detect_overall_supervised}. Each table follows the two-part layout of Table~\ref{table:olmo_detect_overall_supervised}: the top part gives the AUCs of Sup-CAMIA and MIA-Tuner under both settings, with Zlib repeated as an unsupervised reference, and the bottom part gives FSD's AUC gain over its base MIA, since FSD is only meaningful relative to the detector it is built on. \textbf{In-domain} trains on the evaluated domain's development set, and \textbf{LODO} trains on the other eight domains' development sets.

Two patterns hold across domains. First, supervision rarely pays: the better of Sup-CAMIA and MIA-Tuner exceeds Zlib on only five of the nine domains, ties on OpenWebMath, and falls below it on DCLM-Baseline ($-0.05$), SFT ($-0.03$) and RLVR ($-0.03$); the one substantial gain is on DPO ($+0.10$), the domain where Zlib is weakest at $0.48$. Second, fine-tuning on non-members seldom helps its base detector: of the $45$ domain-by-base-MIA combinations, FSD improves on its base in $13$ under In-domain and $11$ under LODO, and its largest losses reach $-0.10$ or worse on DCLM-Baseline, where every base detector degrades under both settings.

\begin{table*}[t]
\centering
\resizebox{0.7\textwidth}{!}{%
\begin{tabular}{@{}l@{\;\;}l@{\;\;\;}c@{\;\;}c@{\;\;}c@{\;\;}c@{\;\;\;}c@{}}
\toprule
\textbf{Method} & \textbf{Setting} & \textbf{1B} & \textbf{7B} & \textbf{13B} & \textbf{32B} & \textbf{Model-Avg} \\
\midrule
\textit{Zlib (unsup.)} & --- & 0.59 & 0.64 & 0.65 & 0.67 & 0.64 \\
\cmidrule(l){1-7}
Sup-CAMIA & In-domain & 0.51 & 0.55 & 0.54 & 0.56 & 0.54 \\
 & LODO & \textbf{0.55} & \textbf{0.64} & \textbf{0.58} & \textbf{0.57} & \textbf{0.58} \\
\cmidrule(l){1-7}
MIA-Tuner & In-domain & 0.45 & 0.45 & 0.53 & 0.55 & 0.50 \\
 & LODO & 0.47 & 0.51 & 0.49 & 0.47 & 0.49 \\
\bottomrule
\end{tabular}%
}

\vspace{4pt}

\resizebox{0.9\textwidth}{!}{%
\begin{tabular}{@{}lc@{\;\;}c@{\;\;}c@{\;\;}c@{\;\;}c@{\;\;}c@{\;\;\;}c@{\;\;}c@{\;\;}c@{\;\;}c@{\;\;}c@{}}
\toprule
\multirow{2.5}{*}{\textbf{Method}} & \multirow{2.5}{*}{\textbf{Base AUC}} & \multicolumn{5}{c}{\textbf{In-domain $\Delta$AUC}} & \multicolumn{5}{c}{\textbf{LODO $\Delta$AUC}} \\
\cmidrule(lr){3-7} \cmidrule(lr){8-12}
 & & 1B & 7B & 13B & 32B & \textbf{Model-Avg} & 1B & 7B & 13B & 32B & \textbf{Model-Avg} \\
\midrule
\textbf{FSD (PPL)} & 0.64 & $-$0.02 & $-$0.06 & $-$0.06 & $-$0.05 & $-$0.05 & $-$0.10 & $-$0.04 & $-$0.10 & $-$0.18 & $-$0.10 \\
\textbf{FSD (Zlib)} & 0.64 & $-$0.01 & $-$0.06 & $-$0.04 & $-$0.05 & $-$0.04 & $-$0.09 & $-$0.05 & $-$0.08 & $-$0.19 & $-$0.10 \\
\textbf{FSD (Lower)} & 0.63 & $-$0.12 & $-$0.09 & $-$0.12 & $-$0.09 & $-$0.10 & $-$0.09 & $-$0.11 & $-$0.11 & $-$0.13 & $-$0.11 \\
\textbf{FSD (MinK)} & 0.61 & $-$0.07 & $-$0.07 & $-$0.07 & $-$0.01 & $-$0.05 & $-$0.08 & $-$0.04 & $-$0.07 & $-$0.18 & $-$0.09 \\
\textbf{FSD (MinK++)} & 0.53 & $-$0.03 & $-$0.02 & $-$0.01 & $-$0.03 & $-$0.02 & $-$0.01 & $-$0.03 & $-$0.04 & $-$0.15 & $-$0.06 \\
\bottomrule
\end{tabular}%
}
\caption{Supervised MIAs on the DCLM-Baseline domain of \textsc{OLMo-Detect}, evaluated via AUC. \textbf{Top:} AUCs of Sup-CAMIA and MIA-Tuner under the \textbf{In-domain} setting (trained on this domain's dev set) and the \textbf{LODO} setting (trained on the other eight domains' dev sets); Zlib, the strongest unsupervised MIA, is repeated as a reference and is excluded from the bold, which marks the best supervised entry per column. \textbf{Bottom:} FSD measures the change in a base MIA's score after fine-tuning the target model on non-members, so we report its AUC gain ($\Delta$AUC) over the base MIA (in parentheses) on this domain. \textbf{Base AUC}: the base MIA's AUC on this domain averaged across model sizes; positive gains are bolded. \textbf{Model-Avg} averages across model sizes.}
\label{table:auc_sup_dclm}
\end{table*}

\begin{table*}[t]
\centering
\resizebox{0.7\textwidth}{!}{%
\begin{tabular}{@{}l@{\;\;}l@{\;\;\;}c@{\;\;}c@{\;\;}c@{\;\;}c@{\;\;\;}c@{}}
\toprule
\textbf{Method} & \textbf{Setting} & \textbf{1B} & \textbf{7B} & \textbf{13B} & \textbf{32B} & \textbf{Model-Avg} \\
\midrule
\textit{Zlib (unsup.)} & --- & 0.49 & 0.49 & 0.50 & 0.51 & 0.50 \\
\cmidrule(l){1-7}
Sup-CAMIA & In-domain & \textbf{0.55} & 0.47 & 0.52 & 0.50 & 0.51 \\
 & LODO & 0.50 & 0.52 & 0.48 & 0.55 & 0.51 \\
\cmidrule(l){1-7}
MIA-Tuner & In-domain & 0.54 & \textbf{0.57} & \textbf{0.56} & 0.57 & \textbf{0.56} \\
 & LODO & 0.55 & 0.51 & 0.39 & \textbf{0.57} & 0.51 \\
\bottomrule
\end{tabular}%
}

\vspace{4pt}

\resizebox{0.9\textwidth}{!}{%
\begin{tabular}{@{}lc@{\;\;}c@{\;\;}c@{\;\;}c@{\;\;}c@{\;\;}c@{\;\;\;}c@{\;\;}c@{\;\;}c@{\;\;}c@{\;\;}c@{}}
\toprule
\multirow{2.5}{*}{\textbf{Method}} & \multirow{2.5}{*}{\textbf{Base AUC}} & \multicolumn{5}{c}{\textbf{In-domain $\Delta$AUC}} & \multicolumn{5}{c}{\textbf{LODO $\Delta$AUC}} \\
\cmidrule(lr){3-7} \cmidrule(lr){8-12}
 & & 1B & 7B & 13B & 32B & \textbf{Model-Avg} & 1B & 7B & 13B & 32B & \textbf{Model-Avg} \\
\midrule
\textbf{FSD (PPL)} & 0.48 & \textbf{$+$0.07} & \textbf{$+$0.04} & \textbf{$+$0.05} & \textbf{$+$0.03} & \textbf{$+$0.05} & \textbf{$+$0.08} & \textbf{$+$0.05} & \textbf{$+$0.08} & \textbf{$+$0.07} & \textbf{$+$0.07} \\
\textbf{FSD (Zlib)} & 0.50 & \textbf{$+$0.05} & \textbf{$+$0.02} & \textbf{$+$0.01} & $-$0.01 & \textbf{$+$0.02} & \textbf{$+$0.04} & \textbf{$+$0.03} & \textbf{$+$0.05} & \textbf{$+$0.05} & \textbf{$+$0.04} \\
\textbf{FSD (Lower)} & 0.49 & \textbf{$+$0.07} & \textbf{$+$0.04} & $-$0.00 & \textbf{$+$0.02} & \textbf{$+$0.03} & \textbf{$+$0.04} & \textbf{$+$0.00} & $-$0.01 & \textbf{$+$0.00} & \textbf{$+$0.01} \\
\textbf{FSD (MinK)} & 0.50 & \textbf{$+$0.06} & \textbf{$+$0.01} & $-$0.00 & \textbf{$+$0.02} & \textbf{$+$0.02} & \textbf{$+$0.05} & \textbf{$+$0.01} & \textbf{$+$0.04} & \textbf{$+$0.04} & \textbf{$+$0.03} \\
\textbf{FSD (MinK++)} & 0.50 & \textbf{$+$0.04} & \textbf{$+$0.01} & $-$0.01 & \textbf{$+$0.02} & \textbf{$+$0.01} & \textbf{$+$0.01} & \textbf{$+$0.01} & $-$0.01 & \textbf{$+$0.06} & \textbf{$+$0.02} \\
\bottomrule
\end{tabular}%
}
\caption{Supervised MIAs on the peS2o domain of \textsc{OLMo-Detect}, evaluated via AUC. \textbf{Top:} AUCs of Sup-CAMIA and MIA-Tuner under the \textbf{In-domain} setting (trained on this domain's dev set) and the \textbf{LODO} setting (trained on the other eight domains' dev sets); Zlib, the strongest unsupervised MIA, is repeated as a reference and is excluded from the bold, which marks the best supervised entry per column. \textbf{Bottom:} FSD measures the change in a base MIA's score after fine-tuning the target model on non-members, so we report its AUC gain ($\Delta$AUC) over the base MIA (in parentheses) on this domain. \textbf{Base AUC}: the base MIA's AUC on this domain averaged across model sizes; positive gains are bolded. \textbf{Model-Avg} averages across model sizes.}
\label{table:auc_sup_pes2o}
\end{table*}

\begin{table*}[t]
\centering
\resizebox{0.7\textwidth}{!}{%
\begin{tabular}{@{}l@{\;\;}l@{\;\;\;}c@{\;\;}c@{\;\;}c@{\;\;}c@{\;\;\;}c@{}}
\toprule
\textbf{Method} & \textbf{Setting} & \textbf{1B} & \textbf{7B} & \textbf{13B} & \textbf{32B} & \textbf{Model-Avg} \\
\midrule
\textit{Zlib (unsup.)} & --- & 0.56 & 0.57 & 0.58 & 0.64 & 0.59 \\
\cmidrule(l){1-7}
Sup-CAMIA & In-domain & \textbf{0.60} & \textbf{0.59} & \textbf{0.59} & \textbf{0.58} & \textbf{0.59} \\
 & LODO & 0.54 & 0.56 & 0.53 & 0.56 & 0.55 \\
\cmidrule(l){1-7}
MIA-Tuner & In-domain & 0.57 & 0.58 & 0.52 & 0.54 & 0.55 \\
 & LODO & 0.59 & 0.54 & 0.56 & 0.51 & 0.55 \\
\bottomrule
\end{tabular}%
}

\vspace{4pt}

\resizebox{0.9\textwidth}{!}{%
\begin{tabular}{@{}lc@{\;\;}c@{\;\;}c@{\;\;}c@{\;\;}c@{\;\;}c@{\;\;\;}c@{\;\;}c@{\;\;}c@{\;\;}c@{\;\;}c@{}}
\toprule
\multirow{2.5}{*}{\textbf{Method}} & \multirow{2.5}{*}{\textbf{Base AUC}} & \multicolumn{5}{c}{\textbf{In-domain $\Delta$AUC}} & \multicolumn{5}{c}{\textbf{LODO $\Delta$AUC}} \\
\cmidrule(lr){3-7} \cmidrule(lr){8-12}
 & & 1B & 7B & 13B & 32B & \textbf{Model-Avg} & 1B & 7B & 13B & 32B & \textbf{Model-Avg} \\
\midrule
\textbf{FSD (PPL)} & 0.53 & \textbf{$+$0.04} & \textbf{$+$0.03} & \textbf{$+$0.07} & \textbf{$+$0.09} & \textbf{$+$0.06} & \textbf{$+$0.03} & \textbf{$+$0.04} & \textbf{$+$0.02} & \textbf{$+$0.02} & \textbf{$+$0.03} \\
\textbf{FSD (Zlib)} & 0.59 & \textbf{$+$0.02} & \textbf{$+$0.00} & \textbf{$+$0.02} & \textbf{$+$0.01} & \textbf{$+$0.01} & \textbf{$+$0.00} & $-$0.01 & $-$0.05 & $-$0.04 & $-$0.02 \\
\textbf{FSD (Lower)} & 0.60 & $-$0.04 & $-$0.04 & $-$0.01 & $-$0.07 & $-$0.04 & $-$0.10 & $-$0.04 & $-$0.11 & $-$0.07 & $-$0.08 \\
\textbf{FSD (MinK)} & 0.56 & \textbf{$+$0.01} & $-$0.00 & \textbf{$+$0.02} & \textbf{$+$0.04} & \textbf{$+$0.02} & \textbf{$+$0.01} & \textbf{$+$0.01} & $-$0.02 & $-$0.02 & $-$0.01 \\
\textbf{FSD (MinK++)} & 0.52 & $-$0.05 & $-$0.01 & \textbf{$+$0.02} & \textbf{$+$0.00} & $-$0.01 & $-$0.01 & $-$0.00 & $-$0.04 & $-$0.07 & $-$0.03 \\
\bottomrule
\end{tabular}%
}
\caption{Supervised MIAs on the OpenWebMath domain of \textsc{OLMo-Detect}, evaluated via AUC. \textbf{Top:} AUCs of Sup-CAMIA and MIA-Tuner under the \textbf{In-domain} setting (trained on this domain's dev set) and the \textbf{LODO} setting (trained on the other eight domains' dev sets); Zlib, the strongest unsupervised MIA, is repeated as a reference and is excluded from the bold, which marks the best supervised entry per column. \textbf{Bottom:} FSD measures the change in a base MIA's score after fine-tuning the target model on non-members, so we report its AUC gain ($\Delta$AUC) over the base MIA (in parentheses) on this domain. \textbf{Base AUC}: the base MIA's AUC on this domain averaged across model sizes; positive gains are bolded. \textbf{Model-Avg} averages across model sizes.}
\label{table:auc_sup_openwebmath}
\end{table*}

\begin{table*}[t]
\centering
\resizebox{0.7\textwidth}{!}{%
\begin{tabular}{@{}l@{\;\;}l@{\;\;\;}c@{\;\;}c@{\;\;}c@{\;\;}c@{\;\;\;}c@{}}
\toprule
\textbf{Method} & \textbf{Setting} & \textbf{1B} & \textbf{7B} & \textbf{13B} & \textbf{32B} & \textbf{Model-Avg} \\
\midrule
\textit{Zlib (unsup.)} & --- & 0.65 & 0.65 & 0.64 & 0.63 & 0.64 \\
\cmidrule(l){1-7}
Sup-CAMIA & In-domain & 0.57 & 0.54 & 0.56 & 0.56 & 0.56 \\
 & LODO & \textbf{0.64} & \textbf{0.69} & \textbf{0.67} & \textbf{0.67} & \textbf{0.67} \\
\cmidrule(l){1-7}
MIA-Tuner & In-domain & 0.61 & 0.54 & 0.53 & 0.48 & 0.54 \\
 & LODO & 0.46 & 0.55 & 0.44 & 0.50 & 0.49 \\
\bottomrule
\end{tabular}%
}

\vspace{4pt}

\resizebox{0.9\textwidth}{!}{%
\begin{tabular}{@{}lc@{\;\;}c@{\;\;}c@{\;\;}c@{\;\;}c@{\;\;}c@{\;\;\;}c@{\;\;}c@{\;\;}c@{\;\;}c@{\;\;}c@{}}
\toprule
\multirow{2.5}{*}{\textbf{Method}} & \multirow{2.5}{*}{\textbf{Base AUC}} & \multicolumn{5}{c}{\textbf{In-domain $\Delta$AUC}} & \multicolumn{5}{c}{\textbf{LODO $\Delta$AUC}} \\
\cmidrule(lr){3-7} \cmidrule(lr){8-12}
 & & 1B & 7B & 13B & 32B & \textbf{Model-Avg} & 1B & 7B & 13B & 32B & \textbf{Model-Avg} \\
\midrule
\textbf{FSD (PPL)} & 0.59 & \textbf{$+$0.03} & \textbf{$+$0.01} & $-$0.00 & $-$0.02 & \textbf{$+$0.01} & \textbf{$+$0.02} & $-$0.01 & $-$0.11 & $-$0.01 & $-$0.03 \\
\textbf{FSD (Zlib)} & 0.64 & $-$0.02 & $-$0.01 & $-$0.01 & $-$0.05 & $-$0.02 & $-$0.03 & $-$0.06 & $-$0.16 & $-$0.03 & $-$0.07 \\
\textbf{FSD (Lower)} & 0.60 & \textbf{$+$0.02} & $-$0.00 & $-$0.02 & $-$0.02 & $-$0.01 & \textbf{$+$0.04} & $-$0.03 & $-$0.03 & $-$0.05 & $-$0.02 \\
\textbf{FSD (MinK)} & 0.61 & $-$0.03 & $-$0.04 & $-$0.06 & $-$0.09 & $-$0.05 & $-$0.05 & $-$0.01 & $-$0.12 & $-$0.04 & $-$0.05 \\
\textbf{FSD (MinK++)} & 0.49 & \textbf{$+$0.00} & $-$0.04 & $-$0.00 & $-$0.01 & $-$0.01 & \textbf{$+$0.03} & \textbf{$+$0.06} & $-$0.02 & \textbf{$+$0.02} & \textbf{$+$0.02} \\
\bottomrule
\end{tabular}%
}
\caption{Supervised MIAs on the StarCoder domain of \textsc{OLMo-Detect} (macro-averaged over its subsets), evaluated via AUC. \textbf{Top:} AUCs of Sup-CAMIA and MIA-Tuner under the \textbf{In-domain} setting (trained on this domain's dev set) and the \textbf{LODO} setting (trained on the other eight domains' dev sets); Zlib, the strongest unsupervised MIA, is repeated as a reference and is excluded from the bold, which marks the best supervised entry per column. \textbf{Bottom:} FSD measures the change in a base MIA's score after fine-tuning the target model on non-members, so we report its AUC gain ($\Delta$AUC) over the base MIA (in parentheses) on this domain. \textbf{Base AUC}: the base MIA's AUC on this domain averaged across model sizes; positive gains are bolded. \textbf{Model-Avg} averages across model sizes.}
\label{table:auc_sup_starcoder}
\end{table*}

\begin{table*}[t]
\centering
\resizebox{0.7\textwidth}{!}{%
\begin{tabular}{@{}l@{\;\;}l@{\;\;\;}c@{\;\;}c@{\;\;}c@{\;\;}c@{\;\;\;}c@{}}
\toprule
\textbf{Method} & \textbf{Setting} & \textbf{1B} & \textbf{7B} & \textbf{13B} & \textbf{32B} & \textbf{Model-Avg} \\
\midrule
\textit{Zlib (unsup.)} & --- & 0.76 & 0.89 & 0.99 & 0.99 & 0.91 \\
\cmidrule(l){1-7}
Sup-CAMIA & In-domain & \textbf{0.82} & \textbf{0.94} & \textbf{1.00} & \textbf{1.00} & \textbf{0.94} \\
 & LODO & 0.68 & 0.85 & 0.77 & 0.90 & 0.80 \\
\cmidrule(l){1-7}
MIA-Tuner & In-domain & 0.51 & 0.51 & 0.50 & 0.51 & 0.51 \\
 & LODO & 0.49 & 0.51 & 0.51 & 0.52 & 0.51 \\
\bottomrule
\end{tabular}%
}

\vspace{4pt}

\resizebox{0.9\textwidth}{!}{%
\begin{tabular}{@{}lc@{\;\;}c@{\;\;}c@{\;\;}c@{\;\;}c@{\;\;}c@{\;\;\;}c@{\;\;}c@{\;\;}c@{\;\;}c@{\;\;}c@{}}
\toprule
\multirow{2.5}{*}{\textbf{Method}} & \multirow{2.5}{*}{\textbf{Base AUC}} & \multicolumn{5}{c}{\textbf{In-domain $\Delta$AUC}} & \multicolumn{5}{c}{\textbf{LODO $\Delta$AUC}} \\
\cmidrule(lr){3-7} \cmidrule(lr){8-12}
 & & 1B & 7B & 13B & 32B & \textbf{Model-Avg} & 1B & 7B & 13B & 32B & \textbf{Model-Avg} \\
\midrule
\textbf{FSD (PPL)} & 0.89 & $-$0.05 & $-$0.08 & \textbf{$+$0.01} & $-$0.06 & $-$0.04 & $-$0.16 & $-$0.10 & $-$0.01 & $-$0.14 & $-$0.10 \\
\textbf{FSD (Zlib)} & 0.91 & $-$0.07 & $-$0.10 & \textbf{$+$0.00} & $-$0.06 & $-$0.06 & $-$0.19 & $-$0.12 & $-$0.01 & $-$0.13 & $-$0.11 \\
\textbf{FSD (Lower)} & 0.85 & $-$0.03 & \textbf{$+$0.01} & \textbf{$+$0.01} & \textbf{$+$0.01} & $-$0.00 & $-$0.13 & $-$0.05 & $-$0.04 & $-$0.04 & $-$0.06 \\
\textbf{FSD (MinK)} & 0.89 & $-$0.12 & $-$0.27 & $-$0.06 & $-$0.13 & $-$0.15 & $-$0.19 & $-$0.08 & $-$0.01 & $-$0.12 & $-$0.10 \\
\textbf{FSD (MinK++)} & 0.55 & $-$0.02 & $-$0.02 & $-$0.02 & $-$0.09 & $-$0.04 & $-$0.06 & \textbf{$+$0.00} & $-$0.00 & $-$0.05 & $-$0.03 \\
\bottomrule
\end{tabular}%
}
\caption{Supervised MIAs on the GSM8K subset of \textsc{OLMo-Detect}'s Math Mix domain, evaluated via AUC. \textbf{Top:} AUCs of Sup-CAMIA and MIA-Tuner under the \textbf{In-domain} setting (trained on this domain's dev set) and the \textbf{LODO} setting (trained on the other eight domains' dev sets); Zlib, the strongest unsupervised MIA, is repeated as a reference and is excluded from the bold, which marks the best supervised entry per column. \textbf{Bottom:} FSD measures the change in a base MIA's score after fine-tuning the target model on non-members, so we report its AUC gain ($\Delta$AUC) over the base MIA (in parentheses) on this domain. \textbf{Base AUC}: the base MIA's AUC on this domain averaged across model sizes; positive gains are bolded. \textbf{Model-Avg} averages across model sizes.}
\label{table:auc_sup_gsm8k}
\end{table*}

\begin{table*}[t]
\centering
\resizebox{0.7\textwidth}{!}{%
\begin{tabular}{@{}l@{\;\;}l@{\;\;\;}c@{\;\;}c@{\;\;}c@{\;\;}c@{\;\;\;}c@{}}
\toprule
\textbf{Method} & \textbf{Setting} & \textbf{1B} & \textbf{7B} & \textbf{13B} & \textbf{32B} & \textbf{Model-Avg} \\
\midrule
\textit{Zlib (unsup.)} & --- & 0.59 & 0.64 & 0.67 & 0.70 & 0.65 \\
\cmidrule(l){1-7}
Sup-CAMIA & In-domain & 0.61 & 0.61 & \textbf{0.72} & \textbf{0.68} & 0.66 \\
 & LODO & 0.59 & 0.61 & 0.56 & 0.50 & 0.56 \\
\cmidrule(l){1-7}
MIA-Tuner & In-domain & \textbf{0.69} & \textbf{0.72} & 0.70 & 0.55 & \textbf{0.66} \\
 & LODO & 0.51 & 0.46 & 0.44 & 0.54 & 0.49 \\
\bottomrule
\end{tabular}%
}

\vspace{4pt}

\resizebox{0.9\textwidth}{!}{%
\begin{tabular}{@{}lc@{\;\;}c@{\;\;}c@{\;\;}c@{\;\;}c@{\;\;}c@{\;\;\;}c@{\;\;}c@{\;\;}c@{\;\;}c@{\;\;}c@{}}
\toprule
\multirow{2.5}{*}{\textbf{Method}} & \multirow{2.5}{*}{\textbf{Base AUC}} & \multicolumn{5}{c}{\textbf{In-domain $\Delta$AUC}} & \multicolumn{5}{c}{\textbf{LODO $\Delta$AUC}} \\
\cmidrule(lr){3-7} \cmidrule(lr){8-12}
 & & 1B & 7B & 13B & 32B & \textbf{Model-Avg} & 1B & 7B & 13B & 32B & \textbf{Model-Avg} \\
\midrule
\textbf{FSD (PPL)} & 0.59 & \textbf{$+$0.01} & \textbf{$+$0.06} & \textbf{$+$0.11} & $-$0.00 & \textbf{$+$0.05} & $-$0.08 & \textbf{$+$0.03} & \textbf{$+$0.01} & $-$0.07 & $-$0.03 \\
\textbf{FSD (Zlib)} & 0.65 & $-$0.02 & \textbf{$+$0.03} & \textbf{$+$0.03} & $-$0.08 & $-$0.01 & $-$0.11 & $-$0.02 & $-$0.08 & $-$0.14 & $-$0.09 \\
\textbf{FSD (Lower)} & 0.53 & $-$0.01 & $-$0.01 & $-$0.03 & $-$0.07 & $-$0.03 & \textbf{$+$0.01} & \textbf{$+$0.06} & \textbf{$+$0.08} & $-$0.08 & \textbf{$+$0.02} \\
\textbf{FSD (MinK)} & 0.60 & \textbf{$+$0.01} & \textbf{$+$0.03} & \textbf{$+$0.04} & $-$0.01 & \textbf{$+$0.02} & $-$0.08 & \textbf{$+$0.01} & $-$0.01 & $-$0.08 & $-$0.04 \\
\textbf{FSD (MinK++)} & 0.57 & \textbf{$+$0.00} & \textbf{$+$0.00} & $-$0.02 & $-$0.03 & $-$0.01 & $-$0.03 & \textbf{$+$0.01} & $-$0.02 & $-$0.03 & $-$0.02 \\
\bottomrule
\end{tabular}%
}
\caption{Supervised MIAs on the Stack Exchange domain of \textsc{OLMo-Detect} (macro-averaged over its subsets), evaluated via AUC. \textbf{Top:} AUCs of Sup-CAMIA and MIA-Tuner under the \textbf{In-domain} setting (trained on this domain's dev set) and the \textbf{LODO} setting (trained on the other eight domains' dev sets); Zlib, the strongest unsupervised MIA, is repeated as a reference and is excluded from the bold, which marks the best supervised entry per column. \textbf{Bottom:} FSD measures the change in a base MIA's score after fine-tuning the target model on non-members, so we report its AUC gain ($\Delta$AUC) over the base MIA (in parentheses) on this domain. \textbf{Base AUC}: the base MIA's AUC on this domain averaged across model sizes; positive gains are bolded. \textbf{Model-Avg} averages across model sizes.}
\label{table:auc_sup_stackexchange}
\end{table*}

\begin{table*}[t]
\centering
\resizebox{0.7\textwidth}{!}{%
\begin{tabular}{@{}l@{\;\;}l@{\;\;\;}c@{\;\;}c@{\;\;}c@{\;\;}c@{\;\;\;}c@{}}
\toprule
\textbf{Method} & \textbf{Setting} & \textbf{1B} & \textbf{7B} & \textbf{13B} & \textbf{32B} & \textbf{Model-Avg} \\
\midrule
\textit{Zlib (unsup.)} & --- & 0.64 & 0.70 & 0.61 & 0.64 & 0.65 \\
\cmidrule(l){1-7}
Sup-CAMIA & In-domain & 0.51 & \textbf{0.73} & \textbf{0.66} & \textbf{0.55} & \textbf{0.61} \\
 & LODO & \textbf{0.55} & 0.61 & 0.56 & 0.49 & 0.55 \\
\cmidrule(l){1-7}
MIA-Tuner & In-domain & 0.53 & 0.55 & 0.50 & 0.52 & 0.53 \\
 & LODO & 0.50 & 0.46 & 0.48 & 0.51 & 0.49 \\
\bottomrule
\end{tabular}%
}

\vspace{4pt}

\resizebox{0.9\textwidth}{!}{%
\begin{tabular}{@{}lc@{\;\;}c@{\;\;}c@{\;\;}c@{\;\;}c@{\;\;}c@{\;\;\;}c@{\;\;}c@{\;\;}c@{\;\;}c@{\;\;}c@{}}
\toprule
\multirow{2.5}{*}{\textbf{Method}} & \multirow{2.5}{*}{\textbf{Base AUC}} & \multicolumn{5}{c}{\textbf{In-domain $\Delta$AUC}} & \multicolumn{5}{c}{\textbf{LODO $\Delta$AUC}} \\
\cmidrule(lr){3-7} \cmidrule(lr){8-12}
 & & 1B & 7B & 13B & 32B & \textbf{Model-Avg} & 1B & 7B & 13B & 32B & \textbf{Model-Avg} \\
\midrule
\textbf{FSD (PPL)} & 0.66 & \textbf{$+$0.04} & $-$0.05 & \textbf{$+$0.02} & \textbf{$+$0.06} & \textbf{$+$0.02} & \textbf{$+$0.06} & $-$0.04 & \textbf{$+$0.02} & \textbf{$+$0.04} & \textbf{$+$0.02} \\
\textbf{FSD (Zlib)} & 0.65 & \textbf{$+$0.01} & \textbf{$+$0.02} & \textbf{$+$0.08} & \textbf{$+$0.05} & \textbf{$+$0.04} & \textbf{$+$0.04} & \textbf{$+$0.02} & \textbf{$+$0.09} & \textbf{$+$0.04} & \textbf{$+$0.05} \\
\textbf{FSD (Lower)} & 0.51 & $-$0.03 & \textbf{$+$0.01} & $-$0.05 & $-$0.00 & $-$0.02 & $-$0.00 & $-$0.04 & $-$0.02 & \textbf{$+$0.02} & $-$0.01 \\
\textbf{FSD (MinK)} & 0.66 & $-$0.02 & $-$0.07 & \textbf{$+$0.02} & \textbf{$+$0.06} & $-$0.00 & \textbf{$+$0.03} & $-$0.06 & \textbf{$+$0.02} & \textbf{$+$0.05} & \textbf{$+$0.01} \\
\textbf{FSD (MinK++)} & 0.72 & $-$0.10 & $-$0.06 & $-$0.03 & $-$0.01 & $-$0.05 & $-$0.06 & $-$0.07 & $-$0.02 & $-$0.02 & $-$0.04 \\
\bottomrule
\end{tabular}%
}
\caption{Supervised MIAs on the SFT domain of \textsc{OLMo-Detect} (macro-averaged over its subsets), evaluated via AUC. \textbf{Top:} AUCs of Sup-CAMIA and MIA-Tuner under the \textbf{In-domain} setting (trained on this domain's dev set) and the \textbf{LODO} setting (trained on the other eight domains' dev sets); Zlib, the strongest unsupervised MIA, is repeated as a reference and is excluded from the bold, which marks the best supervised entry per column. \textbf{Bottom:} FSD measures the change in a base MIA's score after fine-tuning the target model on non-members, so we report its AUC gain ($\Delta$AUC) over the base MIA (in parentheses) on this domain. \textbf{Base AUC}: the base MIA's AUC on this domain averaged across model sizes; positive gains are bolded. \textbf{Model-Avg} averages across model sizes.}
\label{table:auc_sup_sft}
\end{table*}

\begin{table*}[t]
\centering
\resizebox{0.7\textwidth}{!}{%
\begin{tabular}{@{}l@{\;\;}l@{\;\;\;}c@{\;\;}c@{\;\;}c@{\;\;}c@{\;\;\;}c@{}}
\toprule
\textbf{Method} & \textbf{Setting} & \textbf{1B} & \textbf{7B} & \textbf{13B} & \textbf{32B} & \textbf{Model-Avg} \\
\midrule
\textit{Zlib (unsup.)} & --- & 0.55 & 0.39 & 0.47 & 0.53 & 0.48 \\
\cmidrule(l){1-7}
Sup-CAMIA & In-domain & \textbf{0.60} & \textbf{0.58} & \textbf{0.58} & \textbf{0.57} & \textbf{0.58} \\
 & LODO & 0.40 & 0.42 & 0.40 & 0.44 & 0.42 \\
\cmidrule(l){1-7}
MIA-Tuner & In-domain & 0.49 & 0.55 & 0.56 & 0.47 & 0.52 \\
 & LODO & 0.41 & 0.44 & 0.42 & 0.48 & 0.44 \\
\bottomrule
\end{tabular}%
}

\vspace{4pt}

\resizebox{0.9\textwidth}{!}{%
\begin{tabular}{@{}lc@{\;\;}c@{\;\;}c@{\;\;}c@{\;\;}c@{\;\;}c@{\;\;\;}c@{\;\;}c@{\;\;}c@{\;\;}c@{\;\;}c@{}}
\toprule
\multirow{2.5}{*}{\textbf{Method}} & \multirow{2.5}{*}{\textbf{Base AUC}} & \multicolumn{5}{c}{\textbf{In-domain $\Delta$AUC}} & \multicolumn{5}{c}{\textbf{LODO $\Delta$AUC}} \\
\cmidrule(lr){3-7} \cmidrule(lr){8-12}
 & & 1B & 7B & 13B & 32B & \textbf{Model-Avg} & 1B & 7B & 13B & 32B & \textbf{Model-Avg} \\
\midrule
\textbf{FSD (PPL)} & 0.42 & $-$0.04 & $-$0.01 & $-$0.04 & $-$0.02 & $-$0.03 & $-$0.03 & $-$0.01 & $-$0.06 & $-$0.01 & $-$0.03 \\
\textbf{FSD (Zlib)} & 0.48 & $-$0.13 & $-$0.02 & $-$0.13 & $-$0.04 & $-$0.08 & $-$0.12 & $-$0.01 & $-$0.15 & $-$0.06 & $-$0.09 \\
\textbf{FSD (Lower)} & 0.60 & $-$0.08 & \textbf{$+$0.02} & $-$0.03 & $-$0.06 & $-$0.04 & $-$0.06 & \textbf{$+$0.05} & $-$0.03 & $-$0.10 & $-$0.04 \\
\textbf{FSD (MinK)} & 0.43 & $-$0.07 & \textbf{$+$0.00} & $-$0.04 & $-$0.02 & $-$0.03 & $-$0.06 & $-$0.00 & $-$0.06 & $-$0.01 & $-$0.03 \\
\textbf{FSD (MinK++)} & 0.44 & $-$0.08 & $-$0.01 & $-$0.03 & \textbf{$+$0.00} & $-$0.03 & $-$0.07 & $-$0.01 & $-$0.02 & \textbf{$+$0.02} & $-$0.02 \\
\bottomrule
\end{tabular}%
}
\caption{Supervised MIAs on the UF-Wildchat subset of \textsc{OLMo-Detect}'s DPO domain, evaluated via AUC. \textbf{Top:} AUCs of Sup-CAMIA and MIA-Tuner under the \textbf{In-domain} setting (trained on this domain's dev set) and the \textbf{LODO} setting (trained on the other eight domains' dev sets); Zlib, the strongest unsupervised MIA, is repeated as a reference and is excluded from the bold, which marks the best supervised entry per column. \textbf{Bottom:} FSD measures the change in a base MIA's score after fine-tuning the target model on non-members, so we report its AUC gain ($\Delta$AUC) over the base MIA (in parentheses) on this domain. \textbf{Base AUC}: the base MIA's AUC on this domain averaged across model sizes; positive gains are bolded. \textbf{Model-Avg} averages across model sizes.}
\label{table:auc_sup_dpo}
\end{table*}

\begin{table*}[t]
\centering
\resizebox{0.7\textwidth}{!}{%
\begin{tabular}{@{}l@{\;\;}l@{\;\;\;}c@{\;\;}c@{\;\;}c@{\;\;}c@{\;\;\;}c@{}}
\toprule
\textbf{Method} & \textbf{Setting} & \textbf{1B} & \textbf{7B} & \textbf{13B} & \textbf{32B} & \textbf{Model-Avg} \\
\midrule
\textit{Zlib (unsup.)} & --- & 0.79 & 0.88 & 0.94 & 0.91 & 0.88 \\
\cmidrule(l){1-7}
Sup-CAMIA & In-domain & \textbf{0.73} & \textbf{0.86} & \textbf{0.91} & 0.90 & \textbf{0.85} \\
 & LODO & 0.62 & 0.83 & 0.68 & \textbf{0.94} & 0.77 \\
\cmidrule(l){1-7}
MIA-Tuner & In-domain & 0.56 & 0.54 & 0.53 & 0.54 & 0.54 \\
 & LODO & 0.51 & 0.48 & 0.48 & 0.52 & 0.50 \\
\bottomrule
\end{tabular}%
}

\vspace{4pt}

\resizebox{0.9\textwidth}{!}{%
\begin{tabular}{@{}lc@{\;\;}c@{\;\;}c@{\;\;}c@{\;\;}c@{\;\;}c@{\;\;\;}c@{\;\;}c@{\;\;}c@{\;\;}c@{\;\;}c@{}}
\toprule
\multirow{2.5}{*}{\textbf{Method}} & \multirow{2.5}{*}{\textbf{Base AUC}} & \multicolumn{5}{c}{\textbf{In-domain $\Delta$AUC}} & \multicolumn{5}{c}{\textbf{LODO $\Delta$AUC}} \\
\cmidrule(lr){3-7} \cmidrule(lr){8-12}
 & & 1B & 7B & 13B & 32B & \textbf{Model-Avg} & 1B & 7B & 13B & 32B & \textbf{Model-Avg} \\
\midrule
\textbf{FSD (PPL)} & 0.89 & $-$0.05 & $-$0.21 & $-$0.12 & $-$0.31 & $-$0.17 & $-$0.02 & $-$0.02 & $-$0.03 & $-$0.07 & $-$0.03 \\
\textbf{FSD (Zlib)} & 0.88 & $-$0.11 & $-$0.25 & $-$0.14 & $-$0.31 & $-$0.20 & $-$0.05 & $-$0.02 & $-$0.02 & $-$0.05 & $-$0.04 \\
\textbf{FSD (Lower)} & 0.79 & $-$0.04 & $-$0.11 & $-$0.05 & $-$0.14 & $-$0.08 & $-$0.06 & $-$0.07 & $-$0.01 & $-$0.01 & $-$0.03 \\
\textbf{FSD (MinK)} & 0.90 & $-$0.19 & $-$0.59 & $-$0.48 & $-$0.59 & $-$0.46 & $-$0.10 & $-$0.07 & $-$0.04 & $-$0.11 & $-$0.08 \\
\textbf{FSD (MinK++)} & 0.66 & $-$0.02 & $-$0.07 & $-$0.02 & $-$0.14 & $-$0.06 & $-$0.01 & $-$0.02 & $-$0.00 & $-$0.06 & $-$0.02 \\
\bottomrule
\end{tabular}%
}
\caption{Supervised MIAs on the RLVR domain of \textsc{OLMo-Detect} (macro-averaged over its subsets), evaluated via AUC. \textbf{Top:} AUCs of Sup-CAMIA and MIA-Tuner under the \textbf{In-domain} setting (trained on this domain's dev set) and the \textbf{LODO} setting (trained on the other eight domains' dev sets); Zlib, the strongest unsupervised MIA, is repeated as a reference and is excluded from the bold, which marks the best supervised entry per column. \textbf{Bottom:} FSD measures the change in a base MIA's score after fine-tuning the target model on non-members, so we report its AUC gain ($\Delta$AUC) over the base MIA (in parentheses) on this domain. \textbf{Base AUC}: the base MIA's AUC on this domain averaged across model sizes; positive gains are bolded. \textbf{Model-Avg} averages across model sizes.}
\label{table:auc_sup_rlvr}
\end{table*}

\section{TPR@5\%FPR Results for Supervised MIAs on \textsc{OLMo-Detect}}
\label{appendix:tpr_results_sup}

The main text reports supervised MIAs by AUC only, so we give their TPR@5\% FPR here. Table~\ref{table:tpr_sup:all} is the overall table, aggregated by training stage, and Tables~\ref{table:tpr_sup_dclm}--\ref{table:tpr_sup_rlvr} give one table per domain. All follow the two-part layout of Table~\ref{table:olmo_detect_overall_supervised}: the top part reports Sup-CAMIA and MIA-Tuner under both settings, with Zlib repeated as an unsupervised reference, and the bottom part reports FSD's gain over the base MIA it is built on. \textbf{In-domain} trains on the evaluated domain's development set, and \textbf{LODO} trains on the other eight domains' development sets.

TPR@5\% FPR reproduces the AUC picture and sharpens it. Overall, the best supervised configuration, Sup-CAMIA under In-domain, reaches $0.26$ against Zlib's $0.28$, so supervision does not improve the operating point that matters for an attacker; moving to LODO halves it to $0.14$, and MIA-Tuner reaches only $0.09$ In-domain and $0.05$ under LODO, the latter indistinguishable from the $0.05$ of a random detector. Per domain, the better of Sup-CAMIA and MIA-Tuner exceeds Zlib on four of the nine domains and falls below it on the other five, with the largest shortfall on StarCoder ($-0.10$) and the largest gains on Math Mix and DPO ($+0.08$ each). FSD again rarely helps its base MIA: of the $45$ domain-by-base-MIA combinations it improves in $12$ under In-domain and $11$ under LODO.

\begin{table*}[t]
\centering
\resizebox{\textwidth}{!}{%
\begin{tabular}{@{}ll@{\;\;}c@{\;\;}c@{\;\;}c@{\;\;}c@{\;\;}c@{\;\;}c@{\;\;}c@{\;\;}c@{\;\;}c@{\;\;}c@{\;\;}c@{\;\;}c@{\;\;}c@{\;\;}c@{\;\;}c@{\;\;}c@{\;\;}c@{\;\;}c@{\;\;}c@{\;\;}c@{}}
\toprule
\multirow{2.5}{*}{\textbf{Method}} & \multirow{2.5}{*}{\textbf{Setting}} & \multicolumn{4}{c}{\textbf{1B}} & \multicolumn{4}{c}{\textbf{7B}} & \multicolumn{4}{c}{\textbf{13B}} & \multicolumn{4}{c}{\textbf{32B}} & \multicolumn{4}{c}{\textbf{Model-Avg}} \\
\cmidrule(lr){3-6} \cmidrule(lr){7-10} \cmidrule(lr){11-14} \cmidrule(lr){15-18} \cmidrule(lr){19-22}
 & & Pre & Mid & Post & All & Pre & Mid & Post & All & Pre & Mid & Post & All & Pre & Mid & Post & All & Pre & Mid & Post & All \\
\midrule
\textit{Zlib (unsup.)} & --- & 0.09 & 0.17 & 0.18 & 0.15 & 0.13 & 0.33 & 0.32 & 0.26 & 0.13 & 0.57 & 0.34 & 0.35 & 0.16 & 0.58 & 0.31 & 0.35 & 0.13 & 0.41 & 0.29 & 0.28 \\
\midrule
\multirow{2}{*}{\textbf{Sup-CAMIA}} & In-domain & \textbf{0.08} & \textbf{0.24} & \textbf{0.19} & \textbf{0.17} & 0.07 & \textbf{0.38} & \textbf{0.30} & \textbf{0.25} & \textbf{0.08} & \textbf{0.57} & \textbf{0.36} & \textbf{0.34} & 0.06 & \textbf{0.56} & 0.27 & \textbf{0.30} & 0.07 & \textbf{0.44} & \textbf{0.28} & \textbf{0.26} \\
 & LODO & 0.07 & 0.09 & 0.08 & 0.08 & 0.06 & 0.22 & 0.25 & 0.18 & 0.07 & 0.07 & 0.11 & 0.08 & \textbf{0.10} & 0.24 & \textbf{0.28} & 0.21 & \textbf{0.07} & 0.16 & 0.18 & 0.14 \\
\midrule
\multirow{2}{*}{\textbf{MIA-Tuner}} & In-domain & 0.07 & 0.13 & 0.10 & 0.10 & \textbf{0.08} & 0.20 & 0.09 & 0.12 & 0.04 & 0.15 & 0.08 & 0.09 & 0.06 & 0.07 & 0.06 & 0.06 & 0.06 & 0.14 & 0.08 & 0.09 \\
 & LODO & 0.07 & 0.05 & 0.05 & 0.06 & 0.04 & 0.07 & 0.05 & 0.05 & 0.04 & 0.05 & 0.06 & 0.05 & 0.05 & 0.06 & 0.06 & 0.06 & 0.05 & 0.06 & 0.05 & 0.05 \\
\bottomrule
\end{tabular}%
}

\vspace{4pt}

\resizebox{0.9\textwidth}{!}{%
\begin{tabular}{@{}lc@{\;\;}c@{\;\;}c@{\;\;}c@{\;\;}c@{\;\;}c@{\;\;\;}c@{\;\;}c@{\;\;}c@{\;\;}c@{\;\;}c@{}}
\toprule
\multirow{2.5}{*}{\textbf{Method}} & \multirow{2.5}{*}{\textbf{Base TPR@5\% FPR}} & \multicolumn{5}{c}{\textbf{In-domain $\Delta$}} & \multicolumn{5}{c}{\textbf{LODO $\Delta$}} \\
\cmidrule(lr){3-7} \cmidrule(lr){8-12}
 & & 1B & 7B & 13B & 32B & \textbf{Model-Avg} & 1B & 7B & 13B & 32B & \textbf{Model-Avg} \\
\midrule
\textbf{FSD (PPL)} & 0.25 & $-$0.00 & $-$0.06 & \textbf{$+$0.01} & $-$0.10 & $-$0.04 & $-$0.02 & $-$0.05 & $-$0.02 & $-$0.13 & $-$0.06 \\
\textbf{FSD (Zlib)} & 0.28 & $-$0.03 & $-$0.10 & $-$0.06 & $-$0.12 & $-$0.08 & $-$0.04 & $-$0.07 & $-$0.04 & $-$0.14 & $-$0.07 \\
\textbf{FSD (Lower)} & 0.18 & $-$0.02 & $-$0.02 & $-$0.00 & $-$0.05 & $-$0.02 & $-$0.04 & $-$0.02 & $-$0.03 & $-$0.05 & $-$0.04 \\
\textbf{FSD (MinK)} & 0.26 & $-$0.05 & $-$0.15 & $-$0.11 & $-$0.15 & $-$0.11 & $-$0.05 & $-$0.08 & $-$0.03 & $-$0.14 & $-$0.07 \\
\textbf{FSD (MinK++)} & 0.12 & $-$0.04 & $-$0.04 & $-$0.01 & $-$0.06 & $-$0.04 & $-$0.06 & $-$0.04 & $-$0.01 & $-$0.04 & $-$0.04 \\
\bottomrule
\end{tabular}%
}
\caption{Supervised MIAs on \textsc{OLMo-Detect}, evaluated via TPR@5\% FPR. \textbf{Top:} scores of Sup-CAMIA and MIA-Tuner under the \textbf{In-domain} setting (trained on the evaluated domain's dev set) and the \textbf{LODO} setting (trained on the other eight domains' dev sets); Zlib, the strongest unsupervised MIA, is repeated as a reference and is excluded from the bold, which marks the best supervised entry per column. \textbf{Pre}/\textbf{Mid}/\textbf{Post} macro-average over each stage's domains and \textbf{All} macro-averages the three stages, as in Table~\ref{table:olmo_detect_overall_unsupervised}. \textbf{Bottom:} FSD's gain ($\Delta$) over its base MIA (in parentheses) under the \textbf{All} aggregation; \textbf{Base TPR@5\% FPR} is the base MIA's own score averaged across model sizes, and positive gains are bolded. \textbf{Model-Avg} averages across model sizes.}
\label{table:tpr_sup:all}
\end{table*}

\begin{table*}[t]
\centering
\resizebox{0.7\textwidth}{!}{%
\begin{tabular}{@{}l@{\;\;}l@{\;\;\;}c@{\;\;}c@{\;\;}c@{\;\;}c@{\;\;\;}c@{}}
\toprule
\textbf{Method} & \textbf{Setting} & \textbf{1B} & \textbf{7B} & \textbf{13B} & \textbf{32B} & \textbf{Model-Avg} \\
\midrule
\textit{Zlib (unsup.)} & --- & 0.06 & 0.11 & 0.13 & 0.18 & 0.12 \\
\cmidrule(l){1-7}
Sup-CAMIA & In-domain & 0.01 & 0.05 & \textbf{0.04} & 0.06 & 0.04 \\
 & LODO & \textbf{0.05} & \textbf{0.08} & 0.04 & \textbf{0.17} & \textbf{0.09} \\
\cmidrule(l){1-7}
MIA-Tuner & In-domain & 0.01 & 0.01 & 0.00 & 0.06 & 0.02 \\
 & LODO & 0.02 & 0.02 & 0.03 & 0.01 & 0.02 \\
\bottomrule
\end{tabular}%
}

\vspace{4pt}

\resizebox{0.9\textwidth}{!}{%
\begin{tabular}{@{}lc@{\;\;}c@{\;\;}c@{\;\;}c@{\;\;}c@{\;\;}c@{\;\;\;}c@{\;\;}c@{\;\;}c@{\;\;}c@{\;\;}c@{}}
\toprule
\multirow{2.5}{*}{\textbf{Method}} & \multirow{2.5}{*}{\textbf{Base TPR@5\% FPR}} & \multicolumn{5}{c}{\textbf{In-domain $\Delta$}} & \multicolumn{5}{c}{\textbf{LODO $\Delta$}} \\
\cmidrule(lr){3-7} \cmidrule(lr){8-12}
 & & 1B & 7B & 13B & 32B & \textbf{Model-Avg} & 1B & 7B & 13B & 32B & \textbf{Model-Avg} \\
\midrule
\textbf{FSD (PPL)} & 0.15 & $-$0.05 & $-$0.10 & $-$0.08 & $-$0.13 & $-$0.09 & $-$0.06 & $-$0.01 & $-$0.09 & $-$0.17 & $-$0.08 \\
\textbf{FSD (Zlib)} & 0.12 & $-$0.01 & $-$0.07 & $-$0.10 & $-$0.07 & $-$0.06 & $-$0.02 & \textbf{$+$0.03} & $-$0.04 & $-$0.11 & $-$0.04 \\
\textbf{FSD (Lower)} & 0.14 & $-$0.07 & $-$0.04 & $-$0.02 & $-$0.03 & $-$0.04 & $-$0.07 & $-$0.06 & \textbf{$+$0.00} & $-$0.04 & $-$0.04 \\
\textbf{FSD (MinK)} & 0.19 & $-$0.11 & $-$0.19 & $-$0.15 & $-$0.11 & $-$0.14 & $-$0.09 & $-$0.11 & $-$0.14 & $-$0.18 & $-$0.13 \\
\textbf{FSD (MinK++)} & 0.06 & \textbf{$+$0.01} & $-$0.01 & \textbf{$+$0.01} & $-$0.06 & $-$0.01 & \textbf{$+$0.04} & $-$0.04 & $-$0.01 & $-$0.09 & $-$0.03 \\
\bottomrule
\end{tabular}%
}
\caption{Supervised MIAs on the DCLM-Baseline domain of \textsc{OLMo-Detect}, evaluated via TPR@5\% FPR. \textbf{Top:} scores of Sup-CAMIA and MIA-Tuner under the \textbf{In-domain} setting (trained on this domain's dev set) and the \textbf{LODO} setting (trained on the other eight domains' dev sets); Zlib, the strongest unsupervised MIA, is repeated as a reference and is excluded from the bold, which marks the best supervised entry per column. \textbf{Bottom:} FSD measures the change in a base MIA's score after fine-tuning the target model on non-members, so we report its gain ($\Delta$) over the base MIA (in parentheses) on this domain. \textbf{Base TPR@5\% FPR}: the base MIA's own score on this domain averaged across model sizes; positive gains are bolded. \textbf{Model-Avg} averages across model sizes.}
\label{table:tpr_sup_dclm}
\end{table*}

\begin{table*}[t]
\centering
\resizebox{0.7\textwidth}{!}{%
\begin{tabular}{@{}l@{\;\;}l@{\;\;\;}c@{\;\;}c@{\;\;}c@{\;\;}c@{\;\;\;}c@{}}
\toprule
\textbf{Method} & \textbf{Setting} & \textbf{1B} & \textbf{7B} & \textbf{13B} & \textbf{32B} & \textbf{Model-Avg} \\
\midrule
\textit{Zlib (unsup.)} & --- & 0.07 & 0.07 & 0.07 & 0.09 & 0.07 \\
\cmidrule(l){1-7}
Sup-CAMIA & In-domain & 0.08 & 0.04 & 0.05 & 0.08 & 0.06 \\
 & LODO & 0.05 & 0.03 & \textbf{0.06} & 0.05 & 0.05 \\
\cmidrule(l){1-7}
MIA-Tuner & In-domain & \textbf{0.09} & 0.08 & 0.06 & 0.09 & 0.08 \\
 & LODO & 0.07 & \textbf{0.12} & 0.01 & \textbf{0.13} & \textbf{0.08} \\
\bottomrule
\end{tabular}%
}

\vspace{4pt}

\resizebox{0.9\textwidth}{!}{%
\begin{tabular}{@{}lc@{\;\;}c@{\;\;}c@{\;\;}c@{\;\;}c@{\;\;}c@{\;\;\;}c@{\;\;}c@{\;\;}c@{\;\;}c@{\;\;}c@{}}
\toprule
\multirow{2.5}{*}{\textbf{Method}} & \multirow{2.5}{*}{\textbf{Base TPR@5\% FPR}} & \multicolumn{5}{c}{\textbf{In-domain $\Delta$}} & \multicolumn{5}{c}{\textbf{LODO $\Delta$}} \\
\cmidrule(lr){3-7} \cmidrule(lr){8-12}
 & & 1B & 7B & 13B & 32B & \textbf{Model-Avg} & 1B & 7B & 13B & 32B & \textbf{Model-Avg} \\
\midrule
\textbf{FSD (PPL)} & 0.07 & $-$0.02 & $-$0.02 & $-$0.03 & $-$0.04 & $-$0.03 & $-$0.01 & $-$0.01 & \textbf{$+$0.05} & $-$0.01 & \textbf{$+$0.01} \\
\textbf{FSD (Zlib)} & 0.07 & $-$0.02 & $+$0.00 & $-$0.01 & $-$0.05 & $-$0.02 & $-$0.02 & $-$0.02 & \textbf{$+$0.03} & $-$0.04 & $-$0.01 \\
\textbf{FSD (Lower)} & 0.07 & \textbf{$+$0.00} & $-$0.03 & $-$0.03 & $-$0.01 & $-$0.02 & \textbf{$+$0.01} & $-$0.03 & $-$0.03 & $-$0.04 & $-$0.02 \\
\textbf{FSD (MinK)} & 0.08 & $-$0.03 & $-$0.03 & $-$0.04 & $-$0.05 & $-$0.04 & $-$0.02 & $-$0.03 & \textbf{$+$0.00} & $-$0.03 & $-$0.02 \\
\textbf{FSD (MinK++)} & 0.07 & $-$0.04 & $-$0.00 & $-$0.03 & $-$0.01 & $-$0.02 & $-$0.02 & $-$0.00 & \textbf{$+$0.01} & \textbf{$+$0.00} & $-$0.00 \\
\bottomrule
\end{tabular}%
}
\caption{Supervised MIAs on the peS2o domain of \textsc{OLMo-Detect}, evaluated via TPR@5\% FPR. \textbf{Top:} scores of Sup-CAMIA and MIA-Tuner under the \textbf{In-domain} setting (trained on this domain's dev set) and the \textbf{LODO} setting (trained on the other eight domains' dev sets); Zlib, the strongest unsupervised MIA, is repeated as a reference and is excluded from the bold, which marks the best supervised entry per column. \textbf{Bottom:} FSD measures the change in a base MIA's score after fine-tuning the target model on non-members, so we report its gain ($\Delta$) over the base MIA (in parentheses) on this domain. \textbf{Base TPR@5\% FPR}: the base MIA's own score on this domain averaged across model sizes; positive gains are bolded. \textbf{Model-Avg} averages across model sizes.}
\label{table:tpr_sup_pes2o}
\end{table*}
\clearpage

\begin{table*}[t]
\centering
\resizebox{0.7\textwidth}{!}{%
\begin{tabular}{@{}l@{\;\;}l@{\;\;\;}c@{\;\;}c@{\;\;}c@{\;\;}c@{\;\;\;}c@{}}
\toprule
\textbf{Method} & \textbf{Setting} & \textbf{1B} & \textbf{7B} & \textbf{13B} & \textbf{32B} & \textbf{Model-Avg} \\
\midrule
\textit{Zlib (unsup.)} & --- & 0.07 & 0.07 & 0.10 & 0.17 & 0.10 \\
\cmidrule(l){1-7}
Sup-CAMIA & In-domain & 0.10 & 0.06 & \textbf{0.10} & 0.04 & \textbf{0.07} \\
 & LODO & 0.05 & 0.09 & 0.03 & \textbf{0.09} & 0.06 \\
\cmidrule(l){1-7}
MIA-Tuner & In-domain & 0.05 & \textbf{0.09} & 0.06 & 0.05 & 0.06 \\
 & LODO & \textbf{0.11} & 0.01 & 0.06 & 0.07 & 0.06 \\
\bottomrule
\end{tabular}%
}

\vspace{4pt}

\resizebox{0.9\textwidth}{!}{%
\begin{tabular}{@{}lc@{\;\;}c@{\;\;}c@{\;\;}c@{\;\;}c@{\;\;}c@{\;\;\;}c@{\;\;}c@{\;\;}c@{\;\;}c@{\;\;}c@{}}
\toprule
\multirow{2.5}{*}{\textbf{Method}} & \multirow{2.5}{*}{\textbf{Base TPR@5\% FPR}} & \multicolumn{5}{c}{\textbf{In-domain $\Delta$}} & \multicolumn{5}{c}{\textbf{LODO $\Delta$}} \\
\cmidrule(lr){3-7} \cmidrule(lr){8-12}
 & & 1B & 7B & 13B & 32B & \textbf{Model-Avg} & 1B & 7B & 13B & 32B & \textbf{Model-Avg} \\
\midrule
\textbf{FSD (PPL)} & 0.06 & \textbf{$+$0.04} & \textbf{$+$0.06} & \textbf{$+$0.04} & \textbf{$+$0.06} & \textbf{$+$0.05} & \textbf{$+$0.01} & $+$0.00 & $-$0.02 & $-$0.05 & $-$0.02 \\
\textbf{FSD (Zlib)} & 0.10 & \textbf{$+$0.02} & \textbf{$+$0.04} & \textbf{$+$0.02} & $-$0.01 & \textbf{$+$0.02} & $-$0.01 & $-$0.04 & $-$0.05 & $-$0.07 & $-$0.04 \\
\textbf{FSD (Lower)} & 0.06 & $-$0.00 & $-$0.05 & \textbf{$+$0.01} & $-$0.01 & $-$0.01 & $-$0.03 & $-$0.03 & $-$0.02 & $-$0.00 & $-$0.02 \\
\textbf{FSD (MinK)} & 0.06 & \textbf{$+$0.04} & \textbf{$+$0.05} & \textbf{$+$0.03} & \textbf{$+$0.11} & \textbf{$+$0.05} & \textbf{$+$0.01} & $-$0.00 & $-$0.02 & $-$0.03 & $-$0.01 \\
\textbf{FSD (MinK++)} & 0.12 & $-$0.03 & $-$0.10 & $-$0.07 & $-$0.09 & $-$0.07 & \textbf{$+$0.01} & $-$0.09 & $-$0.08 & $-$0.06 & $-$0.06 \\
\bottomrule
\end{tabular}%
}
\caption{Supervised MIAs on the OpenWebMath domain of \textsc{OLMo-Detect}, evaluated via TPR@5\% FPR. \textbf{Top:} scores of Sup-CAMIA and MIA-Tuner under the \textbf{In-domain} setting (trained on this domain's dev set) and the \textbf{LODO} setting (trained on the other eight domains' dev sets); Zlib, the strongest unsupervised MIA, is repeated as a reference and is excluded from the bold, which marks the best supervised entry per column. \textbf{Bottom:} FSD measures the change in a base MIA's score after fine-tuning the target model on non-members, so we report its gain ($\Delta$) over the base MIA (in parentheses) on this domain. \textbf{Base TPR@5\% FPR}: the base MIA's own score on this domain averaged across model sizes; positive gains are bolded. \textbf{Model-Avg} averages across model sizes.}
\label{table:tpr_sup_openwebmath}
\end{table*}

\begin{table*}[t]
\centering
\resizebox{0.7\textwidth}{!}{%
\begin{tabular}{@{}l@{\;\;}l@{\;\;\;}c@{\;\;}c@{\;\;}c@{\;\;}c@{\;\;\;}c@{}}
\toprule
\textbf{Method} & \textbf{Setting} & \textbf{1B} & \textbf{7B} & \textbf{13B} & \textbf{32B} & \textbf{Model-Avg} \\
\midrule
\textit{Zlib (unsup.)} & --- & 0.17 & 0.27 & 0.24 & 0.19 & 0.22 \\
\cmidrule(l){1-7}
Sup-CAMIA & In-domain & 0.12 & \textbf{0.13} & 0.13 & 0.08 & \textbf{0.12} \\
 & LODO & 0.12 & 0.04 & \textbf{0.14} & \textbf{0.10} & 0.10 \\
\cmidrule(l){1-7}
MIA-Tuner & In-domain & \textbf{0.12} & 0.11 & 0.02 & 0.05 & 0.08 \\
 & LODO & 0.09 & 0.02 & 0.05 & 0.01 & 0.04 \\
\bottomrule
\end{tabular}%
}

\vspace{4pt}

\resizebox{0.9\textwidth}{!}{%
\begin{tabular}{@{}lc@{\;\;}c@{\;\;}c@{\;\;}c@{\;\;}c@{\;\;}c@{\;\;\;}c@{\;\;}c@{\;\;}c@{\;\;}c@{\;\;}c@{}}
\toprule
\multirow{2.5}{*}{\textbf{Method}} & \multirow{2.5}{*}{\textbf{Base TPR@5\% FPR}} & \multicolumn{5}{c}{\textbf{In-domain $\Delta$}} & \multicolumn{5}{c}{\textbf{LODO $\Delta$}} \\
\cmidrule(lr){3-7} \cmidrule(lr){8-12}
 & & 1B & 7B & 13B & 32B & \textbf{Model-Avg} & 1B & 7B & 13B & 32B & \textbf{Model-Avg} \\
\midrule
\textbf{FSD (PPL)} & 0.08 & $-$0.03 & $-$0.01 & $-$0.01 & $-$0.02 & $-$0.02 & \textbf{$+$0.06} & $-$0.01 & $-$0.03 & $-$0.02 & \textbf{$+$0.00} \\
\textbf{FSD (Zlib)} & 0.22 & $-$0.05 & $-$0.17 & $-$0.11 & $-$0.05 & $-$0.09 & $-$0.03 & $-$0.18 & $-$0.15 & $-$0.19 & $-$0.14 \\
\textbf{FSD (Lower)} & 0.07 & \textbf{$+$0.06} & \textbf{$+$0.02} & $-$0.05 & \textbf{$+$0.06} & \textbf{$+$0.02} & \textbf{$+$0.03} & \textbf{$+$0.03} & $-$0.05 & $-$0.02 & \textbf{$+$0.00} \\
\textbf{FSD (MinK)} & 0.08 & $-$0.06 & $-$0.05 & $-$0.03 & $-$0.03 & $-$0.04 & \textbf{$+$0.07} & \textbf{$+$0.02} & $-$0.03 & $-$0.02 & \textbf{$+$0.01} \\
\textbf{FSD (MinK++)} & 0.08 & $-$0.11 & $-$0.02 & \textbf{$+$0.00} & $-$0.02 & $-$0.04 & $-$0.04 & \textbf{$+$0.03} & $-$0.01 & \textbf{$+$0.02} & $-$0.00 \\
\bottomrule
\end{tabular}%
}
\caption{Supervised MIAs on the StarCoder domain of \textsc{OLMo-Detect} (macro-averaged over its subsets), evaluated via TPR@5\% FPR. \textbf{Top:} scores of Sup-CAMIA and MIA-Tuner under the \textbf{In-domain} setting (trained on this domain's dev set) and the \textbf{LODO} setting (trained on the other eight domains' dev sets); Zlib, the strongest unsupervised MIA, is repeated as a reference and is excluded from the bold, which marks the best supervised entry per column. \textbf{Bottom:} FSD measures the change in a base MIA's score after fine-tuning the target model on non-members, so we report its gain ($\Delta$) over the base MIA (in parentheses) on this domain. \textbf{Base TPR@5\% FPR}: the base MIA's own score on this domain averaged across model sizes; positive gains are bolded. \textbf{Model-Avg} averages across model sizes.}
\label{table:tpr_sup_starcoder}
\end{table*}

\begin{table*}[t]
\centering
\resizebox{0.7\textwidth}{!}{%
\begin{tabular}{@{}l@{\;\;}l@{\;\;\;}c@{\;\;}c@{\;\;}c@{\;\;}c@{\;\;\;}c@{}}
\toprule
\textbf{Method} & \textbf{Setting} & \textbf{1B} & \textbf{7B} & \textbf{13B} & \textbf{32B} & \textbf{Model-Avg} \\
\midrule
\textit{Zlib (unsup.)} & --- & 0.25 & 0.54 & 0.97 & 0.96 & 0.68 \\
\cmidrule(l){1-7}
Sup-CAMIA & In-domain & \textbf{0.39} & \textbf{0.69} & \textbf{0.97} & \textbf{0.97} & \textbf{0.76} \\
 & LODO & 0.13 & 0.38 & 0.07 & 0.43 & 0.25 \\
\cmidrule(l){1-7}
MIA-Tuner & In-domain & 0.06 & 0.05 & 0.06 & 0.06 & 0.06 \\
 & LODO & 0.05 & 0.07 & 0.05 & 0.06 & 0.06 \\
\bottomrule
\end{tabular}%
}

\vspace{4pt}

\resizebox{0.9\textwidth}{!}{%
\begin{tabular}{@{}lc@{\;\;}c@{\;\;}c@{\;\;}c@{\;\;}c@{\;\;}c@{\;\;\;}c@{\;\;}c@{\;\;}c@{\;\;}c@{\;\;}c@{}}
\toprule
\multirow{2.5}{*}{\textbf{Method}} & \multirow{2.5}{*}{\textbf{Base TPR@5\% FPR}} & \multicolumn{5}{c}{\textbf{In-domain $\Delta$}} & \multicolumn{5}{c}{\textbf{LODO $\Delta$}} \\
\cmidrule(lr){3-7} \cmidrule(lr){8-12}
 & & 1B & 7B & 13B & 32B & \textbf{Model-Avg} & 1B & 7B & 13B & 32B & \textbf{Model-Avg} \\
\midrule
\textbf{FSD (PPL)} & 0.61 & $-$0.02 & $-$0.06 & \textbf{$+$0.05} & $-$0.16 & $-$0.05 & $-$0.17 & $-$0.22 & $-$0.05 & $-$0.53 & $-$0.24 \\
\textbf{FSD (Zlib)} & 0.68 & $-$0.08 & $-$0.15 & \textbf{$+$0.01} & $-$0.21 & $-$0.11 & $-$0.20 & $-$0.29 & $-$0.03 & $-$0.49 & $-$0.25 \\
\textbf{FSD (Lower)} & 0.38 & $-$0.01 & \textbf{$+$0.07} & \textbf{$+$0.16} & \textbf{$+$0.05} & \textbf{$+$0.06} & $-$0.12 & \textbf{$+$0.01} & $-$0.09 & $-$0.18 & $-$0.09 \\
\textbf{FSD (MinK)} & 0.63 & $-$0.11 & $-$0.39 & $-$0.21 & $-$0.32 & $-$0.26 & $-$0.20 & $-$0.25 & $-$0.09 & $-$0.41 & $-$0.24 \\
\textbf{FSD (MinK++)} & 0.16 & $-$0.04 & $-$0.03 & $-$0.04 & $-$0.09 & $-$0.05 & $-$0.10 & $-$0.07 & $-$0.03 & $-$0.07 & $-$0.07 \\
\bottomrule
\end{tabular}%
}
\caption{Supervised MIAs on the GSM8K subset of \textsc{OLMo-Detect}'s Math Mix domain, evaluated via TPR@5\% FPR. \textbf{Top:} scores of Sup-CAMIA and MIA-Tuner under the \textbf{In-domain} setting (trained on this domain's dev set) and the \textbf{LODO} setting (trained on the other eight domains' dev sets); Zlib, the strongest unsupervised MIA, is repeated as a reference and is excluded from the bold, which marks the best supervised entry per column. \textbf{Bottom:} FSD measures the change in a base MIA's score after fine-tuning the target model on non-members, so we report its gain ($\Delta$) over the base MIA (in parentheses) on this domain. \textbf{Base TPR@5\% FPR}: the base MIA's own score on this domain averaged across model sizes; positive gains are bolded. \textbf{Model-Avg} averages across model sizes.}
\label{table:tpr_sup_gsm8k}
\end{table*}
\clearpage

\begin{table*}[t]
\centering
\resizebox{0.7\textwidth}{!}{%
\begin{tabular}{@{}l@{\;\;}l@{\;\;\;}c@{\;\;}c@{\;\;}c@{\;\;}c@{\;\;\;}c@{}}
\toprule
\textbf{Method} & \textbf{Setting} & \textbf{1B} & \textbf{7B} & \textbf{13B} & \textbf{32B} & \textbf{Model-Avg} \\
\midrule
\textit{Zlib (unsup.)} & --- & 0.08 & 0.12 & 0.17 & 0.20 & 0.14 \\
\cmidrule(l){1-7}
Sup-CAMIA & In-domain & 0.08 & 0.06 & 0.17 & \textbf{0.16} & 0.12 \\
 & LODO & 0.05 & 0.06 & 0.08 & 0.06 & 0.06 \\
\cmidrule(l){1-7}
MIA-Tuner & In-domain & \textbf{0.19} & \textbf{0.36} & \textbf{0.24} & 0.09 & \textbf{0.22} \\
 & LODO & 0.06 & 0.08 & 0.04 & 0.05 & 0.06 \\
\bottomrule
\end{tabular}%
}

\vspace{4pt}

\resizebox{0.9\textwidth}{!}{%
\begin{tabular}{@{}lc@{\;\;}c@{\;\;}c@{\;\;}c@{\;\;}c@{\;\;}c@{\;\;\;}c@{\;\;}c@{\;\;}c@{\;\;}c@{\;\;}c@{}}
\toprule
\multirow{2.5}{*}{\textbf{Method}} & \multirow{2.5}{*}{\textbf{Base TPR@5\% FPR}} & \multicolumn{5}{c}{\textbf{In-domain $\Delta$}} & \multicolumn{5}{c}{\textbf{LODO $\Delta$}} \\
\cmidrule(lr){3-7} \cmidrule(lr){8-12}
 & & 1B & 7B & 13B & 32B & \textbf{Model-Avg} & 1B & 7B & 13B & 32B & \textbf{Model-Avg} \\
\midrule
\textbf{FSD (PPL)} & 0.07 & \textbf{$+$0.02} & \textbf{$+$0.08} & \textbf{$+$0.15} & \textbf{$+$0.02} & \textbf{$+$0.07} & $-$0.03 & $-$0.01 & \textbf{$+$0.02} & $-$0.04 & $-$0.02 \\
\textbf{FSD (Zlib)} & 0.14 & $-$0.00 & \textbf{$+$0.02} & $-$0.03 & $-$0.08 & $-$0.02 & $-$0.05 & $-$0.08 & $-$0.04 & $-$0.15 & $-$0.08 \\
\textbf{FSD (Lower)} & 0.13 & $-$0.07 & $-$0.03 & $-$0.00 & $-$0.12 & $-$0.06 & $-$0.06 & $-$0.00 & $-$0.06 & $-$0.11 & $-$0.06 \\
\textbf{FSD (MinK)} & 0.08 & \textbf{$+$0.00} & $-$0.01 & \textbf{$+$0.08} & $-$0.02 & \textbf{$+$0.01} & $-$0.02 & $-$0.01 & \textbf{$+$0.02} & $-$0.07 & $-$0.02 \\
\textbf{FSD (MinK++)} & 0.07 & $-$0.02 & $-$0.00 & $-$0.00 & $-$0.04 & $-$0.01 & $-$0.02 & \textbf{$+$0.03} & \textbf{$+$0.04} & $-$0.04 & \textbf{$+$0.00} \\
\bottomrule
\end{tabular}%
}
\caption{Supervised MIAs on the Stack Exchange domain of \textsc{OLMo-Detect} (macro-averaged over its subsets), evaluated via TPR@5\% FPR. \textbf{Top:} scores of Sup-CAMIA and MIA-Tuner under the \textbf{In-domain} setting (trained on this domain's dev set) and the \textbf{LODO} setting (trained on the other eight domains' dev sets); Zlib, the strongest unsupervised MIA, is repeated as a reference and is excluded from the bold, which marks the best supervised entry per column. \textbf{Bottom:} FSD measures the change in a base MIA's score after fine-tuning the target model on non-members, so we report its gain ($\Delta$) over the base MIA (in parentheses) on this domain. \textbf{Base TPR@5\% FPR}: the base MIA's own score on this domain averaged across model sizes; positive gains are bolded. \textbf{Model-Avg} averages across model sizes.}
\label{table:tpr_sup_stackexchange}
\end{table*}

\begin{table*}[t]
\centering
\resizebox{0.7\textwidth}{!}{%
\begin{tabular}{@{}l@{\;\;}l@{\;\;\;}c@{\;\;}c@{\;\;}c@{\;\;}c@{\;\;\;}c@{}}
\toprule
\textbf{Method} & \textbf{Setting} & \textbf{1B} & \textbf{7B} & \textbf{13B} & \textbf{32B} & \textbf{Model-Avg} \\
\midrule
\textit{Zlib (unsup.)} & --- & 0.14 & 0.29 & 0.14 & 0.15 & 0.18 \\
\cmidrule(l){1-7}
Sup-CAMIA & In-domain & \textbf{0.09} & 0.20 & \textbf{0.14} & 0.08 & \textbf{0.13} \\
 & LODO & 0.06 & \textbf{0.23} & 0.05 & \textbf{0.10} & 0.11 \\
\cmidrule(l){1-7}
MIA-Tuner & In-domain & 0.09 & 0.10 & 0.07 & 0.04 & 0.07 \\
 & LODO & 0.05 & 0.03 & 0.07 & 0.03 & 0.04 \\
\bottomrule
\end{tabular}%
}

\vspace{4pt}

\resizebox{0.9\textwidth}{!}{%
\begin{tabular}{@{}lc@{\;\;}c@{\;\;}c@{\;\;}c@{\;\;}c@{\;\;}c@{\;\;\;}c@{\;\;}c@{\;\;}c@{\;\;}c@{\;\;}c@{}}
\toprule
\multirow{2.5}{*}{\textbf{Method}} & \multirow{2.5}{*}{\textbf{Base TPR@5\% FPR}} & \multicolumn{5}{c}{\textbf{In-domain $\Delta$}} & \multicolumn{5}{c}{\textbf{LODO $\Delta$}} \\
\cmidrule(lr){3-7} \cmidrule(lr){8-12}
 & & 1B & 7B & 13B & 32B & \textbf{Model-Avg} & 1B & 7B & 13B & 32B & \textbf{Model-Avg} \\
\midrule
\textbf{FSD (PPL)} & 0.27 & \textbf{$+$0.08} & \textbf{$+$0.03} & \textbf{$+$0.16} & \textbf{$+$0.01} & \textbf{$+$0.07} & \textbf{$+$0.19} & $-$0.01 & \textbf{$+$0.03} & $-$0.08 & \textbf{$+$0.03} \\
\textbf{FSD (Zlib)} & 0.18 & \textbf{$+$0.12} & $-$0.00 & \textbf{$+$0.07} & \textbf{$+$0.12} & \textbf{$+$0.08} & \textbf{$+$0.18} & \textbf{$+$0.08} & \textbf{$+$0.07} & \textbf{$+$0.05} & \textbf{$+$0.09} \\
\textbf{FSD (Lower)} & 0.09 & $-$0.03 & $-$0.01 & \textbf{$+$0.00} & $-$0.02 & $-$0.02 & $-$0.05 & $-$0.05 & \textbf{$+$0.01} & $-$0.03 & $-$0.03 \\
\textbf{FSD (MinK)} & 0.27 & \textbf{$+$0.09} & \textbf{$+$0.01} & \textbf{$+$0.13} & $-$0.09 & \textbf{$+$0.04} & \textbf{$+$0.14} & $-$0.06 & \textbf{$+$0.08} & $-$0.12 & \textbf{$+$0.01} \\
\textbf{FSD (MinK++)} & 0.36 & $-$0.12 & $-$0.09 & \textbf{$+$0.01} & $-$0.20 & $-$0.10 & $-$0.28 & $-$0.21 & $-$0.04 & $-$0.15 & $-$0.17 \\
\bottomrule
\end{tabular}%
}
\caption{Supervised MIAs on the SFT domain of \textsc{OLMo-Detect} (macro-averaged over its subsets), evaluated via TPR@5\% FPR. \textbf{Top:} scores of Sup-CAMIA and MIA-Tuner under the \textbf{In-domain} setting (trained on this domain's dev set) and the \textbf{LODO} setting (trained on the other eight domains' dev sets); Zlib, the strongest unsupervised MIA, is repeated as a reference and is excluded from the bold, which marks the best supervised entry per column. \textbf{Bottom:} FSD measures the change in a base MIA's score after fine-tuning the target model on non-members, so we report its gain ($\Delta$) over the base MIA (in parentheses) on this domain. \textbf{Base TPR@5\% FPR}: the base MIA's own score on this domain averaged across model sizes; positive gains are bolded. \textbf{Model-Avg} averages across model sizes.}
\label{table:tpr_sup_sft}
\end{table*}

\begin{table*}[t]
\centering
\resizebox{0.7\textwidth}{!}{%
\begin{tabular}{@{}l@{\;\;}l@{\;\;\;}c@{\;\;}c@{\;\;}c@{\;\;}c@{\;\;\;}c@{}}
\toprule
\textbf{Method} & \textbf{Setting} & \textbf{1B} & \textbf{7B} & \textbf{13B} & \textbf{32B} & \textbf{Model-Avg} \\
\midrule
\textit{Zlib (unsup.)} & --- & 0.06 & 0.03 & 0.06 & 0.09 & 0.06 \\
\cmidrule(l){1-7}
Sup-CAMIA & In-domain & \textbf{0.21} & 0.06 & \textbf{0.21} & 0.05 & \textbf{0.13} \\
 & LODO & 0.06 & 0.06 & 0.11 & \textbf{0.08} & 0.08 \\
\cmidrule(l){1-7}
MIA-Tuner & In-domain & 0.05 & \textbf{0.08} & 0.12 & 0.06 & 0.08 \\
 & LODO & 0.06 & 0.07 & 0.07 & 0.07 & 0.07 \\
\bottomrule
\end{tabular}%
}

\vspace{4pt}

\resizebox{0.9\textwidth}{!}{%
\begin{tabular}{@{}lc@{\;\;}c@{\;\;}c@{\;\;}c@{\;\;}c@{\;\;}c@{\;\;\;}c@{\;\;}c@{\;\;}c@{\;\;}c@{\;\;}c@{}}
\toprule
\multirow{2.5}{*}{\textbf{Method}} & \multirow{2.5}{*}{\textbf{Base TPR@5\% FPR}} & \multicolumn{5}{c}{\textbf{In-domain $\Delta$}} & \multicolumn{5}{c}{\textbf{LODO $\Delta$}} \\
\cmidrule(lr){3-7} \cmidrule(lr){8-12}
 & & 1B & 7B & 13B & 32B & \textbf{Model-Avg} & 1B & 7B & 13B & 32B & \textbf{Model-Avg} \\
\midrule
\textbf{FSD (PPL)} & 0.04 & $-$0.00 & $-$0.01 & $-$0.02 & \textbf{$+$0.01} & $-$0.01 & $-$0.02 & \textbf{$+$0.01} & $-$0.01 & \textbf{$+$0.01} & $-$0.00 \\
\textbf{FSD (Zlib)} & 0.06 & \textbf{$+$0.03} & $-$0.03 & $-$0.04 & $-$0.01 & $-$0.01 & \textbf{$+$0.00} & $-$0.01 & $-$0.03 & $-$0.03 & $-$0.02 \\
\textbf{FSD (Lower)} & 0.15 & $-$0.01 & \textbf{$+$0.01} & $-$0.01 & $-$0.02 & $-$0.01 & \textbf{$+$0.01} & \textbf{$+$0.05} & \textbf{$+$0.00} & $-$0.03 & \textbf{$+$0.01} \\
\textbf{FSD (MinK)} & 0.05 & $-$0.02 & $-$0.01 & $-$0.02 & \textbf{$+$0.05} & \textbf{$+$0.00} & $-$0.01 & \textbf{$+$0.03} & $-$0.02 & \textbf{$+$0.04} & \textbf{$+$0.01} \\
\textbf{FSD (MinK++)} & 0.04 & \textbf{$+$0.01} & $-$0.01 & \textbf{$+$0.01} & \textbf{$+$0.03} & \textbf{$+$0.01} & $-$0.01 & \textbf{$+$0.01} & $-$0.01 & \textbf{$+$0.05} & \textbf{$+$0.01} \\
\bottomrule
\end{tabular}%
}
\caption{Supervised MIAs on the UF-Wildchat subset of \textsc{OLMo-Detect}'s DPO domain, evaluated via TPR@5\% FPR. \textbf{Top:} scores of Sup-CAMIA and MIA-Tuner under the \textbf{In-domain} setting (trained on this domain's dev set) and the \textbf{LODO} setting (trained on the other eight domains' dev sets); Zlib, the strongest unsupervised MIA, is repeated as a reference and is excluded from the bold, which marks the best supervised entry per column. \textbf{Bottom:} FSD measures the change in a base MIA's score after fine-tuning the target model on non-members, so we report its gain ($\Delta$) over the base MIA (in parentheses) on this domain. \textbf{Base TPR@5\% FPR}: the base MIA's own score on this domain averaged across model sizes; positive gains are bolded. \textbf{Model-Avg} averages across model sizes.}
\label{table:tpr_sup_dpo}
\end{table*}
\clearpage

\begin{table*}[t]
\centering
\resizebox{0.7\textwidth}{!}{%
\begin{tabular}{@{}l@{\;\;}l@{\;\;\;}c@{\;\;}c@{\;\;}c@{\;\;}c@{\;\;\;}c@{}}
\toprule
\textbf{Method} & \textbf{Setting} & \textbf{1B} & \textbf{7B} & \textbf{13B} & \textbf{32B} & \textbf{Model-Avg} \\
\midrule
\textit{Zlib (unsup.)} & --- & 0.35 & 0.65 & 0.81 & 0.68 & 0.62 \\
\cmidrule(l){1-7}
Sup-CAMIA & In-domain & \textbf{0.27} & \textbf{0.64} & \textbf{0.72} & \textbf{0.67} & \textbf{0.58} \\
 & LODO & 0.12 & 0.45 & 0.18 & 0.66 & 0.35 \\
\cmidrule(l){1-7}
MIA-Tuner & In-domain & 0.15 & 0.10 & 0.07 & 0.07 & 0.10 \\
 & LODO & 0.04 & 0.04 & 0.03 & 0.08 & 0.05 \\
\bottomrule
\end{tabular}%
}

\vspace{4pt}

\resizebox{0.9\textwidth}{!}{%
\begin{tabular}{@{}lc@{\;\;}c@{\;\;}c@{\;\;}c@{\;\;}c@{\;\;}c@{\;\;\;}c@{\;\;}c@{\;\;}c@{\;\;}c@{\;\;}c@{}}
\toprule
\multirow{2.5}{*}{\textbf{Method}} & \multirow{2.5}{*}{\textbf{Base TPR@5\% FPR}} & \multicolumn{5}{c}{\textbf{In-domain $\Delta$}} & \multicolumn{5}{c}{\textbf{LODO $\Delta$}} \\
\cmidrule(lr){3-7} \cmidrule(lr){8-12}
 & & 1B & 7B & 13B & 32B & \textbf{Model-Avg} & 1B & 7B & 13B & 32B & \textbf{Model-Avg} \\
\midrule
\textbf{FSD (PPL)} & 0.62 & $-$0.06 & $-$0.51 & $-$0.33 & $-$0.60 & $-$0.37 & $-$0.07 & $-$0.05 & $-$0.11 & $-$0.06 & $-$0.07 \\
\textbf{FSD (Zlib)} & 0.62 & $-$0.23 & $-$0.53 & $-$0.43 & $-$0.59 & $-$0.44 & $-$0.08 & $-$0.01 & $-$0.09 & $-$0.03 & $-$0.05 \\
\textbf{FSD (Lower)} & 0.32 & $-$0.03 & $-$0.18 & $-$0.18 & $-$0.31 & $-$0.17 & $-$0.06 & $-$0.09 & $-$0.01 & \textbf{$+$0.09} & $-$0.02 \\
\textbf{FSD (MinK)} & 0.63 & $-$0.21 & $-$0.57 & $-$0.77 & $-$0.70 & $-$0.56 & $-$0.22 & $-$0.19 & $-$0.13 & $-$0.23 & $-$0.19 \\
\textbf{FSD (MinK++)} & 0.11 & $-$0.01 & $-$0.07 & $-$0.01 & $-$0.06 & $-$0.04 & $-$0.01 & $-$0.01 & $+$0.00 & $-$0.02 & $-$0.01 \\
\bottomrule
\end{tabular}%
}
\caption{Supervised MIAs on the RLVR domain of \textsc{OLMo-Detect} (macro-averaged over its subsets), evaluated via TPR@5\% FPR. \textbf{Top:} scores of Sup-CAMIA and MIA-Tuner under the \textbf{In-domain} setting (trained on this domain's dev set) and the \textbf{LODO} setting (trained on the other eight domains' dev sets); Zlib, the strongest unsupervised MIA, is repeated as a reference and is excluded from the bold, which marks the best supervised entry per column. \textbf{Bottom:} FSD measures the change in a base MIA's score after fine-tuning the target model on non-members, so we report its gain ($\Delta$) over the base MIA (in parentheses) on this domain. \textbf{Base TPR@5\% FPR}: the base MIA's own score on this domain averaged across model sizes; positive gains are bolded. \textbf{Model-Avg} averages across model sizes.}
\label{table:tpr_sup_rlvr}
\end{table*}

\clearpage

\FloatBarrier

\section{Effect of the Labeled-Data Budget on Supervised MIAs}
\label{appendix:learning_curve}

The supervised methods in Section~\ref{sec:sup_results} are trained on each domain's dev split, i.e., 10\% of its members and non-members. To test whether this budget limits their performance, we measure Sup-CAMIA as a function of the number of labeled examples. For each domain and model size, we hold out half of the members and half of the non-members as a fixed evaluation set and train on nested subsets of the remaining half containing 10\%, 20\%, 30\%, 40\%, and 50\% of the domain's members and non-members; the 10\% point matches the budget of the main tables. We repeat this over five random splits and report the mean AUC on the held-out half. Because the evaluation set comprises half of each domain rather than the 90\% test split, absolute AUCs are not comparable to those in the main tables; only the trend is relevant.

Table~\ref{table:lc_supcamia} reports the results. Raising the budget from 10\% to 50\% increases the macro-averaged AUC by 0.04--0.05 at every model size. The gains are confined to the smallest pre-training domains (DCLM-Baseline, OpenWebMath, and StarCoder, with at most 151 members), where the 10\% budget provides only 15--16 labeled members and non-members; peS2o and DPO stay within 0.03 of their 10\% value, and GSM8K and CoT-GSM8K are saturated at every budget. A five-fold larger labeled set therefore does not change the conclusion of Section~\ref{sec:sup_results}: supervision from the matched distribution adds little beyond unsupervised membership signals.

\begin{table*}[t]
\centering
\resizebox{\textwidth}{!}{%
\begin{tabular}{@{}l@{\;\;}r@{\;\;\;}c@{\;\;}c@{\;\;}c@{\;\;}c@{\;\;}c@{\;\;\;}c@{\;\;}c@{\;\;}c@{\;\;}c@{\;\;}c@{\;\;\;}c@{\;\;}c@{\;\;}c@{\;\;}c@{\;\;}c@{\;\;\;}c@{\;\;}c@{\;\;}c@{\;\;}c@{\;\;}c@{}}
\toprule
\multirow{2}{*}{\textbf{Domain}} & \multirow{2}{*}{$N$} & \multicolumn{5}{c}{\textbf{1B}} & \multicolumn{5}{c}{\textbf{7B}} & \multicolumn{5}{c}{\textbf{13B}} & \multicolumn{5}{c}{\textbf{32B}} \\
\cmidrule(lr){3-7} \cmidrule(lr){8-12} \cmidrule(lr){13-17} \cmidrule(lr){18-22}
 & & 10\% & 20\% & 30\% & 40\% & 50\% & 10\% & 20\% & 30\% & 40\% & 50\% & 10\% & 20\% & 30\% & 40\% & 50\% & 10\% & 20\% & 30\% & 40\% & 50\% \\
\midrule
DCLM-Baseline & 150 & 0.52 & 0.57 & 0.56 & 0.57 & 0.58 & 0.56 & 0.58 & 0.62 & 0.62 & 0.63 & 0.58 & 0.56 & 0.59 & 0.60 & 0.60 & 0.58 & 0.60 & 0.63 & 0.63 & 0.64 \\
peS2o & 268 & 0.52 & 0.53 & 0.55 & 0.55 & 0.55 & 0.52 & 0.55 & 0.55 & 0.55 & 0.55 & 0.49 & 0.50 & 0.49 & 0.49 & 0.48 & 0.49 & 0.49 & 0.50 & 0.49 & 0.49 \\
OpenWebMath & 151 & 0.54 & 0.57 & 0.60 & 0.62 & 0.62 & 0.52 & 0.55 & 0.58 & 0.61 & 0.61 & 0.52 & 0.57 & 0.58 & 0.60 & 0.60 & 0.55 & 0.60 & 0.61 & 0.63 & 0.63 \\
StarCoder & 156 & 0.61 & 0.63 & 0.65 & 0.65 & 0.66 & 0.59 & 0.61 & 0.65 & 0.67 & 0.67 & 0.59 & 0.61 & 0.64 & 0.65 & 0.66 & 0.61 & 0.64 & 0.66 & 0.66 & 0.68 \\
Math Mix & 1040 & 0.83 & 0.83 & 0.84 & 0.84 & 0.84 & 0.94 & 0.94 & 0.95 & 0.95 & 0.95 & 1.00 & 1.00 & 1.00 & 1.00 & 1.00 & 1.00 & 1.00 & 1.00 & 1.00 & 1.00 \\
Stack Exchange & 430 & 0.58 & 0.58 & 0.58 & 0.58 & 0.59 & 0.57 & 0.59 & 0.60 & 0.61 & 0.62 & 0.63 & 0.64 & 0.67 & 0.69 & 0.70 & 0.60 & 0.63 & 0.65 & 0.67 & 0.68 \\
SFT & 164--225 & 0.59 & 0.62 & 0.62 & 0.64 & 0.64 & 0.67 & 0.76 & 0.77 & 0.78 & 0.77 & 0.58 & 0.64 & 0.64 & 0.64 & 0.65 & 0.58 & 0.59 & 0.62 & 0.63 & 0.64 \\
DPO & 110--113 & 0.61 & 0.61 & 0.63 & 0.64 & 0.65 & 0.54 & 0.59 & 0.60 & 0.61 & 0.61 & 0.64 & 0.65 & 0.66 & 0.67 & 0.67 & 0.58 & 0.59 & 0.61 & 0.61 & 0.63 \\
RLVR & 290 & 0.73 & 0.78 & 0.80 & 0.80 & 0.80 & 0.86 & 0.87 & 0.87 & 0.88 & 0.89 & 0.94 & 0.95 & 0.95 & 0.95 & 0.95 & 0.90 & 0.92 & 0.92 & 0.92 & 0.92 \\
\midrule
\textbf{Macro-Avg} & & \textbf{0.61} & \textbf{0.63} & \textbf{0.65} & \textbf{0.65} & \textbf{0.66} & \textbf{0.64} & \textbf{0.67} & \textbf{0.69} & \textbf{0.70} & \textbf{0.70} & \textbf{0.66} & \textbf{0.68} & \textbf{0.69} & \textbf{0.70} & \textbf{0.70} & \textbf{0.65} & \textbf{0.67} & \textbf{0.69} & \textbf{0.69} & \textbf{0.70} \\
\bottomrule
\end{tabular}%
}
\caption{AUC of Sup-CAMIA as a function of the labeled-data budget on \textsc{OLMo-Detect}. For each subset and model size, half of the members and half of the non-members are held out as a fixed evaluation set; Sup-CAMIA is trained on nested subsets of the remaining half containing 10\%--50\% of the subset's members and non-members (10\% equals the dev-set budget of the main tables). Entries are the mean AUC on the held-out half over five random splits, averaged over the domain's subsets (DPO over the chosen and rejected responses); $N$ is the number of members (equal to the number of non-members) in the domain. SFT and DPO members come from the size-specific post-training mixes (SFT: one mix shared by 1B and 32B and another by 7B and 13B; DPO: one preference mix per size), so their $N$ varies across model sizes and is given as a range. Macro-Avg is the mean over the nine domains. At the subset level the standard deviation across splits is below 0.05 in 211 of 260 cases; it reaches 0.17 for the smallest subsets (at most 100 members) at the 10\%--20\% budgets and is at most 0.09 at budgets of 30\% or more. Absolute values are not comparable to Table~\ref{table:olmo_detect_overall_supervised}, which uses the 90\% test split.}
\label{table:lc_supcamia}
\end{table*}

\begin{table}[t]
\centering\small
\begin{tabular}{@{}lcccc@{}}
\toprule
\multirow{2.5}{*}{\textbf{Method}} & \multicolumn{2}{c}{\textbf{Given the domain}} & \multicolumn{2}{c}{\textbf{Domain unknown}} \\
\cmidrule(lr){2-3} \cmidrule(lr){4-5}
  & AUC (macro) & Acc. & AUC (pooled) & Acc. \\
\midrule
PPL$^{\mathrm{L}}$ & 0.60 & 0.57 & 0.56 & 0.56 \\
Zlib$^{\mathrm{L}}$ & \textbf{0.64} & \textbf{0.59} & 0.59 & 0.57 \\
Lower$^{\mathrm{L}}$ & 0.61 & 0.57 & 0.59 & 0.57 \\
MinK$^{\mathrm{L}}$ & 0.61 & 0.58 & 0.58 & 0.57 \\
MinK++$^{\mathrm{L}}$ & 0.54 & 0.52 & 0.52 & 0.53 \\
DCPDD$^{\mathrm{L}}$ & 0.63 & 0.59 & 0.60 & 0.58 \\
PAC$^{\mathrm{L}}$ & 0.63 & 0.58 & 0.59 & 0.58 \\
ReCaLL$^{\mathrm{L}}$ & 0.48 & 0.52 & 0.50 & 0.53 \\
EMMIA$^{\mathrm{L}}$ & 0.55 & --- & 0.51 & --- \\
CAMIA$^{\mathrm{L}}$ & 0.62 & 0.59 & \textbf{0.62} & \textbf{0.60} \\
\cmidrule(lr){1-5}
Neigh.$^{\mathrm{P}}$ & 0.57 & 0.53 & 0.55 & 0.54 \\
\cmidrule(lr){1-5}
Guided$^{\mathrm{Pr}}$ & 0.49 & 0.48 & 0.49 & 0.50 \\
DCQ$^{\mathrm{Pr}}$ & 0.49 & 0.50 & 0.50 & 0.53 \\
\cmidrule(lr){1-5}
CDD$^{\mathrm{O}}$ & 0.52 & 0.50 & 0.51 & 0.50 \\
SelfCrit$^{\mathrm{O}}$$^\dagger$ & 0.48 & --- & 0.48 & --- \\
\midrule
\textbf{Method-Avg (unsup.)} & 0.56 & 0.55 & 0.55 & 0.55 \\
\midrule
\textbf{Sup-CAMIA} & \textbf{0.66} & \textbf{0.62} & \textbf{0.60} & \textbf{0.58} \\
\textbf{MIA-Tuner} & 0.56 & 0.52 & 0.55 & 0.54 \\
\bottomrule
\end{tabular}
\caption{Two evaluation settings on \textsc{OLMo-Detect}, averaged over model sizes (1B--32B). Every domain is subsampled to 100 members and 100 non-members (20 random draws; multi-subset domains pooled over their subsets; methods, dev sets and hyperparameters exactly as in the main tables). \textbf{Given the domain}: AUC per domain, macro-averaged (the paper's protocol), and accuracy with a decision threshold tuned within the evaluated domain (2-fold). \textbf{Domain unknown}: one AUC over the pooled instances of all nine domains, and accuracy with a single threshold tuned on the pooled data; supervised MIAs are retrained on the pooled dev sets of all nine domains. Thresholds maximise accuracy on their tuning data. Best per column and block in bold. $^\dagger$SelfCrit is evaluated on DPO and RLVR only. Accuracies are omitted for EMMIA, which reports the best of its four base scores per subset and thus has no single score to threshold and for SelfCrit.}
\label{table:three_settings}
\end{table}

\section{Domain-Unknown Evaluation}
\label{appendix:three_settings}

The main text evaluates MIAs \emph{given the domain}: each domain is scored separately and the AUCs are macro-averaged, which assumes the auditor knows which domain a text comes from. In practice that is rarely true. To simulate the realistic case, the \emph{domain-unknown} setting pools members and non-members from all nine domains and computes a single AUC, so a score must separate members from non-members in general rather than within a known domain.\footnote{This differs from the LODO setting: the test instance still comes from a domain the MIA was tuned on, we simply do not know which one.} Table~\ref{table:three_settings} contrasts the two settings on the same subsampled data, 100 members and 100 non-members per domain over 20 draws, with supervised MIAs retrained on the pooled dev sets. For comparison on a common metric we also report accuracy, with a threshold tuned within each evaluated domain in the first setting and a single threshold tuned on the pooled data in the second.

Pooling lowers AUC for 12 of the 15 unsupervised MIAs. The three that rise, ReCaLL, Guided and DCQ, are at or below chance in both settings, so the increase carries no information. The largest drops fall on Zlib ($0.64$ to $0.59$), PPL ($0.60$ to $0.56$) and EMMIA ($0.55$ to $0.51$), while CAMIA is unaffected ($0.62$ in both), making it the strongest method once the domain is unknown. Averaged over unsupervised MIAs the cost is $0.02$ AUC. Supervised MIAs pay more, but unevenly: Sup-CAMIA falls from $0.66$ to $0.60$, having lost the ability to calibrate to a single domain, whereas MIA-Tuner is almost unchanged at $0.56$ to $0.55$. Accuracy tells the same story, with the unsupervised average flat at $0.55$ and Sup-CAMIA falling from $0.62$ to $0.58$. Crucially the ranking survives: Spearman $\rho = 0.95$ between the two settings across the 17 MIAs. Absolute performance therefore declines when the domain is unknown, but the conclusions of the main text do not depend on knowing it.

\section{Per-Domain Matched-vs-Shifted Results}
\label{appendix:distribution_shifts}

Table~\ref{table:distribution_shifts_per_domain} reports the per-domain changes behind the aggregate counts of Table~\ref{table:distribution_shifts}. Each entry is the AUC on \textsc{OLMo-Detect (Shifted)} minus the AUC on \textsc{OLMo-Detect}, averaged over the four model sizes, with significance from a paired instance-level bootstrap. Two patterns are visible directly. The direction of the change is a property of the method--domain pair rather than of the domain: OpenWebMath inflates PPL, MinK, CAMIA and EMMIA but degrades Zlib, Lower and PAC. And the largest change of every MIA falls in a pre-training domain, most often OpenWebMath (7 of 14 MIAs), then StarCoder (4) and peS2o (3), while DCLM never carries a method's largest change despite being shifted on the same axes.

\begin{table*}[t]
\centering
\resizebox{\textwidth}{!}{%
\begin{tabular}{@{}l@{\;\;}c@{\;\;}c@{\;\;}c@{\;\;}c@{\;\;}c@{\;\;}c@{\;\;}c@{\;\;\;}c@{}}
\toprule
\textbf{Method} & \textbf{DCLM} & \textbf{peS2o} & \textbf{OWM} & \textbf{SC} & \textbf{SE} & \textbf{SFT} & \textbf{DPO} & \textbf{Largest} \\
\midrule
PPL & \textbf{$+$0.06}$^{\ddagger}$ & \textbf{$+$0.22}$^{\ddagger}$ & \textbf{$+$0.29}$^{\ddagger}$ & \textbf{$+$0.04}$^{\dagger}$ & \textbf{$-$0.06}$^{\ddagger}$ & \textbf{$-$0.08}$^{\ddagger}$ & $-$0.01 & OWM \\
Zlib & $+$0.01 & \textbf{$-$0.20}$^{\ddagger}$ & \textbf{$-$0.31}$^{\ddagger}$ & \textbf{$+$0.25}$^{\ddagger}$ & \textbf{$+$0.05}$^{\ddagger}$ & $-$0.01 & $+$0.01 & OWM \\
Lower & \textbf{$+$0.05}$^{\ddagger}$ & \textbf{$+$0.06}$^{\ddagger}$ & \textbf{$-$0.30}$^{\ddagger}$ & \textbf{$+$0.14}$^{\ddagger}$ & \textbf{$+$0.04}$^{\ddagger}$ & \textbf{$+$0.06}$^{\ddagger}$ & $+$0.01 & OWM \\
MinK & $+$0.02 & \textbf{$+$0.21}$^{\ddagger}$ & \textbf{$+$0.23}$^{\ddagger}$ & $+$0.03 & $-$0.02 & \textbf{$-$0.06}$^{\ddagger}$ & $+$0.01 & OWM \\
MinK++ & \textbf{$-$0.06}$^{\ddagger}$ & \textbf{$-$0.03}$^{\dagger}$ & $-$0.02 & \textbf{$-$0.16}$^{\ddagger}$ & \textbf{$+$0.07}$^{\ddagger}$ & $-$0.02 & $+$0.01 & SC \\
DCPDD & $+$0.03 & \textbf{$+$0.04}$^{\ddagger}$ & \textbf{$-$0.05}$^{\ddagger}$ & \textbf{$+$0.16}$^{\ddagger}$ & $+$0.02 & \textbf{$-$0.10}$^{\ddagger}$ & \textbf{$-$0.04}$^{\ddagger}$ & SC \\
PAC & \textbf{$+$0.05}$^{\dagger}$ & \textbf{$+$0.09}$^{\ddagger}$ & \textbf{$-$0.20}$^{\ddagger}$ & \textbf{$+$0.19}$^{\ddagger}$ & $-$0.00 & \textbf{$+$0.04}$^{\dagger}$ & \textbf{$+$0.09}$^{\ddagger}$ & OWM \\
ReCaLL & \textbf{$+$0.06}$^{\ddagger}$ & \textbf{$+$0.24}$^{\ddagger}$ & \textbf{$+$0.04}$^{\ddagger}$ & \textbf{$+$0.13}$^{\ddagger}$ & $+$0.01 & $+$0.02 & $+$0.01 & peS2o \\
EMMIA & $-$0.01 & \textbf{$-$0.17}$^{\ddagger}$ & \textbf{$+$0.42}$^{\ddagger}$ & \textbf{$+$0.06}$^{\dagger}$ & \textbf{$-$0.06}$^{\ddagger}$ & \textbf{$+$0.06}$^{\ddagger}$ & $+$0.02 & OWM \\
CAMIA & $+$0.04 & \textbf{$+$0.28}$^{\ddagger}$ & \textbf{$+$0.18}$^{\ddagger}$ & \textbf{$+$0.23}$^{\ddagger}$ & \textbf{$-$0.12}$^{\ddagger}$ & $+$0.00 & \textbf{$+$0.03}$^{\dagger}$ & peS2o \\
Neigh. & $+$0.03 & \textbf{$+$0.10}$^{\ddagger}$ & \textbf{$+$0.04}$^{\dagger}$ & \textbf{$+$0.12}$^{\ddagger}$ & \textbf{$-$0.03}$^{\dagger}$ & \textbf{$+$0.06}$^{\ddagger}$ & \textbf{$+$0.06}$^{\ddagger}$ & SC \\
Guided & $+$0.03 & \textbf{$-$0.05}$^{\ddagger}$ & $+$0.03 & \textbf{$+$0.10}$^{\ddagger}$ & $+$0.00 & \textbf{$-$0.03}$^{\dagger}$ & $-$0.01 & SC \\
DCQ & $-$0.02 & \textbf{$+$0.09}$^{\ddagger}$ & \textbf{$-$0.03}$^{\dagger}$ & $+$0.01 & \textbf{$+$0.03}$^{\dagger}$ & $+$0.02 & $-$0.01 & peS2o \\
CDD & $+$0.01 & \textbf{$+$0.05}$^{\ddagger}$ & \textbf{$+$0.12}$^{\ddagger}$ & \textbf{$+$0.04}$^{\ddagger}$ & $-$0.01 & $-$0.01 & $-$0.01 & OWM \\
\bottomrule
\end{tabular}%
}
\caption{Per-domain effect of distribution shift on unsupervised MIAs, $\Delta$ = AUC on \textsc{OLMo-Detect (Shifted)} minus AUC on \textsc{OLMo-Detect}, averaged over the four model sizes. Significance is assessed with a paired instance-level bootstrap: $^{\dagger}$ the $95\%$ interval excludes zero, $^{\ddagger}$ the $99\%$ interval excludes zero; significant entries are in bold. \textbf{Largest}: the domain carrying that MIA's largest change. GSM8K and RLVR have no shifted variant. OWM: OpenWebMath; SC: StarCoder; SE: Stack Exchange. Aggregate counts are in Table~\ref{table:distribution_shifts}.}
\label{table:distribution_shifts_per_domain}
\end{table*}

\section{Non-Member Verification for SmolLM2}
\label{appendix:smollm2_data}

SmolLM2-1.7B is trained on DCLM together with FineWeb-Edu, both publicly released. Its members therefore overlap \textsc{OLMo-Detect}'s DCLM domain by construction, but its non-members are valid only if they are absent from \emph{every} corpus SmolLM2 saw, not from DCLM alone. Because \textsc{OLMo-Detect}'s DCLM non-members are drawn from 25 Common Crawl dumps spanning 2013--2021, and FineWeb-Edu deduplicates within rather than across dumps, absence from DCLM does not imply absence from FineWeb-Edu.

We therefore applied the same 13-gram filter used in benchmark construction, with the same $0.20$ threshold, to the whole of FineWeb-Edu: 2{,}410 files and $1.53 \times 10^{9}$ documents, sharded nine ways across the fleet. Eight of the 134 DCLM test non-members exceed the threshold, at overlaps of $0.855$, $0.797$, $0.790$, $0.746$, $0.746$, $0.733$, $0.661$ and $0.616$, together with one of the 16 development non-members. The remaining 126 reach at most $0.128$, so the separation is unambiguous rather than borderline. The excluded indices are released with the benchmark.

We evaluate SmolLM2 on the 126 verified non-members, and re-evaluate OLMo~2-1B on exactly the same set so that the comparison in Table~\ref{table:generalization} is paired. Members are not filtered: a SmolLM2 member appearing in FineWeb-Edu is expected and harmless, since it remains a member of that model's training data.

\end{document}